\documentclass[11pt]{article}

\usepackage[preprint]{acl}

\usepackage{booktabs}
\usepackage{multirow}
\usepackage{amsmath}

\usepackage{times}
\usepackage{latexsym}
\usepackage{dblfloatfix}
\usepackage{placeins}
\usepackage{subcaption}

\usepackage{pgfplots}
\usepgfplotslibrary{groupplots}
\pgfplotsset{compat=1.18}

\usepackage[T1]{fontenc}

\usepackage[utf8]{inputenc}
\usepackage{amssymb}

\usepackage{microtype}

\usepackage{inconsolata}

\usepackage[table]{xcolor}
\usepackage{array}
\usepackage{dblfloatfix}
\usepackage{tikz}
\usetikzlibrary{arrows.meta, positioning, fit, calc}
\usepackage{fontawesome5}

\usepackage{graphicx}
\graphicspath{{latex/}}
\usepackage{todonotes}

\definecolor{jack}{rgb}{1.0, 0.6, 0.4}
\definecolor{jerry}{rgb}{0.6, 0.2, 1.0}
\definecolor{tengfei}{rgb}{0.5, 0.2, 0.8}

\title{Do LLMs Understand Sequential Structure? A Controlled Study of Inference and Generation}

\author{
\textbf{Jerry Wang}\textsuperscript{1} \quad
\textbf{Zhengxiang Wang}\textsuperscript{2} \quad
\textbf{Ting Yu Liu}\textsuperscript{3}
\\
\textbf{Hsin-Ling Hsu}\textsuperscript{3} \quad
\textbf{Yi-Cheng Lai}\textsuperscript{3} \quad
\textbf{Tengfei Ma}\textsuperscript{2}
\\[0.6em]
\textsuperscript{1}University of Illinois Urbana-Champaign
\\
\textsuperscript{2}Stony Brook University
\\
\textsuperscript{3}National Chengchi University
}

\begin{document}

\maketitle

\begin{abstract}
% Large language models (LLMs) are increasingly used as interactive agents and simulators, but it remains unclear whether they can distinguish surface-level action frequencies from underlying sequential strategies. We study this question in a controlled two-player Rock--Paper--Scissors setting, where players follow either statistical policies with fixed marginal action distributions or Markov policies whose actions depend on recent interaction history. Across three experiments, we evaluate whether LLMs can identify latent player strategies from observed trajectories, generate future trajectories under simple first-order Markov rules, and maintain more complex history-dependent rules over long horizons. Our framework separates distribution matching from conditional rule following, allowing us to test whether models recover genuine temporal structure rather than merely reproducing marginal statistics. We further conduct a one-player stochastic continuation analysis to remove player-attribution effects and test whether models can use and recover higher-order transition dynamics. Our results show that recognition, marginal distribution matching, and conditional rule execution are separable abilities, providing a controlled evaluation of LLMs' ability to infer, represent, and sustain latent sequential dynamics in interactive settings.

Large language models (LLMs) are increasingly used as interactive agents and simulators, yet it remains unclear whether they can recover latent sequential structure beyond surface action frequencies. \textbf{This distinction is critical for behavioral simulation, where actions are often shaped by prior context rather than marginal frequencies alone.} We study this question using controlled two-player Rock--Paper--Scissors interactions and a one-player stochastic $n$-gram continuation task. Across these experiments, we test whether LLMs can identify latent strategies, follow simple Markov rules, and sustain higher-order conditional dependencies. Our framework separates distribution matching from conditional rule following. Results show that longer context does not improve identification, correct recognition does not ensure faithful simulation, and higher-order dependencies substantially degrade rule recovery. Apparent behavioral fidelity can therefore mask incorrect generative mechanisms.

% \jack{Does this conclusion sound true?}
% \jerry{YES}
\end{abstract}

\section{Introduction}

Large language models (LLMs) are increasingly used as interactive agents and behavioral simulators. These applications reflect the sequential nature of human behavior, where a user's response to a recommendation, question, or system action depends on what has happened earlier in the interaction and shapes what happens next. Recent work has used LLMs to simulate human-like agents in sandbox environments, evaluate social intelligence through open-ended interaction, and complete multi-turn tool-use tasks with simulated users \citep{park2023generativeagentsinteractivesimulacra, zhou2024sotopia, ma2024agentboardanalyticalevaluationboard, yao2025taubench}. These settings require models not only to produce plausible actions, but also to infer latent strategies, track evolving context, and act consistently across turns.

A key challenge is that observed behavior does not uniquely determine the process that generated it. As illustrated in Figure~\ref{fig:spirit-figure}, two action trajectories may exhibit similar surface statistics while arising from different latent mechanisms. One may be generated by independent sampling from a fixed distribution, whereas another may depend on recent history. Matching aggregate behavior therefore does not establish that a model has recovered the conditional structure that generated the sequence.

% ======================================================
% Figure 1: Observed behavior -> proportions
%           -> latent generators -> policies
%
% Preamble:
% \usepackage{tikz}
% \usetikzlibrary{arrows.meta}
% \usepackage{fontawesome5}
% ======================================================

\newcommand{\rpsicon}[1]{\makebox[0.42cm][c]{#1}}
\newcommand{\rock}{\rpsicon{\faIcon[regular]{hand-rock}}}
\newcommand{\paper}{\rpsicon{\faIcon[regular]{hand-paper}}}
\newcommand{\scissors}{\rpsicon{\faIcon[regular]{hand-scissors}}}

% ---------- restrained paper-friendly palette ----------
\definecolor{figborder}{RGB}{156,163,175}
\definecolor{figtext}{RGB}{55,65,81}
\definecolor{surfacebg}{RGB}{247,249,251}
\definecolor{panelbg}{RGB}{255,255,255}
\definecolor{claimbg}{RGB}{239,247,241}
\definecolor{statbg}{RGB}{235,244,252}
\definecolor{markovbg}{RGB}{252,241,229}
\definecolor{stataccent}{RGB}{87,126,160}
\definecolor{markovaccent}{RGB}{171,116,72}

\begin{figure}[t]
\centering
\resizebox{0.98\linewidth}{!}{%
\begin{tikzpicture}[
    font=\small,
    >={Latex[length=2.0mm,width=1.4mm]},
    flow/.style={
        ->,
        draw=figborder,
        line width=0.75pt
    },
    dep/.style={
        ->,
        draw=markovaccent,
        line width=0.75pt,
        shorten >=1pt,
        shorten <=1pt
    },
    outer/.style={
        draw=figborder!75,
        line width=0.55pt,
        rounded corners=3mm,
        fill=surfacebg
    },
    panel/.style={
        draw=figborder!55,
        line width=0.45pt,
        rounded corners=2.2mm,
        fill=panelbg
    },
    generator/.style={
        draw=figborder!70,
        line width=0.55pt,
        rounded corners=2.7mm
    },
    policy/.style={
        draw=figborder!65,
        line width=0.50pt,
        rounded corners=2.2mm,
        align=center
    }
]

% ======================================================
% LAYER 1 — OBSERVED BEHAVIOR
% ======================================================

\node[
    outer,
    minimum width=10.35cm,
    minimum height=3.20cm
] (observed) at (0,0) {};

\node[
    font=\bfseries\small,
    text=figtext
] at (0,1.32) {Observed behavior};

% ------------------------------------------------------
% Interaction A
% ------------------------------------------------------

\node[
    panel,
    minimum width=4.60cm,
    minimum height=2.15cm
] (panelA) at (-2.55,-0.05) {};

\node[
    font=\scriptsize\bfseries,
    text=figtext
] at (-2.55,0.72) {Interaction A};

\node[
    font=\tiny\bfseries,
    anchor=east,
    text=figtext
] at (-4.12,0.28) {P1};

\node[
    font=\tiny\bfseries,
    anchor=east,
    text=figtext
] at (-4.12,-0.14) {P2};

% P1: R P S R P S
\node[font=\tiny] at (-3.78,0.28) {\rock};
\node[font=\tiny] at (-3.32,0.28) {\paper};
\node[font=\tiny] at (-2.86,0.28) {\scissors};
\node[font=\tiny] at (-2.40,0.28) {\rock};
\node[font=\tiny] at (-1.94,0.28) {\paper};
\node[font=\tiny] at (-1.48,0.28) {\scissors};

% P2: S R P S R P
\node[font=\tiny] at (-3.78,-0.14) {\scissors};
\node[font=\tiny] at (-3.32,-0.14) {\rock};
\node[font=\tiny] at (-2.86,-0.14) {\paper};
\node[font=\tiny] at (-2.40,-0.14) {\scissors};
\node[font=\tiny] at (-1.94,-0.14) {\rock};
\node[font=\tiny] at (-1.48,-0.14) {\paper};

% compact visual ratio strip
\draw[
    rounded corners=0.7pt,
    fill=stataccent!28,
    draw=none
] (-3.38,-0.72) rectangle (-2.94,-0.53);

\draw[
    rounded corners=0.7pt,
    fill=stataccent!45,
    draw=none
] (-2.90,-0.72) rectangle (-2.46,-0.53);

\draw[
    rounded corners=0.7pt,
    fill=stataccent!62,
    draw=none
] (-2.42,-0.72) rectangle (-1.98,-0.53);

\node[
    font=\tiny,
    text=figtext
] at (-2.68,-0.93) {$R:P:S=1:1:1$};

% ------------------------------------------------------
% Interaction B
% ------------------------------------------------------

\node[
    panel,
    minimum width=4.60cm,
    minimum height=2.15cm
] (panelB) at (2.55,-0.05) {};

\node[
    font=\scriptsize\bfseries,
    text=figtext
] at (2.55,0.72) {Interaction B};

\node[
    font=\tiny\bfseries,
    anchor=east,
    text=figtext
] at (0.98,0.31) {P2};

\node[
    font=\tiny\bfseries,
    anchor=east,
    text=figtext
] at (0.98,-0.25) {P1};

% P2: R P S R P S
\node[font=\tiny] (b21) at (1.32,0.31) {\rock};
\node[font=\tiny] (b22) at (1.78,0.31) {\paper};
\node[font=\tiny] (b23) at (2.24,0.31) {\scissors};
\node[font=\tiny] (b24) at (2.70,0.31) {\rock};
\node[font=\tiny] (b25) at (3.16,0.31) {\paper};
\node[font=\tiny] (b26) at (3.62,0.31) {\scissors};

% P1: R P S R P S
\node[font=\tiny] (b11) at (1.32,-0.25) {\rock};
\node[font=\tiny] (b12) at (1.78,-0.25) {\paper};
\node[font=\tiny] (b13) at (2.24,-0.25) {\scissors};
\node[font=\tiny] (b14) at (2.70,-0.25) {\rock};
\node[font=\tiny] (b15) at (3.16,-0.25) {\paper};
\node[font=\tiny] (b16) at (3.62,-0.25) {\scissors};

% previous-opponent-action dependencies
\draw[dep] (b21.south) -- (b12.north);
\draw[dep] (b22.south) -- (b13.north);
\draw[dep] (b23.south) -- (b14.north);
\draw[dep] (b24.south) -- (b15.north);
\draw[dep] (b25.south) -- (b16.north);

% same compact visual ratio strip
\draw[
    rounded corners=0.7pt,
    fill=stataccent!28,
    draw=none
] (1.72,-0.72) rectangle (2.16,-0.53);

\draw[
    rounded corners=0.7pt,
    fill=stataccent!45,
    draw=none
] (2.20,-0.72) rectangle (2.64,-0.53);

\draw[
    rounded corners=0.7pt,
    fill=stataccent!62,
    draw=none
] (2.68,-0.72) rectangle (3.12,-0.53);

\node[
    font=\tiny,
    text=figtext
] at (2.42,-0.93) {$R:P:S=1:1:1$};

% icon legend
\node[
    font=\scriptsize,
    text=figtext
] at (0,-1.27) {
    \rock\, Rock
    \qquad
    \paper\, Paper
    \qquad
    \scissors\, Scissors
};

% ======================================================
% LAYER 2 — SAME SURFACE PROPORTIONS
% ======================================================

\node[
    draw=figborder!70,
    line width=0.50pt,
    rounded corners=2.6mm,
    fill=claimbg,
    minimum width=4.70cm,
    minimum height=0.92cm,
    align=center
] (ratio) at (0,-3.00) {
    \bfseries Similar action proportions
    \\[-1pt]
    \scriptsize\mdseries $R:P:S=1:1:1$
};

\draw[flow] (observed.south) -- (ratio.north);

% ======================================================
% LAYER 3 — LATENT GENERATORS
% ======================================================

% ------------------------------------------------------
% Independent generator
% ------------------------------------------------------

\node[
    generator,
    fill=statbg,
    minimum width=4.15cm,
    minimum height=2.35cm
] (indgen) at (-2.45,-5.25) {};

% probability bars — visual first
\draw[
    draw=figborder!60,
    line width=0.40pt
] (-3.55,-5.18) -- (-2.45,-5.18);

\draw[
    fill=stataccent!40,
    draw=none,
    rounded corners=0.7pt
] (-3.45,-5.18) rectangle (-3.20,-4.62);

\draw[
    fill=stataccent!58,
    draw=none,
    rounded corners=0.7pt
] (-3.08,-5.18) rectangle (-2.83,-4.62);

\draw[
    fill=stataccent!76,
    draw=none,
    rounded corners=0.7pt
] (-2.71,-5.18) rectangle (-2.46,-4.62);

\node[font=\tiny, text=figtext] at (-3.325,-5.38) {R};
\node[font=\tiny, text=figtext] at (-2.955,-5.38) {P};
\node[font=\tiny, text=figtext] at (-2.585,-5.38) {S};

% sample arrow
\draw[
    ->,
    draw=stataccent,
    line width=0.70pt
] (-2.22,-4.90) -- (-1.55,-4.90);

\node[
    circle,
    draw=stataccent,
    line width=0.55pt,
    fill=white,
    minimum size=0.58cm,
    inner sep=0pt,
    font=\scriptsize\bfseries,
    text=figtext
] at (-1.25,-4.90) {$a_t$};

% caption inside card
\node[
    font=\footnotesize\bfseries,
    text=figtext
] at (-2.45,-5.88) {Independent generator};

\node[
    font=\tiny,
    text=figtext!72
] at (-2.45,-6.14) {sample without history};

% ------------------------------------------------------
% History-dependent generator
% ------------------------------------------------------

\node[
    generator,
    fill=markovbg,
    minimum width=4.15cm,
    minimum height=2.35cm
] (histgen) at (2.45,-5.25) {};

% column labels
\node[
    font=\tiny,
    text=figtext!72
] at (1.72,-4.46) {previous};

\node[
    font=\tiny,
    text=figtext!72
] at (3.18,-4.46) {next};

% previous-action column
\node[
    circle,
    draw=markovaccent!80,
    fill=white,
    line width=0.50pt,
    minimum size=0.50cm,
    inner sep=0pt,
    font=\scriptsize,
    text=figtext
] (mr1) at (1.72,-4.84) {R};

\node[
    circle,
    draw=markovaccent!80,
    fill=white,
    line width=0.50pt,
    minimum size=0.50cm,
    inner sep=0pt,
    font=\scriptsize,
    text=figtext
] (mp1) at (1.72,-5.20) {P};

\node[
    circle,
    draw=markovaccent!80,
    fill=white,
    line width=0.50pt,
    minimum size=0.50cm,
    inner sep=0pt,
    font=\scriptsize,
    text=figtext
] (ms1) at (1.72,-5.56) {S};

% next-action column
\node[
    circle,
    draw=markovaccent!80,
    fill=white,
    line width=0.50pt,
    minimum size=0.50cm,
    inner sep=0pt,
    font=\scriptsize,
    text=figtext
] (mr2) at (3.18,-4.84) {P};

\node[
    circle,
    draw=markovaccent!80,
    fill=white,
    line width=0.50pt,
    minimum size=0.50cm,
    inner sep=0pt,
    font=\scriptsize,
    text=figtext
] (mp2) at (3.18,-5.20) {S};

\node[
    circle,
    draw=markovaccent!80,
    fill=white,
    line width=0.50pt,
    minimum size=0.50cm,
    inner sep=0pt,
    font=\scriptsize,
    text=figtext
] (ms2) at (3.18,-5.56) {R};

% explicit conditional mappings
\draw[
    ->,
    draw=markovaccent,
    line width=0.70pt
] (mr1.east) -- (mr2.west);

\draw[
    ->,
    draw=markovaccent,
    line width=0.70pt
] (mp1.east) -- (mp2.west);

\draw[
    ->,
    draw=markovaccent,
    line width=0.70pt
] (ms1.east) -- (ms2.west);

% caption inside card
\node[
    font=\footnotesize\bfseries,
    text=figtext
] at (2.45,-6.14) {History-dependent generator};

% symmetric branching from Layer 2
\draw[flow]
    (ratio.south west)
    .. controls (-0.35,-3.80) and (-2.45,-3.65) ..
    (indgen.north);

\draw[flow]
    (ratio.south east)
    .. controls (0.35,-3.80) and (2.45,-3.65) ..
    (histgen.north);

% ======================================================
% LAYER 4 — POLICIES
% ======================================================

\node[
    policy,
    fill=statbg!72,
    minimum width=4.15cm,
    minimum height=1.18cm
] (statpolicy) at (-2.45,-7.35) {
    {\bfseries Statistical policy}\\[-1pt]
    \footnotesize $P(a_t)=P(a)$
};

\node[
    policy,
    fill=markovbg!78,
    minimum width=4.15cm,
    minimum height=1.18cm
] (markovpolicy) at (2.45,-7.35) {
    {\bfseries Markov policy}\\[-1pt]
    \footnotesize $P(a_t\mid h_t)$
};

\draw[flow] (indgen.south) -- (statpolicy.north);
\draw[flow] (histgen.south) -- (markovpolicy.north);

\end{tikzpicture}%
}

\caption{
Similar action proportions can arise from different latent generative processes and policies.
}
\label{fig:spirit-figure}
\end{figure}
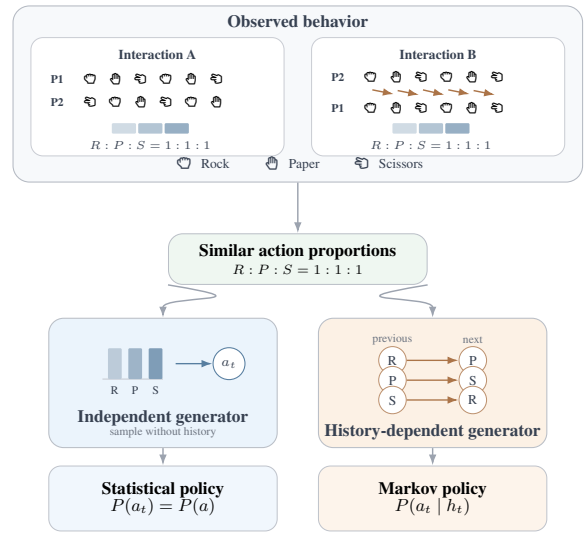

This distinction is fundamental in sequence modeling, where Markov and $n$-gram models explicitly represent history-conditioned distributions \citep{shannon1948mathematical, chen-goodman-1996-empirical}. It is also relevant for LLM-based agents, which may need to infer and respond to other agents over repeated interactions \citep{akata2024playing, yu-etal-2025-llm}. Controlled model problems have also been widely used to study what Transformers infer and implement in context, spanning simple function classes and regression to formal languages, Markov chains, and hidden Markov processes \citep{3600270.3602487, pmlr-v202-von-oswald23a, pmlr-v235-akyurek24a, edelman2024the, NEURIPS2025_29a0ea49}. What remains less clear is whether LLMs can recover the conditional process behind an observed trajectory and sustain it during generation.

We study this question using a controlled Rock--Paper--Scissors (RPS) environment in which the latent generating process is known exactly, allowing us to separately evaluate strategy identification, distributional fidelity, and conditional rule execution. Our three RPS experiments progress from identifying latent strategies, to inference-and-simulation under first-order Markov rules, to controlled variation of joint-state and second-order dependencies. We then remove player interaction and RPS semantics through a one-player stochastic $n$-gram continuation task, extending dependency order from one to eight. Across these settings, more observed context does not reliably improve strategy recovery, correct identification does not guarantee faithful rule execution, and performance degrades most clearly as conditional dependencies become higher-order. This degradation persists in the one-player setting even with longer prefixes or explicit transition rules, while a prefix-only empirical $n$-gram estimator remains comparatively stable at high order. These results identify the recovery and sustained execution of higher-order conditional structure as a key bottleneck in sequential generation.

\section{Related Work}
\paragraph{Game-Based Evaluation of LLM Strategic Behavior.}
Recent work has used games as controlled environments for evaluating LLM strategic reasoning, hidden-state inference, and interactive decision-making \citep{costarelli2024gamebenchevaluatingstrategicreasoning, wang2025tmgbench, light2023avalonbenchevaluatingllmsplaying, wu-etal-2024-deciphering}. 
Related studies further show that LLMs can display strategic behavior in repeated games and can benefit from modeling opponents in game settings \citep{akata2024playing, yu-etal-2025-llm}. 
These works mainly ask whether LLM agents can reason strategically, infer hidden information, model an opponent, or achieve strong game-level performance. 
In contrast, our work examines whether apparent strategic behavior is supported by the correct underlying behavioral mechanism. 
Instead of using successful play or opponent identification as the endpoint, we decompose behavior into marginal distribution matching, strategy recognition, and conditional rule execution, allowing us to test whether models recover the sequential rule behind the observed trajectory.
% \paragraph{LLM Agents and Interactive Game-Based Evaluation}
% Recent work has evaluated LLMs as interactive agents in multi-step environments, including general benchmarks for tool use, web navigation, and long-horizon task completion \citep{liu2024agentbench, zhou2024webarenarealisticwebenvironment}. 
% Another line of work uses games as controlled testbeds for evaluating strategic reasoning, hidden-state inference, and multi-agent interaction \citep{costarelli2024gamebenchevaluatingstrategicreasoning, wang2025tmgbench, light2023avalonbenchevaluatingllmsplaying, wu-etal-2024-deciphering}. 
% While these studies show that interactive environments are useful for evaluating LLM agency and strategic behavior, they mainly focus on task success or game-level performance. 
% In contrast, our work asks whether LLMs can recover the latent process behind observed behavior, separating fixed marginal action distributions from Markovian, history-dependent strategies in repeated Rock--Paper--Scissors, an easy-to-understand game structure.\jack{how this differs from the focus on task success? provides a deeper understanding?}
% \jerry{yes, overall for this part I DO THINK it is kinda sparse cuz the most relevent papers are cited at intro, so these are not so relevant}

\paragraph{In-Context Learning of Sequential Structure}
Large language models can perform new tasks from examples provided in context, making in-context learning a central explanation for their few-shot behavior \citep{NEURIPS2020_1457c0d6}. 
However, prior work suggests that ICL does not necessarily amount to learning an explicit task rule. 
It has been interpreted as implicit Bayesian inference \citep{xie2022explanationincontextlearningimplicit}, shown to depend on training-data distributional properties \citep{10.5555/3600270.3601641}, and found to benefit from demonstrations even when labels are corrupted \citep{min-etal-2022-rethinking}. 
Recent controlled studies further show that LLMs can approximate Bayesian updating in stochastic settings given sufficient in-context evidence \citep{gupta-etal-2025-enough}.
Mechanistic and theoretical studies further connect ICL to induction heads, task retrieval, and label-relationship learning rather than conventional rule learning \citep{olsson2022incontextlearninginductionheads, pmlr-v235-lin24l, kossen2024incontext}. 
These findings motivate a more fine-grained evaluation of what structure models extract from context. 
Our setting extends this question beyond marginal belief updating by testing not only whether LLMs can use previous examples, but whether they can infer and sustain the sequential dependency that generated an interaction trajectory.

\paragraph{Distinguishing Marginal Statistics from Markovian Dynamics}
Classical sequence modeling distinguishes surface-level observation frequencies from latent transition dynamics. 
Markov and hidden Markov models explicitly represent how future observations depend on previous states or histories, rather than only on aggregate token or action counts \citep{70a60b27-b856-312d-8021-d928a796d626, 10.5555/108235.108253}. 
This distinction is central to our setting, where two Rock--Paper--Scissors players may have similar marginal action frequencies but follow different conditional rules. 
Empirical studies of repeated Rock--Paper--Scissors similarly show that human play can exhibit conditional response patterns rather than pure randomization \citep{Wang_2014}. 
Related work in opponent modeling also studies how agents infer other agents' behavior, goals, or strategies from observations \citep{pmlr-v48-he16, ALBRECHT201866, pmlr-v80-raileanu18a}. 
However, these works primarily aim to improve adaptation or decision-making in multi-agent environments. 
Our work instead directly evaluates whether LLMs recover sequential structure beyond fixed marginal behavior, testing whether apparent behavioral understanding reflects conditional rule recovery rather than frequency matching.

\definecolor{pyrblue}{RGB}{222,237,250}
\definecolor{pyrgreen}{RGB}{232,245,234}
\definecolor{pyrorange}{RGB}{250,232,211}
\definecolor{pyrpurple}{RGB}{236,232,246}
\definecolor{linegray}{RGB}{120,120,120}

\begin{figure}[t]
\centering
\resizebox{1.00\columnwidth}{!}{%
\begin{tikzpicture}[
    font=\small,
    >={Latex[length=1.7mm]},
    level/.style={
        draw=linegray!55,
        line width=0.60pt,
        rounded corners=2mm,
        align=center,
        inner xsep=6pt,
        inner ysep=4pt
    },
    trans/.style={
        draw=linegray!50,
        fill=white,
        line width=0.45pt,
        rounded corners=1.5mm,
        align=center,
        inner xsep=5pt,
        inner ysep=2.5pt,
        font=\scriptsize
    },
    arrow/.style={
        ->,
        draw=linegray!75,
        line width=0.65pt
    },
    icon/.style={
        font=\normalsize,
        text=black!65
    }
]

% ======================================================
% Left axis label
% ======================================================
\node[
    rotate=90,
    font=\scriptsize\bfseries,
    text=black!75
] at (-4.55,0.00) {Increasing conditional complexity};

\draw[arrow] (-4.20,-2.60) -- (-4.20,2.60);

% ======================================================
% Level 1
% ======================================================
\node[
    level,
    fill=pyrblue,
    minimum width=7.60cm,
    minimum height=0.88cm
] (l1) at (0,-2.35) {};

\node[icon] at (-3.45,-2.35) {\faEye};

\node[font=\scriptsize\bfseries, align=center] at (0.22,-2.22) {
    Exp. 1: Strategy Identification
};
\node[font=\scriptsize, align=center] at (0.22,-2.51) {
    Infer latent strategies from trajectories.
};

% Transition 1
\node[trans] (t12) at (0,-1.50) {
    identify $\rightarrow$ identify + generate
};

% ======================================================
% Level 2
% ======================================================
\node[
    level,
    fill=pyrgreen,
    minimum width=6.95cm,
    minimum height=0.88cm
] (l2) at (0,-0.70) {};

\node[icon] at (-3.10,-0.70) {\faFilm};

\node[font=\scriptsize\bfseries, align=center] at (0.25,-0.57) {
    Exp. 2: Simple Markov Inference-and-Simulation
};
\node[font=\scriptsize, align=center] at (0.25,-0.86) {
    Generate under first-order rules.
};

% Transition 2
\node[trans] (t23) at (0,0.12) {
    first-order $\rightarrow$ joint / second-order
};

% ======================================================
% Level 3
% ======================================================
\node[
    level,
    fill=pyrorange,
    minimum width=6.85cm,
    minimum height=0.88cm
] (l3) at (0,0.95) {};

\node[icon] at (-3.02,0.95) {\faPuzzlePiece};

\node[font=\scriptsize\bfseries, align=center] at (0.30,1.08) {
    Exp. 3: Complex Markov Inference-and-Simulation
};
\node[font=\scriptsize, align=center] at (0.30,0.79) {
    Joint-state and second-order rules.
};

% Transition 3
\node[trans] (t34) at (0,1.78) {
    remove attribution + extend order
};

% ======================================================
% Level 4
% ======================================================
\node[
    level,
    fill=pyrpurple,
    minimum width=5.15cm,
    minimum height=0.88cm
] (l4) at (0,2.60) {};

\node[icon] at (-2.20,2.60) {\faLayerGroup};

\node[font=\scriptsize\bfseries, align=center] at (0.22,2.73) {
    One-Player Continuation
};
\node[font=\scriptsize, align=center] at (0.22,2.44) {
    Up to 8th-order recovery.
};

% ======================================================
% Connecting arrows
% ======================================================
\draw[arrow] (l1.north) -- (t12.south);
\draw[arrow] (t12.north) -- (l2.south);

\draw[arrow] (l2.north) -- (t23.south);
\draw[arrow] (t23.north) -- (l3.south);

\draw[arrow] (l3.north) -- (t34.south);
\draw[arrow] (t34.north) -- (l4.south);

\end{tikzpicture}%
}
\caption{Ability-based evaluation pyramid, from strategy identification to increasingly complex conditional generation.}
\label{fig:methodology}
\end{figure}
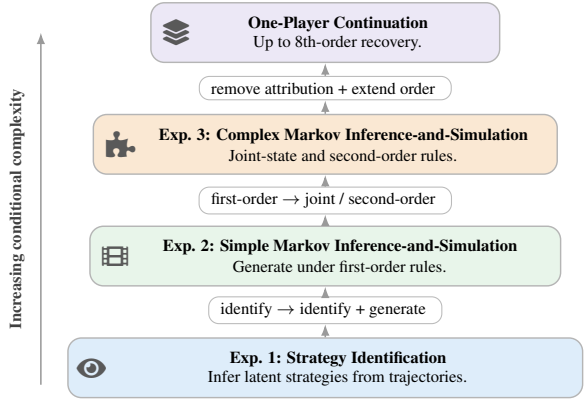
\section{Methodology}

\subsection{Task Description}

We study whether LLMs can infer and reproduce latent strategies from interaction trajectories. The task is instantiated as a two-player Rock--Paper--Scissors game, where each player selects an action from $\mathcal{A}=\{R,P,S\}$ at each round; this provides a minimal interactive setting in which opponent responses can induce history-dependent behavior while the small action space enables controlled rule-level evaluation.
A trajectory of length $T$ is denoted as
\begin{equation}
\label{eq:trajectory}
\tau_{1:T}=\{(a^1_t,a^2_t)\}_{t=1}^{T},
\end{equation}
where $a^i_t\in\mathcal{A}$ is the action of player $i\in\{1,2\}$ at round $t$.

Each player is governed by an unobserved strategy $\pi_i\in\Pi$. Given an observed trajectory $\tau_{1:T}$, the model is asked to infer the latent strategies $(\pi_1,\pi_2)$, estimate action distributions $\hat{\mathbf{p}}_i\in\Delta(\mathcal{A})$ when applicable, and, in generation settings, produce a continuation $\hat{\tau}_{T+1:T+M}=\{(\hat{a}^1_t,\hat{a}^2_t)\}_{t=T+1}^{T+M}$ that remains consistent with the inferred strategies.
\subsection{Player Types}

\paragraph{Statistical Players.}
Statistical players do not condition on previous rounds. This group includes static players, which always choose a fixed action $a^\star\in\mathcal{A}$, and distributional players, which sample independently from one of 16 fixed categorical distributions $\mathbf{p}\in\Delta(\mathcal{A})$ over Rock, Paper, and Scissors (see Table~\ref{tab:candidate_pool_overview} in Appendix~\ref{app:candidate_pool_overview} for details). These players test whether models can recover marginal action statistics.

\paragraph{Markov Players.}
Markov players choose actions according to deterministic rules over recent history, isolating history-conditioned strategy recovery from stochastic transition estimation. We consider policies of the form $a^i_t = f_i(h_t)$, where $h_t$ may include the opponent's previous action $a^{-i}_{t-1}$, the previous joint state $(a^1_{t-1},a^2_{t-1})$, or second-order histories such as $\tau_{t-2:t-1}$. These players test whether models can infer and follow sequential dependencies beyond marginal frequencies.

\subsection{Experiment Types and Scale}

Figure~\ref{fig:methodology} summarizes our evaluation design as a progression from strategy identification to increasingly complex conditional generation. 
Experiment 1 evaluates seven model configurations, Experiment 2 uses four core configurations, and Experiment 3 uses deepseek-reasoner for controlled structural analysis. 
Full sample counts and generation settings are reported in Appendix~\ref{app:experiment_scale}.

\paragraph{Exp. 1: Strategy Identification.}
We test whether models can recover latent player strategies from interaction trajectories. Given $\tau_{1:T}$, the model predicts $(\hat{\pi}_1,\hat{\pi}_2)$ and, for statistical players, estimates $(\hat{\mathbf{p}}_1,\hat{\mathbf{p}}_2)$. We use $T\in\{100,200,500,1000\}$, with 50 statistical-vs-statistical pairs, 50 Markov-as-Player-1 pairs, and 50 Markov-as-Player-2 pairs per context length.

\paragraph{Exp. 2: Simple Markov Inference-and-Simulation.}
We test whether models can infer latent strategies and simulate future trajectories without player identities. Given $\tau_{1:T}$, the model predicts $(\hat{\pi}_1,\hat{\pi}_2)$ and generates $\hat{\tau}_{T+1:T+M}$. Markov players use first-order opponent-only rules. We use 60 samples per pair type and generate 1000 rounds.

\paragraph{Exp. 3: Complex Markov Inference-and-Simulation.}
We extend the same inference-and-simulation framework to more complex Markov rules, where $h_t$ may include joint states or second-order histories. We again use 60 samples for each pair type and generate 1000-round continuations. This setting tests whether models can sustain more demanding history-dependent strategies rather than merely matching marginal action frequencies.

\subsection{Models and Experimental Coverage}

We evaluate both open-weight and proprietary LLMs. Our core model set includes DeepSeek-V4-Flash \cite{deepseekai2026deepseekv4}, evaluated via \texttt{deepseek-chat} and \texttt{deepseek-reasoner}, together with GPT-5 and GPT-5-mini \cite{singh2026openaigpt5card}. For Experiment 1, we additionally evaluate GPT-4.1 \cite{openai2025gpt41}, Gemini 3 Flash \cite{googledeepmind2025gemini3flash}, and Qwen3-8B \cite{yang2025qwen3technicalreport} to broaden coverage across providers, model generations, scales, and open- versus closed-weight settings. All models use identical task instructions; full prompts are provided in Appendix~\ref{app:prompts}, and detailed model descriptions and API-specific information are provided in Appendix~\ref{app:models}.
% All models use identical task instructions; full prompts are provided in Appendix~\ref{app:prompts}.

% \paragraph{Prompting} \jack{Jerry: describe the high-level details of the prompts and then refer the reader to the Appendix~\ref{app:prompts} where you put the prompts. Also mention the decoding strategies used (such as temperature). Default configuration?}

\subsection{Evaluation Metrics} 
% \jack{The evaluation metrics can be tightened up. You should go right in how each metric is computed, no need to say things like ``For strategy identification, the model predicts player identities $(\hat{\pi}_1,\hat{\pi}_2)$ for the true pair $(\pi_1,\pi_2)$. We report exact-match identification accuracy, which requires both predicted identities to match the true pair.'' Simply say we report exact-match between model predicted $(\hat{\pi}_1,\hat{\pi}_2)$ and ground truth $(\pi_1,\pi_2)$. Try to use LLM to make this section more concise. Remember to provide the entire methodology section while asking LLM to shorten this section.}
\label{sec:evaluation_metrics}

% We define only the metrics used in the main results. Appendix-only diagnostics are omitted from this section.

\paragraph{Identification.}
We report exact-match accuracy between the predicted strategy pair $(\hat{\pi}_1,\hat{\pi}_2)$ and the ground-truth pair $(\pi_1,\pi_2)$. Results are reported separately for Markov and non-Markov players, and their difference quantifies whether history-dependent strategies are harder to identify.

\paragraph{Generation.} We evaluate generated continuations $\hat{\tau}_{T+1:T+M}$ in windows $W_1,\ldots,W_K$. For Markov players, we measure rule following by comparing each generated action $\hat{a}^i_t$ with the action prescribed by the true rule $f_i(\hat{h}_t)$ under the generated history:
\begin{equation}
\label{eq:rule_overlap}
\mathrm{Overlap}_{i,k}
=
\frac{1}{|E_{i,k}|}
\sum_{t\in E_{i,k}}
\mathbb{I}\big[\hat{a}^i_t=f_i(\hat{h}_t)\big],
\end{equation}
where $E_{i,k}$ denotes rule-evaluable rounds in window $W_k$. We also report strict rule match, which requires all evaluable actions in a window to satisfy the rule, and cumulative strict match, which measures whether rule following persists over time.

For statistical players, we compare generated and target action distributions using MSE. We also report the relative Markov--Non-Markov MSE gap:
\begin{equation}
\label{eq:mse_delta}
\Delta_{\mathrm{MSE}}(\%)
=
\frac{
\mathrm{MSE}_{\mathrm{Markov}}
-
\mathrm{MSE}_{\mathrm{NonMarkov}}
}{
\mathrm{MSE}_{\mathrm{NonMarkov}}
}
\times 100,
\end{equation}
where positive values indicate higher simulation error for Markov players.

\paragraph{High-Order Generation.}
\label{met:high-o-gen}
For the one-player $n$-gram follow-up experiment, exact match is inappropriate because the target process is stochastic. We therefore use context-conditioned likelihood gain (CCLG), which measures improvement over a unigram baseline, and state-frequency-weighted Jensen--Shannon divergence:
\begin{equation}
\label{eq:wjs}
\mathrm{WJS}
=
\sum_{s} w_s\,
\mathrm{JS}\big(q(\cdot\mid s),\hat{q}(\cdot\mid s)\big),
\end{equation}
where $s$ is an $n$-gram history state, $q(\cdot\mid s)$ is the true next-action distribution, $\hat{q}(\cdot\mid s)$ is the empirical distribution from the generated continuation, and $w_s$ is the empirical state frequency. Higher CCLG and lower WJS indicate better recovery of history-dependent structure.

\section{Results}\label{sec:results}

% Dimensions

% \begin{itemize}
%     \item Experiments dimension 
%     \item Model dimension
%     \item Sequence length
%     \item Metrics dimension 
%     \item Player dimension
% \end{itemize}
% Figure~\ref{fig:methodology} summarizes our evaluation as an ability-based progression from strategy identification to increasingly complex conditional generation. 
We first test whether models can identify statistical and Markovian players from observed trajectories, then examine whether their generated continuations preserve the inferred behavioral process, and finally stress-test conditional generation under more complex Markov structures.
Overall, these experiments evaluate three abilities that can be conflated in surface behavior: \emph{identifying latent strategies, matching marginal action distributions, and executing conditional rules during generation}.
% The results reveal a consistent separation between marginal distribution matching, strategy recognition, and conditional rule execution. 
% Longer context does not reliably improve latent strategy identification, correct recognition does not guarantee faithful simulation, and higher-order dependencies make rule execution substantially harder even when the relevant behavioral structure is recoverable from the trajectory.
\subsection{Can models identify latent player strategies?}
\label{sec:strategy_identification}

\begin{figure}[t]
    \centering
    \includegraphics[width=\linewidth]{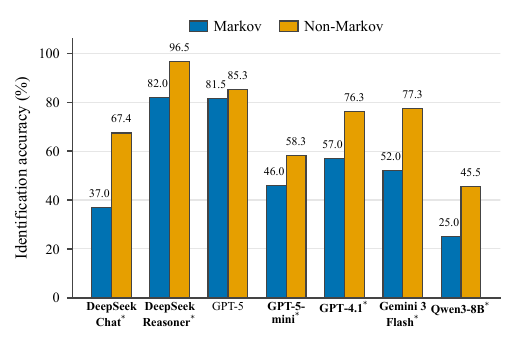}
    \caption{
Identification accuracy across seven model configurations, averaged across context lengths.
Bold model labels and asterisks indicate significant Markov--Non-Markov differences ($p<.001$).
}
    \label{fig:identification_model_coverage}
\end{figure}
Figure~\ref{fig:identification_model_coverage} compares strategy-identification performance across seven model configurations.
Markov identification accuracy is lower on average than Non-Markov accuracy for all seven models, and this difference is significant in six of the seven cases ($p<.001$), with GPT-5 as the sole exception ($p=.095$).
The pattern is therefore highly consistent, with Markov strategies being broadly more difficult to recover than Non-Markov strategies across providers, model families, scales, and both open- and closed-weight settings.

\paragraph{Longer context does not help and often hurts identification.}
The results in Appendix~\ref{app:exp1_full_results} \textbf{to figure}
show that increasing the interaction history from 100 to 1000 rounds
does not improve identification and often degrades it.
All seven models decline on Non-Markov players, and nearly all
decline on Markov players, despite some non-monotonic variation.
More trajectory evidence is therefore insufficient for latent-strategy
recovery and can even make identification harder, despite remaining
within the models' effective context windows.

Appendix~\ref{app:mle_identification_baseline} supports this interpretation with a non-LLM maximum-likelihood identification baseline, which achieves 100\% accuracy across all context lengths under the same closed candidate pool.
This indicates that the observed LLM failures are not due to insufficient recoverable evidence in the trajectories.
For the core models, Appendix~\ref{app:ols_length_analysis} further shows that longer input length predicts lower accuracy, while output length does not reliably improve performance. A prompt-format ablation comparing raw trajectories with row-wise tables is provided in Appendix~\ref{app:structured_format_ablation}.
\paragraph{Takeaway.}
The limitation is not the availability of trajectory evidence, but whether models can recover the behavioral structure encoded in it.
Markov strategies are harder for many, though not all, models, and increasing the amount of interaction does not resolve this difficulty.
Since the same trajectories are perfectly identifiable by the maximum-likelihood baseline, these failures point to a limitation in extracting and maintaining latent decision rules rather than insufficient information.

\subsection{Can models faithfully simulate inferred strategies?}
\label{sec:simulation}
We then evaluate whether generated continuations remain faithful to the underlying behavioral process, rather than only matching surface action frequencies. 
We assess rule-following accuracy, distributional fidelity, and cumulative consistency over generated rounds. 
The Markov player pool consists of three first-order opponent-only rules. 
At each generated round $t$, a rule maps the opponent's previous generated action to a prescribed action $f_i(\hat{h}_t)$, which we compare with the model's generated action. 
Thus, rule-following is evaluated online using the generated history $\hat{h}_t$, so early mistakes can affect later prescribed actions.

\begin{figure}[t]
    \centering
    \includegraphics[width=\columnwidth]{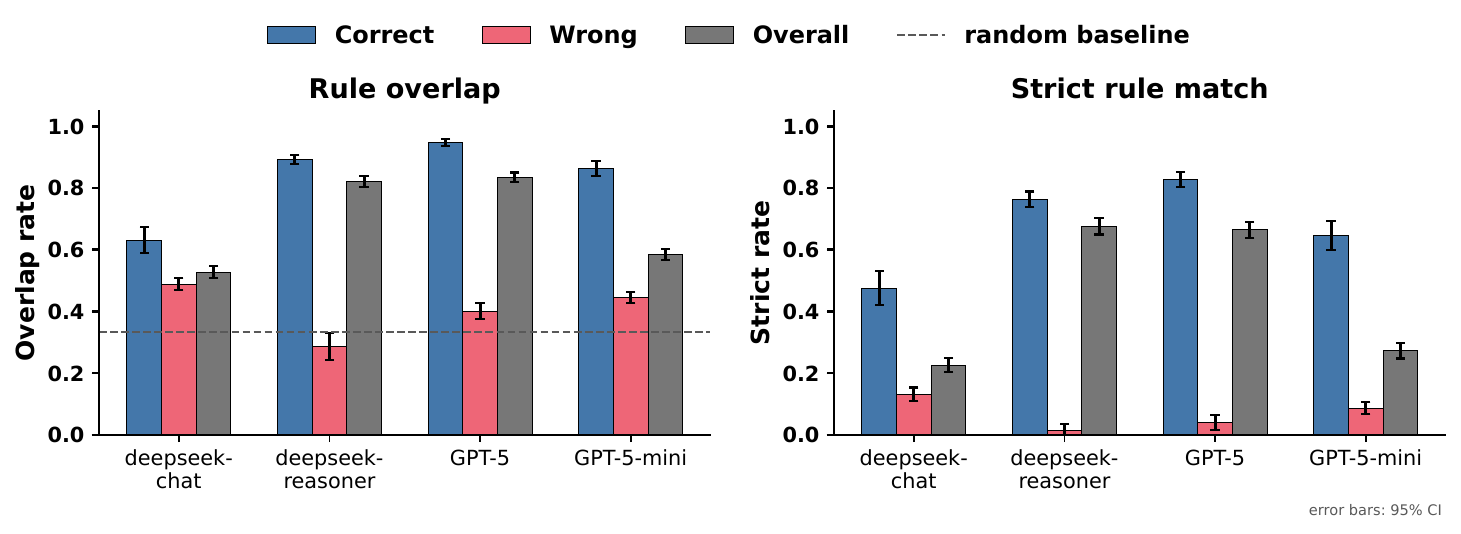}
    \caption{
    Rule-following performance by identity condition.
    }
    \label{fig:rule_following_summary}
\end{figure}

\paragraph{Correct identification improves rule-following, but wrong identities are not random.}
Figure~\ref{fig:rule_following_summary} shows rule overlap and strict rule match under correct, wrong, and overall identity conditions. Models consistently simulate the target rule better when they identify the player correctly, especially under strict rule match. This suggests that recognition is strongly related to simulation quality.

\begin{figure*}[t]
    \centering
    \begin{minipage}[t]{0.485\textwidth}
        \centering
        \includegraphics[width=\linewidth]{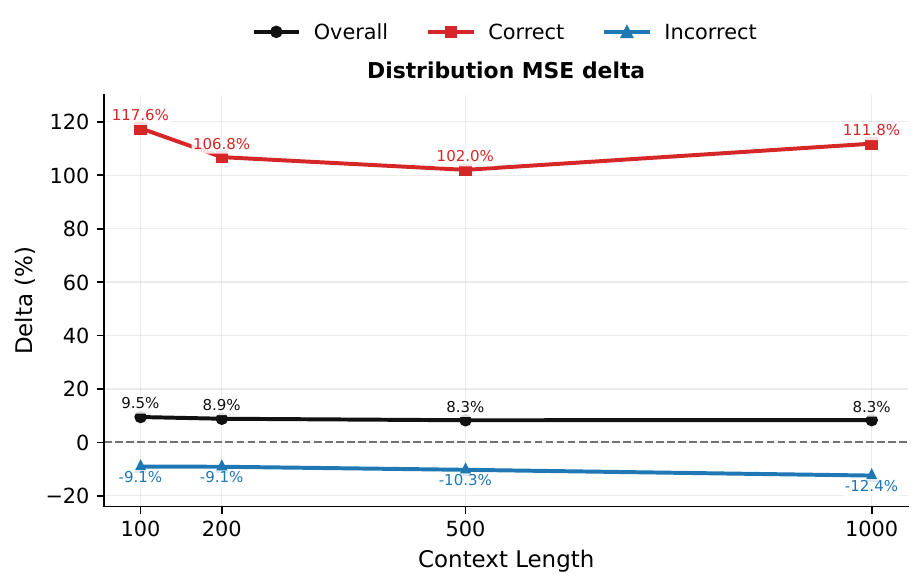}
        \caption{
        Relative MSE gap between Markov and non-Markov players. Positive values indicate higher Markovian structural error.
        }
        \label{fig:mse_delta}
    \end{minipage}
    \hfill
    \begin{minipage}[t]{0.485\textwidth}
        \centering
        \includegraphics[width=\linewidth]{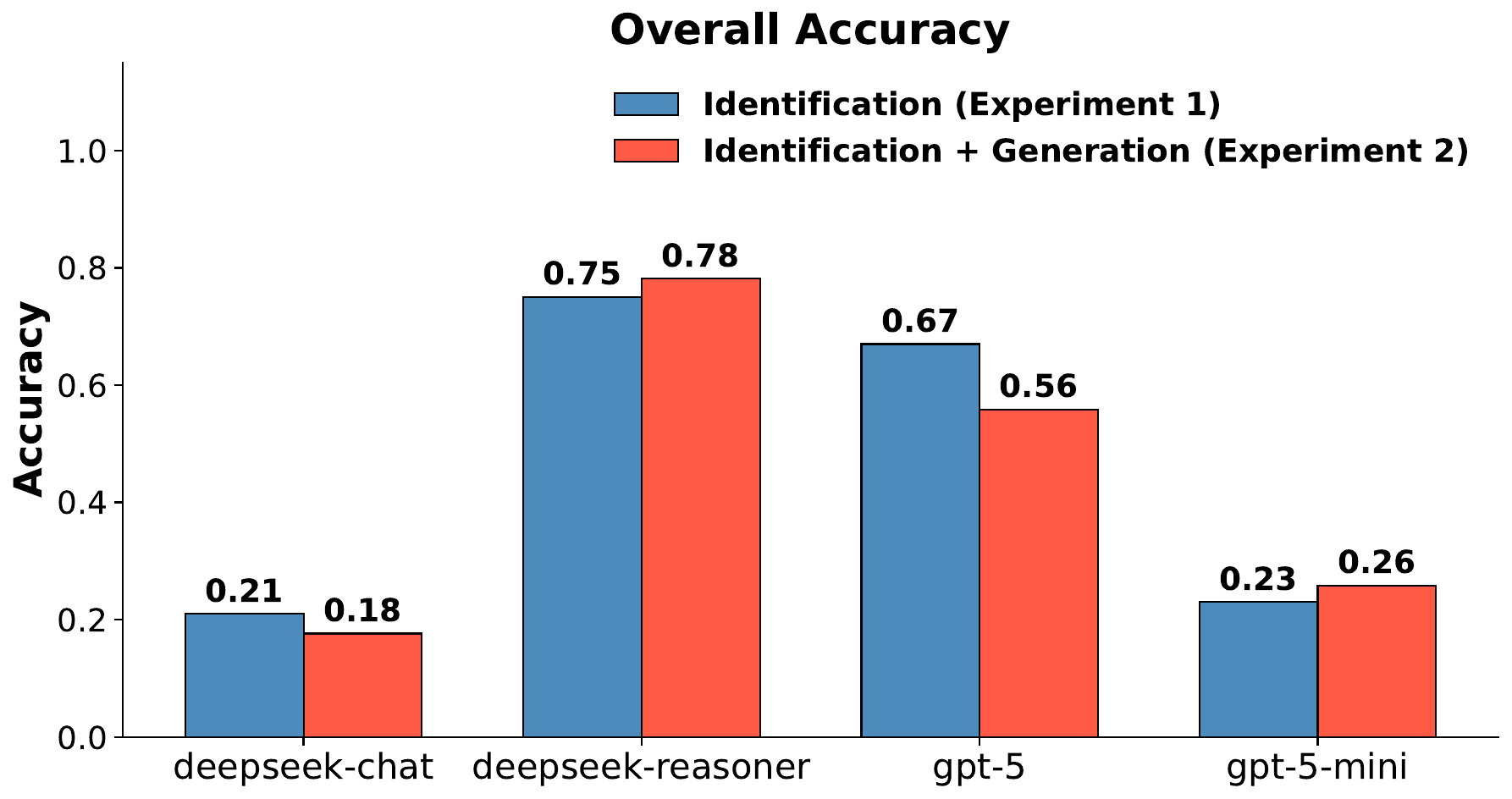}
        \caption{
        Recognition accuracy under recognition-only and recognition-with-generation settings.
        }
        \label{fig:recognition_vs_simulation}
    \end{minipage}
\end{figure*}

However, wrong-identity cases do not collapse to random behavior. In several models, rule overlap remains above the random baseline even when the predicted identity is incorrect. This suggests that identity errors are structured, since a model may select the wrong player while still choosing one with similar behavioral rules. Thus, exact identity recognition and rule-level simulation are related, but they should not be treated as the same target.
\paragraph{Models simulate statistical players better than Markov players.}
We next evaluate distributional simulation quality using the relative MSE gap between Markov and non-Markov players, as defined in Eq.~\ref{eq:mse_delta}. 
Figure~\ref{fig:mse_delta} shows that Markov players are consistently harder to simulate overall, with MSE about 8--9\% higher than non-Markov players. 
The gap becomes much larger in the correct-identity subset, where Markov MSE is more than 100\% higher across context lengths. 
In contrast, the incorrect-identity subset shows the opposite pattern, with wrong Markov predictions producing lower MSE than wrong non-Markov predictions.

This contrast suggests that distributional similarity should not be interpreted as faithful rule execution. 
Even when the model correctly identifies a Markov player, it may still fail to execute the corresponding state-dependent transition rule throughout generation. 
Conversely, incorrect strategy recovery can still produce deceptively accurate marginal behavior. 
Among 282 incorrectly recovered strategies, 41.8\% remain within TV $\leq 0.02$ of the target marginal distribution, and 62.1\% remain within TV $\leq 0.10$.
\textbf{Thus, surface-level distributional fidelity can frequently conceal an incorrect sequential mechanism, while identity recognition, marginal matching, and conditional rule execution capture related but distinct aspects of simulation quality.}
\begin{figure}[t]
\centering
\includegraphics[width=0.98\linewidth]{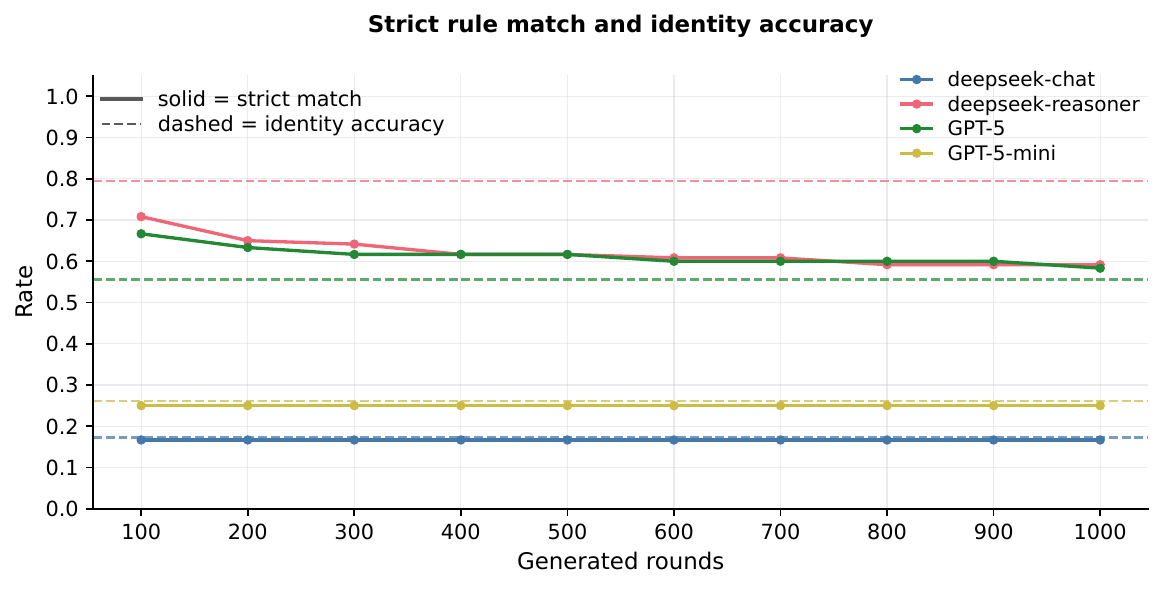}
\caption{
Cumulative strict rule match across generated rounds. Solid lines show cumulative strict rule match, while dashed lines show overall identity accuracy as reference values rather than upper bounds.
}
\label{fig:strict_identity_accuracy}
\end{figure}

\paragraph{Identity accuracy is not an upper bound on rule-level simulation.}
Figure~\ref{fig:strict_identity_accuracy} shows cumulative strict rule match across generated rounds, together with each model's overall identity accuracy. The dashed identity-accuracy lines are reference values rather than upper bounds because exact player identification and rule-level simulation are related but not perfectly coupled. In some cases, strict rule match meets or exceeds identity accuracy, showing that a model can generate actions consistent with the target rule even without naming the exact target player. This distinction matters because faithful rule execution may appear to require correct strategy identification, but our results show that rule-level simulation can still succeed without exact identity recovery.

The cumulative curves also show limited degradation over longer generations. Stronger models decline mildly in early rounds and then stabilize, while weaker models remain mostly flat. This suggests that rule-following quality is largely determined early in generation rather than progressively degrading over time.
% Additional diagnostics are reported in Appendices~\ref{app:distributional_diagnostics} and~\ref{app:identification_diagnostics}.
\paragraph{Teacher forcing separates local rule application from long-horizon execution.}
To diagnose whether the free-running errors above reflect immediate rule-application failures or errors accumulated over generation, we additionally evaluate next-step prediction under teacher forcing.
At each step, the model receives the ground-truth interaction history and predicts only the next action, preventing earlier generation errors from propagating.

\begin{table}[!ht]
\centering
\footnotesize
\setlength{\tabcolsep}{6pt}
\renewcommand{\arraystretch}{1.10}
\begin{tabular}{lcc}
\toprule
\textbf{Model}
& \textbf{All cases}
& \textbf{Correct-ID subset} \\
\midrule
DeepSeek Chat      & 52.5\% & 66.7\% \\
DeepSeek Reasoner  & 93.3\% & 97.2\% \\
GPT-5              & 82.5\% & 89.5\% \\
GPT-5-mini         & 54.2\% & 70.0\% \\
\bottomrule
\end{tabular}
\caption{
Teacher-forced next-step accuracy under ground-truth interaction histories.
\textit{All cases} reports accuracy over the full evaluation set, while
\textit{Correct-ID subset} includes only cases in which the player identity was correctly inferred.
}
\label{tab:teacher_forcing}
\end{table}

The results reveal two sources of simulation failure.
DeepSeek Reasoner and GPT-5 perform strongly under teacher forcing, suggesting that their remaining free-running errors partly reflect difficulty sustaining the rule over longer trajectories.
DeepSeek Chat and GPT-5-mini still struggle with ground-truth histories, indicating that local rule application itself remains a bottleneck.
Teacher forcing thus separates immediate rule-execution failures from failures to maintain the sequential mechanism during generation.
\subsection{Does generation affect recognition?}
\label{sec:complex_input_structures}

We compare strategy recognition in Exp1, where models only identify the players, with Exp2, where they must also generate future behavior. Overall accuracy requires both players to be identified correctly.

Figure~\ref{fig:recognition_vs_simulation} shows no systematic effect of generation on recognition. Some models improve while others decline, but none of the differences are significant under two-proportion z-tests (Appendix~\ref{app:recognition_simulation_significance}). This indicates that generation failures cannot be explained by player-identification errors alone.
\subsection{Which structures make rule execution harder?}
\label{sec:recognition_vs_simulation}
\begin{figure*}[t]
\centering
\begin{minipage}[t]{0.58\textwidth}
\centering
\includegraphics[width=\linewidth]{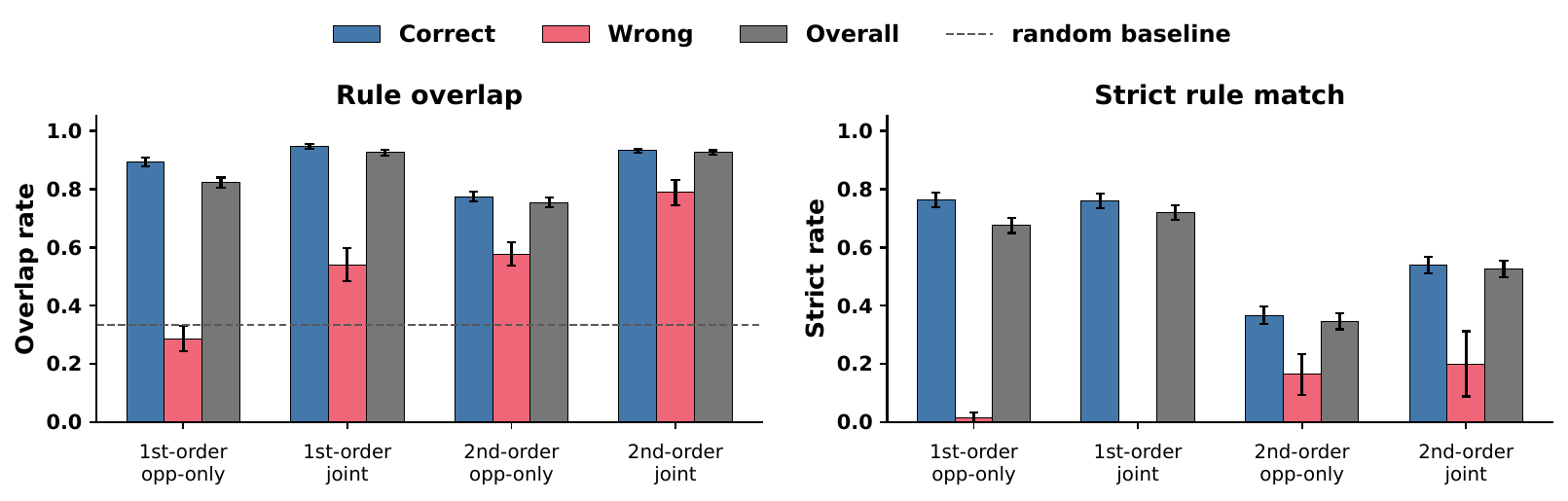}
\smallskip
\textbf{(a)} Rule-following performance.
\end{minipage}
\hspace{0.02\textwidth}
\begin{minipage}[t]{0.38\textwidth}
\centering
\includegraphics[width=\linewidth]{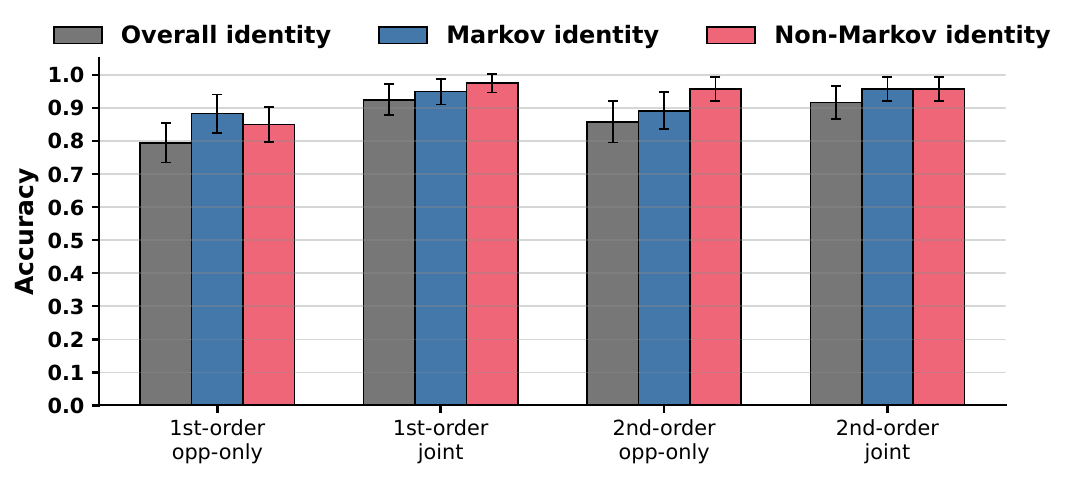}
\smallskip
\textbf{(b)} Identity identification accuracy.
\end{minipage}

\caption{
Results on complex Markov input structures. 
Panel (a) shows rule-following performance, and Panel (b) shows identity identification accuracy across the same settings.
}
\label{fig:complex_input_results}
\end{figure*}

The previous experiments show that conditional generation is the more challenging setting, so we use it as a stress test of which Markov structures make rule execution harder.
We focus on deepseek-reasoner because it achieved the strongest performance in the preceding experiments; the goal is to isolate structural factors rather than compare models.
We vary dependency order and joint-state conditioning to separate longer-history effects from conditioning on both players' previous actions.
Appendix~\ref{app:complex_markov_sets} provides the Markov set details.
\paragraph{Joint-state information is not the main bottleneck.}
Figure~\ref{fig:complex_input_results}a shows that adding joint-state information does not reduce rule-following performance.
The joint-state condition remains strong in both first- and second-order settings, suggesting that conditioning on both players' previous actions is not the primary source of degradation when the transition structure remains locally observable.
\paragraph{Second-order dependency drives the larger degradation.}
The main drop comes from second-order rules. 
In Figure~\ref{fig:complex_input_results}a, both second-order settings show lower strict rule match than the first-order settings, even when rule overlap remains relatively high. 
This suggests that the model can recover local rule elements but fails to compose them into a consistently correct second-order transition rule. 
Figure~\ref{fig:complex_input_results}b further shows that identity accuracy remains high despite the drop in strict rule following, indicating that the harder part is rule execution rather than Markovian player recognition.
These results point to a broader failure mode in higher-order conditional generation: the main bottleneck is maintaining and applying longer historical dependencies, not simply representing joint states.
% Appendices~\ref{app:complex_identification_diagnostics}--\ref{app:complex_identity_rule_diagnostics} further separate attribution errors, distributional drift, and identity-conditioned rule-execution errors.
\subsection{Summary}
Strategy recognition, marginal distribution matching, and conditional rule execution are related but separable abilities, as correct recognition does not guarantee simulation and generation does not systematically change recognition accuracy.
Longer-history dependency, rather than joint-state conditioning, is the bottleneck, especially for second-order rules.
Models handle first-order or locally observable rules, but degrade when generation requires maintaining higher-order conditional structure.
Surface behavioral similarity can mask failures to recover and execute sequential structure.

\section{Follow-Up Experiments}
\label{sec:analyses}

\subsection{One-Player Stochastic Processes}
\label{sec:one_player_ngram}
The preceding results suggest that the core issue is not only whether models can identify a player, but whether they can recover and use the sequential structure behind observed behavior. However, two-player generation conflates player attribution, rule inference, and rule execution, while the main Markov rules are deterministic. We therefore evaluate a one-player stochastic $n$-gram Markov continuation task with probabilistic transition rules. This tests whether models can recover conditional transition distributions rather than merely execute deterministic rules or match marginals. If models still struggle, the bottleneck lies beyond player attribution, pointing to difficulty in maintaining high-order conditional structure during generation.
\subsection{Factors}
\paragraph{Dependency length.}
We generate prefixes from ground-truth $n$-gram Markov processes over an abstract vocabulary $\{A,B,C\}$ to avoid Rock-Paper-Scissors label-semantic confounds. Each process follows a transition table $P(x_t \mid x_{t-n:t-1})$. We vary $n \in \{1,2,4,8\}$ to test degradation with dependency length, from local conditioning at low order to longer-history tracking at high order.

\paragraph{Rule access.}
We use two prompting settings. In the prefix-only setting, the model observes only the generated prefix and must infer the stochastic rule from examples. In the rule-given setting, the model is also given the n-gram rule and transition probabilities, isolating whether failures arise from rule inference or rule execution.
\paragraph{Evaluation.}
Because the process is stochastic, we use CCLG and WJS instead of exact-match accuracy. 
As defined in Section~\ref{met:high-o-gen}, they measure context use beyond unigram frequencies and recovery of the true conditional transition distribution.
\subsection {Key Results}
\begin{figure}[t]
    \centering
    \includegraphics[width=\linewidth]{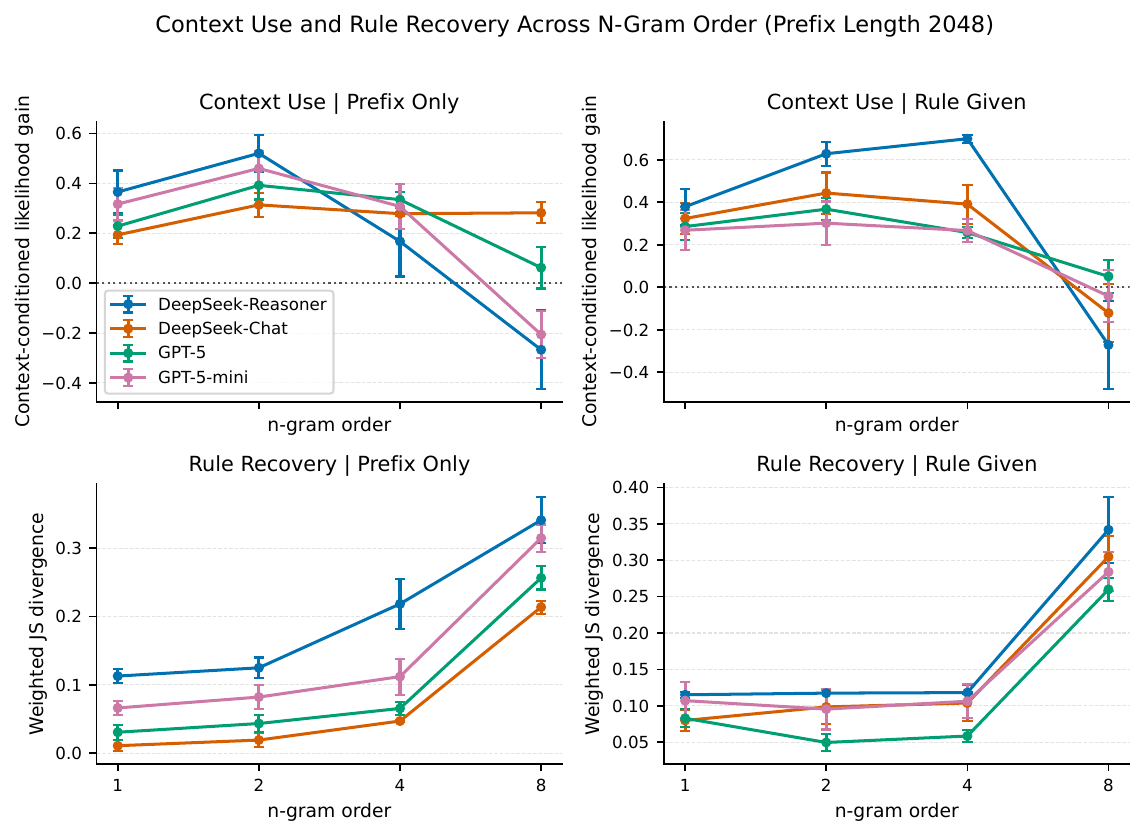}
    \caption{
    Context use and rule recovery across n-gram order with prefix length 2048. 
    }
    \label{fig:one_player_cclg_wjs}
\end{figure}

\paragraph{The task remains recoverable from the observed prefix.}
Appendix~\ref{app:one_player_no_seed_mle} reports a prefix-only n-gram MLE baseline under the same CCLG and WJS metrics.
It remains stable at higher orders using only the observed prefix, confirming that high-order conditional information remains recoverable.
\paragraph{High-order dependency breaks context use and rule recovery.}
Figure~\ref{fig:one_player_cclg_wjs} shows that CCLG is often positive for $n=1$ to $4$, indicating behavior beyond unigram frequency matching. At $n=8$, however, CCLG drops sharply for most models while WJS increases. This suggests that recurring low-order states remain exploitable, whereas sparse high-order states make conditional transition structure harder to recover and sustain during generation.
\paragraph{Stronger context use does not necessarily imply better rule recovery.}
Figure~\ref{fig:one_player_cclg_wjs} shows that higher CCLG does not always correspond to lower WJS. Some models achieve strong context-conditioned gains in lower-order or rule-given settings, yet still diverge from the true transition distributions. This suggests that context use and transition-rule recovery are related but distinct abilities. Appendices~\ref{app:one_player_likelihood_recovery}--\ref{app:one_player_rule_access_effects} further diagnose this gap by separating likelihood effects, output artifacts, and rule-access effects.

\paragraph{Prefix evidence and rule access do not fully explain the failure.}
Neither increasing the prefix length from 256 to 2048 nor providing transition rules consistently removes high-order degradation, suggesting that the bottleneck lies in maintaining transition structure during generation.
Appendix~\ref{app:one_player_prefix_effects} and ~\ref{app:one_player_information_effects} provide the corresponding analyses.
\paragraph{Summary.}
High-order generation failures persist even without player attribution, indicating that the core bottleneck lies in maintaining and applying conditional transition structure during generation.

\section{Conclusion}
\label{sec:conclusion}
We introduced a controlled framework for testing whether LLMs recover latent sequential dynamics beyond marginal action statistics, using two-player Rock--Paper--Scissors and a one-player stochastic $n$-gram task.
Our experiments separate latent strategy recognition, marginal distribution matching, and conditional rule execution over generated trajectories, abilities often conflated in agent evaluation.
Critically, higher-order transition recovery remains unreliable even when recoverable information is present and transition rules are explicitly provided.
Surface behavioral similarity or correct strategy recognition alone does not guarantee faithful conditional generation.
We hope this framework motivates future work on enhancing LLMs’ ability to understand underlying sequential structures.

\section*{Limitations}

% \paragraph{Limited interactive setting.}
% Our experiments use controlled Rock--Paper--Scissors trajectories to isolate marginal statistics from history-dependent dynamics.
% This design makes the latent generating process observable and allows precise evaluation of strategy recovery and rule execution.
% Within this controlled setting, our results show that LLMs can match surface-level behavior while failing to identify, maintain, and execute the underlying sequential strategy.
% In particular, longer context, correct identity recognition, and explicit rule access do not reliably eliminate failures in high-order dependency generation.
% A natural next step is to test whether the same failure modes appear in richer interactive settings that involve natural language observations, changing goals, partial observability, or non-stationary behavior.

\paragraph{Predefined strategy classes.}
Players are sampled from a predefined candidate set $\Pi$, which allows us to evaluate strategy recovery under known ground-truth rules. This setup is appropriate for controlled diagnosis, but it does not cover open-ended strategy discovery. Future work could extend the framework to larger or learned strategy spaces where the model must infer behavioral structure without a closed candidate pool.

\paragraph{Prompt and representation choices.}
Trajectories are represented symbolically, and outputs are evaluated through constrained formats and parsing rules. These choices reduce ambiguity and make rule-level evaluation reliable. Although we tested multiple prompt variants, we cannot guarantee that the chosen prompts, trajectory representations, output schemas, or decoding settings are optimal.

% \section*{Ethical Considerations}

% \paragraph{Synthetic data.}
% All experiments use synthetic players and generated trajectories. We do not collect human-subject data or infer attributes, intentions, or identities of real individuals.

\bibliography{custom}

\clearpage
% =========================
% Colored prompt box styles
% Put these in the preamble
% =========================
\tcbset{
  promptbase/.style={
    enhanced,
    breakable,
    boxrule=0.8pt,
    arc=2mm,
    left=2mm,
    right=2mm,
    top=1mm,
    bottom=1mm,
    fonttitle=\bfseries,
    coltitle=black,
    listing only,
    listing engine=listings,
    listing options={
      basicstyle=\ttfamily\footnotesize,
      breaklines=true,
      breakatwhitespace=false,
      columns=fullflexible,
      keepspaces=true,
      showstringspaces=false,
      showspaces=false,
      showtabs=false
    }
  },
  nonmarkovbox/.style={
    promptbase,
    colback=green!4,
    colframe=green!45!black,
    colbacktitle=green!18
  },
  markovbox/.style={
    promptbase,
    colback=orange!5,
    colframe=orange!70!black,
    colbacktitle=orange!22
  },
  generationbox/.style={
    promptbase,
    colback=blue!4,
    colframe=blue!55!black,
    colbacktitle=blue!15
  },
  identificationbox/.style={
    promptbase,
    colback=purple!4,
    colframe=purple!55!black,
    colbacktitle=purple!15
  }
}

% =========================
% Appendix A
% =========================
\clearpage
\onecolumn

\appendix
\section{Prompt Templates}
\label{app:prompts}

This appendix provides the prompt templates used in our Rock-Paper-Scissors experiments.
Each prompt consists of a candidate pool description and a task-specific instruction.
The candidate pool description includes non-Markovian distribution players and, depending
on the experiment setting, one Markovian player set.

% =========================
% A.1 Candidate Pool Overview
% =========================
\subsection{Candidate Pool Overview}
\label{app:candidate_pool_overview}

\begin{table}[htbp]
\centering
\small
\renewcommand{\arraystretch}{1.15}
\setlength{\tabcolsep}{6pt}

\begin{tabular}{c >{\raggedright\arraybackslash}p{3.2cm} c c c >{\raggedright\arraybackslash}p{5.2cm}}
\toprule
\rowcolor{blue!12}
\textbf{ID} & \textbf{Name} & \textbf{Rock} & \textbf{Paper} & \textbf{Scissors} & \textbf{Type / Rule} \\
\midrule

\rowcolor{green!7}
A & Pure Scissors & 0\% & 0\% & 100\% & Non-Markovian Static \\
\rowcolor{green!7}
B & Pure Rock & 100\% & 0\% & 0\% & Non-Markovian Static \\
\rowcolor{green!7}
C & Pure Paper & 0\% & 100\% & 0\% & Non-Markovian Static \\
\rowcolor{green!7}
D & Uniform Random & 33.3\% & 33.3\% & 33.4\% & Non-Markovian Distribution \\
\rowcolor{green!7}
E & Rock + Paper & 50\% & 50\% & 0\% & Non-Markovian Distribution \\
\rowcolor{green!7}
F & Rock + Scissors & 50\% & 0\% & 50\% & Non-Markovian Distribution \\
\rowcolor{green!7}
G & Paper + Scissors & 0\% & 50\% & 50\% & Non-Markovian Distribution \\
\rowcolor{green!7}
H & Rock Biased & 50\% & 25\% & 25\% & Non-Markovian Distribution \\
\rowcolor{green!7}
I & Paper Biased & 25\% & 50\% & 25\% & Non-Markovian Distribution \\
\rowcolor{green!7}
J & Scissors Biased & 25\% & 25\% & 50\% & Non-Markovian Distribution \\
\rowcolor{green!7}
K & Rock $>$ Paper & 50\% & 33.3\% & 16.7\% & Non-Markovian Distribution \\
\rowcolor{green!7}
L & Rock $>$ Scissors & 50\% & 16.7\% & 33.3\% & Non-Markovian Distribution \\
\rowcolor{green!7}
M & Paper $>$ Rock & 33.3\% & 50\% & 16.7\% & Non-Markovian Distribution \\
\rowcolor{green!7}
N & Paper $>$ Scissors & 16.7\% & 50\% & 33.3\% & Non-Markovian Distribution \\
\rowcolor{green!7}
O & Scissors $>$ Rock & 33.3\% & 16.7\% & 50\% & Non-Markovian Distribution \\
\rowcolor{green!7}
P & Scissors $>$ Paper & 16.7\% & 33.3\% & 50\% & Non-Markovian Distribution \\

\midrule

\rowcolor{orange!12}
X & Win-Last & -- & -- & -- & First-order opponent-only Markov \\
\rowcolor{orange!12}
Y & Lose-Last & -- & -- & -- & First-order opponent-only Markov \\
\rowcolor{orange!12}
Z & Copy-Last & -- & -- & -- & First-order opponent-only Markov \\

\rowcolor{orange!12}
Q & Missing-Action & -- & -- & -- & First-order joint-state Markov \\
\rowcolor{orange!12}
R & Beat-Missing & -- & -- & -- & First-order joint-state Markov \\
\rowcolor{orange!12}
S & Lose-Missing & -- & -- & -- & First-order joint-state Markov \\

\rowcolor{orange!12}
T & 2R-Missing & -- & -- & -- & Second-order joint-state Markov \\
\rowcolor{orange!12}
U & 2R-Beat-Missing & -- & -- & -- & Second-order joint-state Markov \\
\rowcolor{orange!12}
V & 2R-Lose-Missing & -- & -- & -- & Second-order joint-state Markov \\

\rowcolor{orange!12}
x & 2R-Opp-Missing & -- & -- & -- & Second-order opponent-only Markov \\
\rowcolor{orange!12}
y & 2R-Opp-Beat-Missing & -- & -- & -- & Second-order opponent-only Markov \\
\rowcolor{orange!12}
z & 2R-Opp-Lose-Missing & -- & -- & -- & Second-order opponent-only Markov \\

\bottomrule
\end{tabular}

\caption{
Candidate pool of player strategies used in the Rock-Paper-Scissors experiments.
Green rows denote non-Markovian players, including deterministic static players and
distribution-based players.
Orange rows denote Markovian or reactive players that condition their actions on
previous game states.
}
\label{tab:candidate_pool_overview}
\end{table}

% =========================
% A.2 Candidate Pool Description
% =========================
\subsection{Candidate Pool Description}
\label{app:non_markovian_players}

This candidate pool description serves as a shared \textbf{knowledge base} that is
injected into all task-specific prompts. Specifically, it appears as the
\texttt{\{knowledge\_base\}} variable at the beginning of both the
\textbf{Generation Prompt} (Appendix~\ref{app:generation_prompt}) and the
\textbf{Understanding/Identification Prompt} (Appendix~\ref{app:identification_prompt}).
Depending on the experiment setting, one of the Markovian player sets described in
Appendix~\ref{app:markovian_players} is appended to the non-Markovian distribution
player descriptions below before being inserted into the prompt.

\begin{tcblisting}{nonmarkovbox, title={Non-Markovian Distribution Players}}
# Rock-Paper-Scissors Player Behavior Knowledge Base

## Distribution Strategy Players
These players choose actions according to a fixed probability distribution, independently each round.

A - Pure Scissors: Always plays Scissors (0% Rock, 0% Paper, 100% Scissors)
B - Pure Rock: Always plays Rock (100% Rock, 0% Paper, 0% Scissors)
C - Pure Paper: Always plays Paper (0% Rock, 100% Paper, 0% Scissors)
D - Uniform Random: Equal probability for all actions (33.3% Rock, 33.3% Paper, 33.4% Scissors)
E - Rock + Paper: Only plays Rock or Paper (50% Rock, 50% Paper, 0% Scissors)
F - Rock + Scissors: Only plays Rock or Scissors (50% Rock, 0% Paper, 50% Scissors)
G - Paper + Scissors: Only plays Paper or Scissors (0% Rock, 50% Paper, 50% Scissors)
H - Rock Biased: Prefers Rock (50% Rock, 25% Paper, 25% Scissors)
I - Paper Biased: Prefers Paper (25% Rock, 50% Paper, 25% Scissors)
J - Scissors Biased: Prefers Scissors (25% Rock, 25% Paper, 50% Scissors)
K - Rock > Paper: Strong Rock preference (50% Rock, 33.3% Paper, 16.7% Scissors)
L - Rock > Scissors: Strong Rock preference (50% Rock, 16.7% Paper, 33.3% Scissors)
M - Paper > Rock: Strong Paper preference (33.3% Rock, 50% Paper, 16.7% Scissors)
N - Paper > Scissors: Strong Paper preference (16.7% Rock, 50% Paper, 33.3% Scissors)
O - Scissors > Rock: Strong Scissors preference (33.3% Rock, 16.7% Paper, 50% Scissors)
P - Scissors > Paper: Strong Scissors preference (16.7% Rock, 33.3% Paper, 50% Scissors)
\end{tcblisting}

% =========================
% A.3 Markovian Player Sets
% =========================
\subsection{Markovian Player Sets}
\label{app:markovian_players}

Depending on the experiment setting, one of the following Markovian player sets is
appended after the non-Markovian candidate pool description.

\subsubsection{First-Order Opponent-Only Markov Players}

\begin{tcblisting}{markovbox, title={XYZ Markovian Player Set}}
## Markov/Reactive Players (First-Order, Opponent-Only)
These players react to opponent's PREVIOUS move only:

X - Win-Last: Plays the move that would have beaten opponent's last move.
- First move: Random
- Round t: plays beat(opponent_{t-1})

Y - Lose-Last: Plays the move that would have lost to opponent's last move.
- First move: Random
- Round t: plays lose(opponent_{t-1})

Z - Copy-Last: Copies opponent's last move.
- First move: Random
- Round t: plays opponent_{t-1}

Valid player identities: A B C D E F G H I J K L M N O P X Y Z
\end{tcblisting}

\subsubsection{First-Order Joint-State Markov Players}

\begin{tcblisting}{markovbox, title={QRS Markovian Player Set}}
## Joint-State Markov Players (First-Order, Both Players)
These players react to BOTH players' previous moves.
State: (self_prev, opp_prev). Find the action NEITHER player played last round.

Q - Missing-Action: Plays the missing action directly.
R - Beat-Missing: Plays the action that BEATS the missing action.
S - Lose-Missing: Plays the action that LOSES TO the missing action.

Valid player identities: A B C D E F G H I J K L M N O P Q R S
\end{tcblisting}

\subsubsection{Second-Order Joint-State Markov Players}

\begin{tcblisting}{markovbox, title={TUV Markovian Player Set}}
## Second-Order Joint-State Markov Players
These players look at the last TWO rounds for BOTH players, 4 moves total.
Find the action(s) NOT present among those 4 moves.

T - 2R-Missing: Plays the missing action directly.
U - 2R-Beat-Missing: Plays the action that BEATS the missing action.
V - 2R-Lose-Missing: Plays the action that LOSES TO the missing action.

Valid player identities: A B C D E F G H I J K L M N O P T U V
\end{tcblisting}

\subsubsection{Second-Order Opponent-Only Markov Players}

\begin{tcblisting}{markovbox, title={xyz\_opp Markovian Player Set}}
## Second-Order Opponent-Only Markov Players
These players look at the opponent's last TWO moves only.
Find the action NOT present among those 2 opponent moves.

x - 2R-Opp-Missing: Plays the missing action directly.
y - 2R-Opp-Beat-Missing: Plays the action that BEATS the missing action.
z - 2R-Opp-Lose-Missing: Plays the action that LOSES TO the missing action.

Valid player identities: D E F G H I J K L M N O P x y z
\end{tcblisting}

% =========================
% A.4 Generation Prompt
% =========================
\subsection{Generation Prompt}
\label{app:generation_prompt}

\begin{tcblisting}{generationbox, title={Generation Prompt}}
{knowledge_base}

## Rock-Paper-Scissors Trajectory Simulation Task

You are observing a Rock-Paper-Scissors game between two **anonymous** players (P1 and P2).

**Player pool rules:**
- Both players are drawn from the strategy pool described above.
- At most ONE of them is {markov_desc}.
- It is also possible that BOTH are distribution players (neither is Markov).
- You do NOT know which specific strategy each player uses -- you must infer it from the trajectory.

### Previous Game History ({context_rounds} rounds):
Round 1: P1=<move>, P2=<move>
Round 2: P1=<move>, P2=<move>
...

---

### Your Task
Simulate the next **{simulate_rounds} rounds** (rounds {context_rounds + 1} to {context_rounds + simulate_rounds}).

**MANDATORY: You MUST generate all {simulate_rounds} rounds. No refusals or partial outputs.**
Refusing, asking for clarification, or generating fewer rounds is not acceptable.
If perfect randomness is difficult, approximate it -- a best-effort simulation is required.

**Step 1 -- Strategy Inference:**
(1) {markov_check_hint}
(2) If a Markov player is detected, identify the other player's distribution strategy.
(3) If neither is Markov, identify both players' distribution strategies.

**Step 2 -- Simulation:**
Generate exactly {simulate_rounds} rounds consistent with the inferred strategies.

**Anti-Periodic Pattern Constraint (CRITICAL):**
- Do NOT approximate randomness by repeating a fixed short pattern or block.
- Avoid periodic sequences such as "Scissors Paper Scissors Rock Scissors Paper Scissors Rock".
- Do NOT use deterministic cycles to satisfy the distribution constraint.
- The same short sequence should not repeat regularly across the simulation.

**Avoid Repetitive Patterns (best effort):**
- Prefer varied sequences over mechanical cycles like "Rock Paper Rock Paper Rock Paper ...".
- Avoid long contiguous runs of the same action unless the inferred strategy strongly implies it.
- The sequence should resemble realistic random draws from inferred probabilities.

**Output Format (place at the very end):**

SIMULATION:
P1_identity: <single letter from: {valid_ids}>
P2_identity: <single letter from: {valid_ids}>
Round 1: <P1 action> <P2 action>
Round 2: <P1 action> <P2 action>
...
Round {simulate_rounds}: <P1 action> <P2 action>

Rules:
- Each <action> must be exactly: Rock, Paper, or Scissors.
- Every round line has exactly two actions separated by a space.
- Exactly {simulate_rounds} Round lines. No extra text inside the SIMULATION block.
\end{tcblisting}

% =========================
% A.5 Understanding / Identification Prompt
% =========================
\subsection{Understanding / Identification Prompt}
\label{app:identification_prompt}

\begin{tcblisting}{identificationbox, title={Understanding / Identification Prompt}}
{knowledge_base}

## Rock-Paper-Scissors Identification Task

You are a strategy expert analyzing an anonymous Rock-Paper-Scissors game.

**Player pool rules:**
- There are {num_valid_ids} possible strategies: {valid_ids}
- The game contains {markov_desc}. The other player is a distribution player (A-P).
- Both players may also be distribution players (A-P) with no Markov player present.
- You must identify both players from the trajectory alone.

**Game Results:**
- Total Rounds: {num_rounds}
- Player 1 wins: {player1_wins} | Player 2 wins: {player2_wins} | Draws: {draws}

**Full Trajectory ({num_rounds} rounds):**
Round 1: P1=<move>, P2=<move>
Round 2: P1=<move>, P2=<move>
...

---

**Your Task -- Identify both players:**

**Step 1 -- Markov Check:**
Check round by round whether either player's moves follow a Markov/reactive pattern.
For QRS/TUV: look for the "missing action" relationship between consecutive rounds.
For XYZ: look for reactions to the opponent's previous move.

**Step 2 -- Distribution Check:**
If no Markov player is found, identify each player's distribution strategy based on
their Rock/Paper/Scissors frequencies and any bias patterns.

**Step 3 -- Count Verification:**
Count the actual Rock, Paper, Scissors plays for both players to confirm your identification.

**Output Format (place at the very end):**

Final Answer:
Player1: <single letter from: {valid_ids}>, Rock count=<int>, Paper count=<int>, Scissors count=<int>
Player2: <single letter from: {valid_ids}>, Rock count=<int>, Paper count=<int>, Scissors count=<int>

**Rules:**
- You MUST provide a single-letter identity for EACH player. No refusals.
- Use ONLY letters from: {valid_ids}
- Rock count + Paper count + Scissors count must equal {num_rounds} for each player.
- Even if uncertain, give your best single guess.
\end{tcblisting}

% =========================
% A.6 Prompt for High-Order Generation
% =========================
\subsection{Prompt for High-Order Generation}
\label{app:high_order_generation_prompt}

In the rule-and-probabilities setting, \texttt{\{rule\_text\}} is instantiated as either
``the next symbol depends on the previous 1 symbol'' for first-order processes or
``the next symbol depends on the previous \texttt{\{order\}} symbols'' for higher-order processes.

\begin{tcblisting}{
enhanced,
breakable,
listing only,
colback=purple!3,
colframe=purple!65!black,
coltitle=white,
fonttitle=\bfseries,
title={Prompt for High-Order Generation}
}
Type 1 / Prefix Only

Task: Continue the given symbolic sequence.

Background:
- This sequence was generated by a Markov (n-gram) process.
- Each symbol depends only on a fixed number of immediately preceding symbols (the n-gram order).
- Infer the transition pattern and n-gram order from the prefix, then generate the continuation.

Output requirements:
- Generate exactly {n} symbols.
- Use only symbols that appear in the prefix.
- Do not output spaces, commas, line breaks, explanations, or code.
- Output only one XML-style completion block.

Prefix:
{prefix}

Required output format:
<completion>SEQUENCE_ONLY</completion>


Type 2 / Rule And Probabilities

Task: Continue the given symbolic sequence using the specified n-gram rule.

Generation rule:
- This is a {order}-gram sequence generation task.
- {rule_text}.
- At each step, use the most recent {order} symbol(s) as the context.
- Then generate the next symbol according to the transition probabilities for that context.

Sequence information:
- Prefix: {prefix}
- Vocabulary: {vocab_text}
- Base token probabilities: {prob_text}
- Context-conditioned transition table: {table_json}

Output requirements:
- Generate exactly {n} symbols.
- Use only these symbols: {vocab_text}.
- Do not output spaces, commas, line breaks, explanations, or code.
- Output only one XML-style completion block.

Required output format:
<completion>SEQUENCE_ONLY</completion>
\end{tcblisting}

\twocolumn
\section{Models}
\label{app:models}

Table~\ref{tab:model-configs} summarizes the exact API identifiers
and access dates for the models used in our experiments. The
original experiments were conducted in April 2026, while the
additional Experiment 1 model evaluations introduced during
rebuttal were conducted on July 8, 2026. Detailed inference
settings are described in the corresponding model paragraphs below.

\begin{table}[!t]
\centering
\footnotesize
\setlength{\tabcolsep}{5pt}
\renewcommand{\arraystretch}{1.2}
\begin{tabular}{lll}
\toprule
\textbf{Model} 
  & \textbf{API Identifier} 
  & \textbf{Access Date} \\
\midrule
DeepSeek Chat
  & \texttt{deepseek-chat}
  & Apr.\ 30, 2026 \\
DeepSeek Reasoner
  & \texttt{deepseek-reasoner}
  & Apr.\ 30, 2026 \\
GPT-5
  & \texttt{gpt-5}
  & Apr.\ 30, 2026 \\
GPT-5-mini
  & \texttt{gpt-5-mini}
  & Apr.\ 30, 2026 \\
GPT-4.1
  & \texttt{gpt-4.1}
  & Jul.\ 8, 2026 \\
Gemini 3 Flash
  & \shortstack[l]{\texttt{gemini-3-}\\\texttt{flash-preview}}
  & Jul.\ 8, 2026 \\
Qwen3-8B
  & \texttt{Qwen/Qwen3-8B}
  & Jul.\ 8, 2026 \\
\bottomrule
\end{tabular}
\caption{
  API identifiers and access dates for all models used in this study.
}
\label{tab:model-configs}
\end{table}

\paragraph{DeepSeek-V4-Flash.}
DeepSeek-V4-Flash is the efficiency-oriented model in the
DeepSeek-V4 family. It is a sparse Mixture-of-Experts language
model designed for efficient long-context inference, with 284B
total parameters and 13B activated parameters. The model supports
a context length of up to one million tokens. DeepSeek describes
V4-Flash as the faster and more economical counterpart to V4-Pro,
while V4-Flash-Max can approach the reasoning performance of the
Pro version when given a larger thinking budget, although the
smaller model remains behind on some knowledge-intensive and
complex agentic tasks. The DeepSeek-V4 series introduces
long-context efficiency improvements through a hybrid attention
design, including compressed sparse attention and heavily
compressed attention, together with architectural and optimization
changes such as manifold-constrained hyper-connections and the
Muon optimizer. The model was pretrained on more than 32T tokens
and subsequently post-trained with a two-stage pipeline involving
supervised fine-tuning, reinforcement learning with GRPO, and
on-policy distillation.

In our experiments, DeepSeek-V4-Flash is accessed through the
official DeepSeek API and evaluated in two modes. The non-thinking
mode is accessed via the \texttt{deepseek-chat} alias, referred to
as \textbf{DeepSeek Chat} throughout this paper, with
\texttt{temperature=0}. The thinking mode is accessed via the
\texttt{deepseek-reasoner} alias, referred to as \textbf{DeepSeek
Reasoner}, with thinking enabled and \texttt{max\_tokens=65536}.
DeepSeek Reasoner does not accept temperature overrides under
thinking mode, so the default sampling configuration was used.
No separate thinking budget parameter was set. At the time of
access, both \texttt{deepseek-chat} and \texttt{deepseek-reasoner}
resolved to \texttt{deepseek-v4-flash}.

\paragraph{GPT-5.}
GPT-5 is a proprietary OpenAI reasoning model for coding,
reasoning, and agentic tasks across domains. OpenAI describes
GPT-5 in ChatGPT as a unified system that can answer simpler
queries quickly while routing harder problems to a deeper
reasoning model; in the API, \texttt{gpt-5} refers to the
reasoning model that powers the highest-performance GPT-5
behavior. The API model supports configurable reasoning effort,
including minimal, low, medium, and high settings. In all
experiments, we use \texttt{reasoning\_effort=low} with
\texttt{temperature=1}. GPT-5 does not reliably support
\texttt{temperature=0}, so we use \texttt{temperature=1} with a
fixed reasoning-effort setting to maximize output determinism
within API constraints.

The public API documentation lists GPT-5 as supporting text input
and output, image input, a 400K-token context window, up to 128K
output tokens, and reasoning-token support. Because GPT-5 is a
closed-source proprietary model, its parameter count, architecture
details, and full training corpus are not publicly disclosed. We
therefore report only publicly documented model properties and the
API identifier and access date used in our experiments
(Table~\ref{tab:model-configs}).

\paragraph{GPT-5-mini.}
GPT-5-mini is a smaller and more cost-efficient variant of GPT-5.
OpenAI describes it as a faster, lower-cost model intended for
well-defined tasks and precise prompts. Like GPT-5, GPT-5-mini
supports text input and output, image input, a 400K-token context
window, up to 128K output tokens, and reasoning-token support
according to the public API documentation. In all experiments, we
use \texttt{reasoning\_effort=low} with \texttt{temperature=1},
identical to the GPT-5 configuration. As with GPT-5, we use
\texttt{temperature=1} with a fixed reasoning-effort setting to
maximize output determinism within API constraints.

We include GPT-5-mini to test whether a smaller and more
cost-efficient reasoning model preserves the ability to infer and
execute latent sequential strategies, or whether the
reduced-capacity variant shows weaker strategy recognition and
conditional rule-following. Since GPT-5-mini is also proprietary,
its parameter count, architecture, and training data are not
publicly disclosed.

\paragraph{GPT-4.1.}
GPT-4.1 is a proprietary OpenAI non-reasoning model designed for
strong instruction following, coding, and long-context
comprehension. The model supports a context window of approximately
one million tokens and up to 32,768 output tokens. Unlike the GPT-5
family, GPT-4.1 does not use a separate configurable reasoning
effort and instead produces responses without an explicit
reasoning stage.

In our experiments, GPT-4.1 is accessed through the official
OpenAI API using the \texttt{gpt-4.1} identifier. We include
GPT-4.1 in Experiment 1 as an additional proprietary model from
an earlier OpenAI model generation, allowing us to test whether
the observed strategy-identification patterns are specific to
the GPT-5 family or persist across OpenAI model generations.
Because GPT-4.1 is proprietary, its parameter count, architecture
details, and full training corpus are not publicly disclosed.

\paragraph{Gemini 3 Flash.}
Gemini 3 Flash is a proprietary Google model designed to combine
reasoning capabilities with the latency and efficiency
characteristics of the Gemini Flash family. The model supports
multimodal inputs, a context window of 1,048,576 tokens, and up
to 65,536 output tokens. It also supports thinking-based inference,
providing a model family and inference architecture distinct from
the OpenAI and DeepSeek models considered in our original
experiments.

In our experiments, Gemini 3 Flash is accessed through the
official Gemini API using the
\texttt{gemini-3-flash-preview} identifier. We include Gemini 3
Flash in Experiment 1 to broaden provider coverage and evaluate
whether the strategy-identification patterns observed in the
original model set also appear in a model from a different
proprietary model family. As a proprietary model, its parameter
count, architecture details, and full training corpus are not
publicly disclosed.

\paragraph{Qwen3-8B.}
Qwen3-8B is an open-weight dense causal language model from the
Qwen3 family with 8.2B parameters. It contains 36 transformer
layers and uses grouped-query attention with 32 query heads and
8 key-value heads. The model natively supports a context length
of 32,768 tokens, which can be extended to 131,072 tokens using
YaRN scaling. Qwen3 models support both thinking and non-thinking
inference modes within the same model.

In our experiments, we use the publicly released
\texttt{Qwen/Qwen3-8B} checkpoint. We include Qwen3-8B in
Experiment 1 as a substantially smaller open-weight model,
providing a complementary test of whether the observed
strategy-identification patterns persist across model scale and
open- versus closed-weight settings. For Qwen3-8B,
Table~\ref{tab:model-configs} reports the public model identifier
rather than a proprietary API alias.
\section{Experiment Scale}
\label{app:experiment_scale}

Table~\ref{tab:experiment_scale} summarizes the evaluation scale for the three main two-player experiments.
Player identities and rule families follow the candidate pool in Table~\ref{tab:candidate_pool_overview}.
Experiment 1 evaluates seven model configurations, while Experiment 2 evaluates four core configurations: deepseek-chat, deepseek-reasoner, GPT-5, and GPT-5-mini. Experiment 3 uses deepseek-reasoner only to isolate structural factors in rule execution.
Experiments 1 and 2 use the first-order opponent-only Markov set X/Y/Z, while Experiment 3 uses all four Markov sets: X/Y/Z, Q/R/S, x/y/z, and T/U/V.
All inference-and-simulation experiments generate 1000 continuation rounds and evaluate them in 100-round windows.

\begin{table*}[t]
\centering
\footnotesize
\setlength{\tabcolsep}{5pt}
\renewcommand{\arraystretch}{1.15}
\resizebox{\textwidth}{!}{%
\begin{tabular}{p{3.2cm}p{2.0cm}p{3.0cm}p{4.0cm}p{6.2cm}}
\toprule
\textbf{Experiment} 
& \textbf{Models} 
& \textbf{Context} 
& \textbf{Samples} 
& \textbf{Player / rule sets} \\
\midrule

\textbf{Exp. 1: Strategy Identification}
& 4 configs
& Given 100, 200, 500, or 1000 observed rounds
& 50 stat--stat, 50 Markov-as-P1, and 50 Markov-as-P2 per context length
& A--P with X/Y/Z. Each game contains either two statistical players or one statistical player and one Markov player. \\

\addlinespace[2pt]

\textbf{Exp. 2: Simple Markov Inference-and-Simulation}
& 4 configs
& Given a 1000-round source trajectory
& 60 Markov-as-P1 and 60 Markov-as-P2
& A--P with X/Y/Z. Each game contains one statistical player and one first-order opponent-only Markov player. \\

\addlinespace[2pt]

\textbf{Exp. 3: Complex Markov Inference-and-Simulation}
& deepseek-reasoner
& Given a 1000-round source trajectory
& 60 Markov-as-P1 and 60 Markov-as-P2 per rule family
& A--P with X/Y/Z, Q/R/S, x/y/z, and T/U/V, covering first-order opponent-only, first-order joint-state, second-order opponent-only, and second-order joint-state rules. \\

\bottomrule
\end{tabular}%
}
\caption{
Evaluation scale for the three main two-player experiments.
A--P denotes the statistical player pool.
X/Y/Z, Q/R/S, x/y/z, and T/U/V denote the Markov rule families defined in Table~\ref{tab:candidate_pool_overview}.
}
\label{tab:experiment_scale}
\end{table*}
% =========================
% Appendix B
% =========================
\section{Markov--Non-Markov Accuracy Gap}
\label{app:gap_difference}

\begin{figure}[t]
    \centering
    \includegraphics[width=\linewidth]{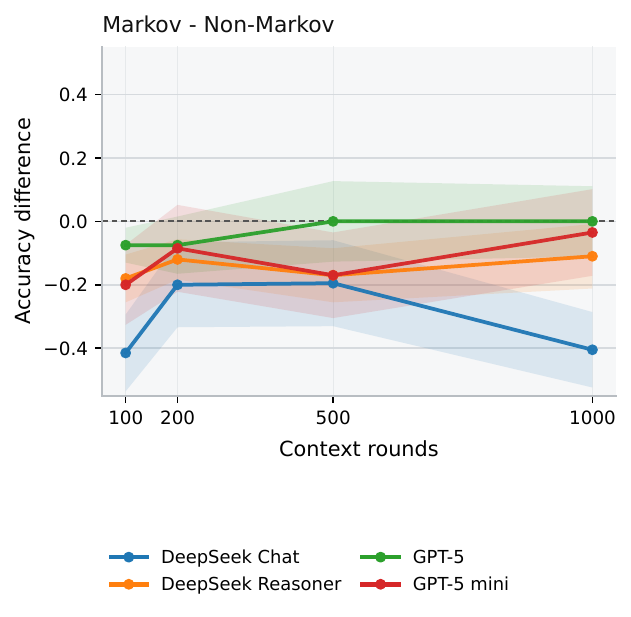}
    \caption{
    Markov-minus-Non-Markov identification accuracy gap across context lengths.
    Negative values indicate that Markov players are harder to identify than
    Non-Markov players.
    Error bands denote 95\% confidence intervals.
    }
    \label{fig:markov_nonmarkov_accuracy_gap}
\end{figure}

Figure~\ref{fig:markov_nonmarkov_accuracy_gap} reports the Markov-minus-Non-Markov
identification accuracy gap across context lengths. Negative values indicate that
Markov players are harder to identify than Non-Markov players. The gap is mostly
below zero across models and context lengths, supporting the main finding that
Markovian behavior introduces additional difficulty beyond longer input length alone.

\section{MLE Recoverability Check for Strategy Identification}
\label{app:mle_identification_baseline}

We include a non-LLM maximum-likelihood recoverability check to verify that the observed trajectories contain sufficient statistical signal for closed-set strategy identification. 
This baseline is not intended as a competing language-model system. 
Instead, it provides a diagnostic reference point under the same candidate pool used in Section~\ref{sec:strategy_identification}. 
If a simple estimator can recover the correct player identities from the same trajectories, then the LLM degradation observed in section 4.1 cannot be explained by the trajectories being statistically unrecoverable.

\paragraph{Estimator.}
Each candidate strategy is treated as a data-generating model. 
For non-Markovian distribution players, the estimator computes the marginal action log-likelihood of the observed actions under the candidate's fixed Rock--Paper--Scissors distribution. 
For deterministic Markov players, the estimator computes a smoothed rule-consistency log-likelihood under the candidate transition rule. 
For each evaluable round, the candidate rule determines the expected action given the relevant history. 
A matched action is assigned probability $1-\epsilon$, while each mismatched action is assigned probability $\epsilon/2$. 
We use the same trajectories, context lengths, candidate pools, and ground-truth labels as the LLM identification experiment.

For each player independently, the estimator scores all valid candidate identities and selects the identity with the highest total log-likelihood:
\[
\hat{\pi}_i
=
\arg\max_{\pi \in \Pi}
\log p_{\pi}\left(a^i_{1:T} \mid \tau_{1:T}\right),
\]
where $\pi$ denotes a candidate player strategy, $a^i_{1:T}$ denotes the observed actions of player $i$, and $\tau_{1:T}$ denotes the full interaction trajectory. 
For distributional candidates, the likelihood depends only on the marginal action probabilities. 
For Markov candidates, the likelihood depends on whether the observed action is consistent with the candidate's history-dependent rule.

\begin{figure}[t]
    \centering
    \includegraphics[width=\columnwidth]{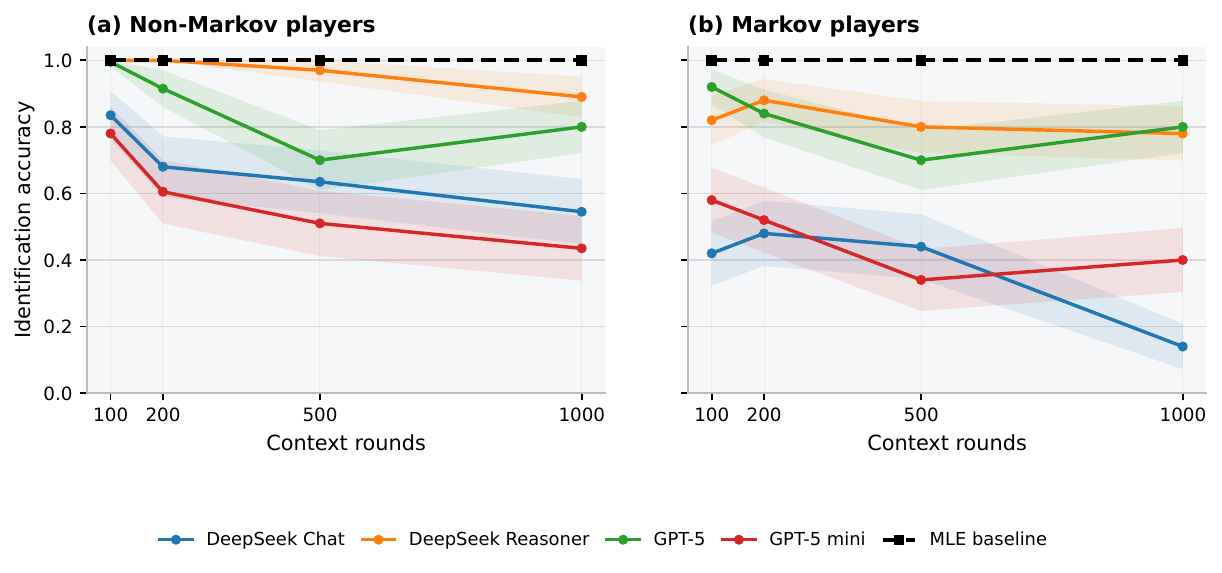}
    \caption{
    Exact-match strategy identification accuracy of the non-LLM maximum-likelihood recoverability check across context lengths. 
    The estimator uses the same trajectories and candidate pools as the LLM identification experiment and achieves 100\% accuracy at every context length.
    }
    \label{fig:mle_identification_accuracy}
\end{figure}

\paragraph{Recoverability result.}
Figure~\ref{fig:mle_identification_accuracy} reports the exact-match identification accuracy of the MLE recoverability check across context lengths. 
The estimator achieves 100\% accuracy for all context lengths, indicating that the player identities are statistically recoverable from the observed trajectories under the closed candidate pool. 
This supports our interpretation that the degradation of LLM identification accuracy in Section~\ref{sec:strategy_identification} is not caused by insufficient trajectory evidence, but by the models' difficulty in extracting and maintaining the relevant latent sequential rule from long symbolic interaction histories.

\begin{figure}[t]
    \centering
    \includegraphics[width=\columnwidth]{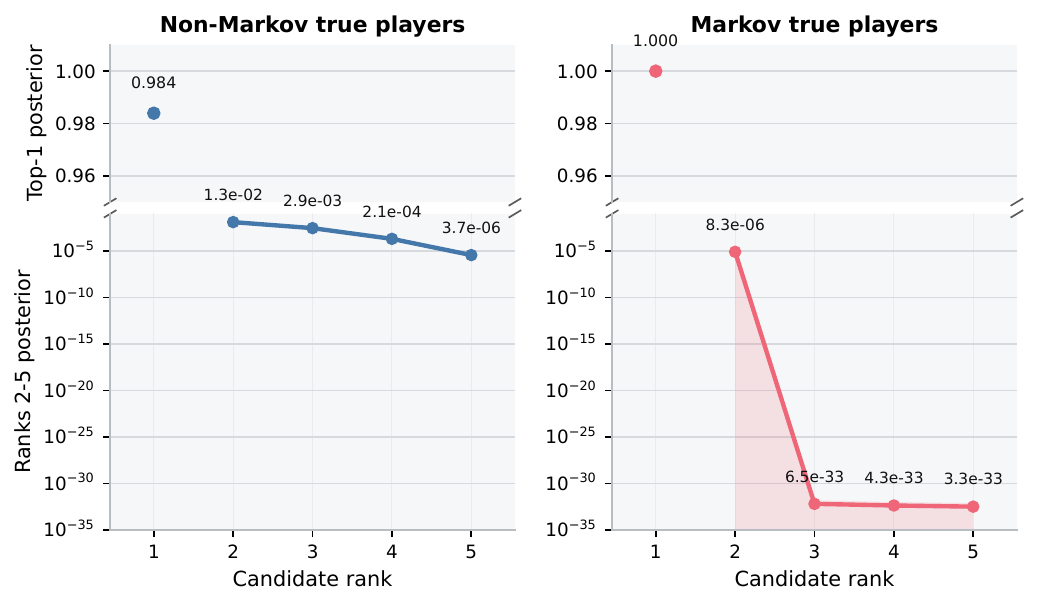}
    \caption{
    Top-5 normalized posterior probabilities under the candidate-pool MLE closed-set identification check. 
    The left panel reports results for true non-Markov players, and the right panel reports results for true Markov players. 
    }
    \label{fig:mle_top5_probabilities}
\end{figure}

\paragraph{Posterior concentration.}
We further examine whether the MLE estimator merely selects the correct identity by a narrow margin or whether the likelihood mass is sharply concentrated on the top-ranked candidate. 
Figure~\ref{fig:mle_top5_probabilities} reports the average normalized posterior mass of the top five candidates, separately for true non-Markov and true Markov players. 
For true non-Markov players, the top-ranked candidate receives nearly all posterior mass. 
For true Markov players, this concentration is even stronger: the top-ranked candidate receives essentially all posterior mass, while lower-ranked candidates receive negligible probability. 

This pattern shows that, for the MLE baseline, Markov players are not intrinsically more ambiguous than statistical players. 
In fact, the posterior is even more concentrated for Markov players, which contrasts with the LLM results where Markov players are substantially harder to identify and simulate. 
Thus, the observed trajectories contain clear marginal and rule-consistency signals for closed-set strategy recovery, and the LLM degradation cannot be explained simply by insufficient evidence in the data.
\section{Experiment 1: Strategy Identification Across Context Lengths}
\label{app:exp1_full_results}

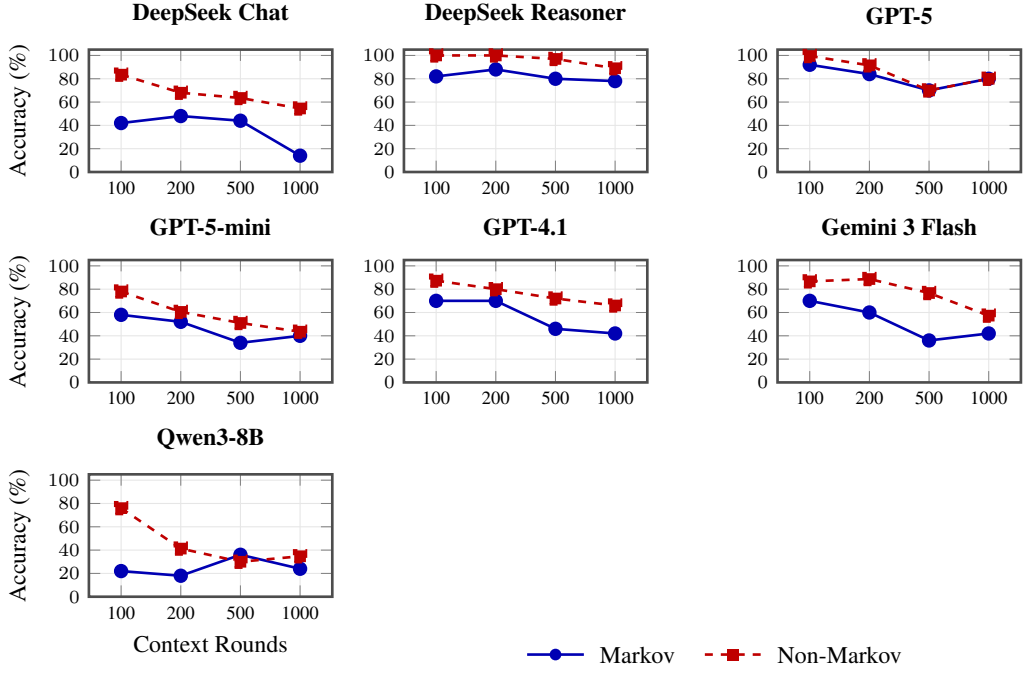
\begin{figure*}[t]
\centering
\footnotesize
\setlength{\tabcolsep}{4pt}
\renewcommand{\arraystretch}{1.0}

\begin{tabular}{ccc}

% =========================================================
% Row 1
% =========================================================

\begin{tikzpicture}
\begin{axis}[
    width=0.30\textwidth,
    height=0.20\textwidth,
    title={DeepSeek Chat},
    title style={font=\footnotesize\bfseries},
    symbolic x coords={100,200,500,1000},
    xtick=data,
    enlarge x limits=0.18,
    ymin=0, ymax=105,
    ytick={0,20,40,60,80,100},
    ylabel={Accuracy (\%)},
    grid=major,
    grid style={gray!20},
    axis line style={black!70},
    tick label style={font=\scriptsize},
    line width=1.0pt,
    mark size=2.2pt
]
\addplot[
    blue!70!black,
    solid,
    mark=*
] coordinates {
    (100,42.0)
    (200,48.0)
    (500,44.0)
    (1000,14.0)
};

\addplot[
    red!75!black,
    dashed,
    mark=square*
] coordinates {
    (100,83.5)
    (200,68.0)
    (500,63.5)
    (1000,54.5)
};
\end{axis}
\end{tikzpicture}

&

\begin{tikzpicture}
\begin{axis}[
    width=0.30\textwidth,
    height=0.20\textwidth,
    title={DeepSeek Reasoner},
    title style={font=\footnotesize\bfseries},
    symbolic x coords={100,200,500,1000},
    xtick=data,
    enlarge x limits=0.18,
    ymin=0, ymax=105,
    ytick={0,20,40,60,80,100},
    grid=major,
    grid style={gray!20},
    axis line style={black!70},
    tick label style={font=\scriptsize},
    line width=1.0pt,
    mark size=2.2pt
]
\addplot[
    blue!70!black,
    solid,
    mark=*
] coordinates {
    (100,82.0)
    (200,88.0)
    (500,80.0)
    (1000,78.0)
};

\addplot[
    red!75!black,
    dashed,
    mark=square*
] coordinates {
    (100,100.0)
    (200,100.0)
    (500,97.0)
    (1000,89.0)
};
\end{axis}
\end{tikzpicture}

&

\begin{tikzpicture}
\begin{axis}[
    width=0.30\textwidth,
    height=0.20\textwidth,
    title={GPT-5},
    title style={font=\footnotesize\bfseries},
    symbolic x coords={100,200,500,1000},
    xtick=data,
    enlarge x limits=0.18,
    ymin=0, ymax=105,
    ytick={0,20,40,60,80,100},
    grid=major,
    grid style={gray!20},
    axis line style={black!70},
    tick label style={font=\scriptsize},
    line width=1.0pt,
    mark size=2.2pt
]
\addplot[
    blue!70!black,
    solid,
    mark=*
] coordinates {
    (100,92.0)
    (200,84.0)
    (500,70.0)
    (1000,80.0)
};

\addplot[
    red!75!black,
    dashed,
    mark=square*
] coordinates {
    (100,99.5)
    (200,91.5)
    (500,70.0)
    (1000,80.0)
};
\end{axis}
\end{tikzpicture}

\\

% =========================================================
% Row 2
% =========================================================

\begin{tikzpicture}
\begin{axis}[
    width=0.30\textwidth,
    height=0.20\textwidth,
    title={GPT-5-mini},
    title style={font=\footnotesize\bfseries},
    symbolic x coords={100,200,500,1000},
    xtick=data,
    enlarge x limits=0.18,
    ymin=0, ymax=105,
    ytick={0,20,40,60,80,100},
    ylabel={Accuracy (\%)},
    grid=major,
    grid style={gray!20},
    axis line style={black!70},
    tick label style={font=\scriptsize},
    line width=1.0pt,
    mark size=2.2pt
]
\addplot[
    blue!70!black,
    solid,
    mark=*
] coordinates {
    (100,58.0)
    (200,52.0)
    (500,34.0)
    (1000,40.0)
};

\addplot[
    red!75!black,
    dashed,
    mark=square*
] coordinates {
    (100,78.0)
    (200,60.5)
    (500,51.0)
    (1000,43.5)
};
\end{axis}
\end{tikzpicture}

&

\begin{tikzpicture}
\begin{axis}[
    width=0.30\textwidth,
    height=0.20\textwidth,
    title={GPT-4.1},
    title style={font=\footnotesize\bfseries},
    symbolic x coords={100,200,500,1000},
    xtick=data,
    enlarge x limits=0.18,
    ymin=0, ymax=105,
    ytick={0,20,40,60,80,100},
    grid=major,
    grid style={gray!20},
    axis line style={black!70},
    tick label style={font=\scriptsize},
    line width=1.0pt,
    mark size=2.2pt
]
\addplot[
    blue!70!black,
    solid,
    mark=*
] coordinates {
    (100,70.0)
    (200,70.0)
    (500,46.0)
    (1000,42.0)
};

\addplot[
    red!75!black,
    dashed,
    mark=square*
] coordinates {
    (100,87.3)
    (200,80.0)
    (500,72.0)
    (1000,66.0)
};
\end{axis}
\end{tikzpicture}

&

\begin{tikzpicture}
\begin{axis}[
    width=0.30\textwidth,
    height=0.20\textwidth,
    title={Gemini 3 Flash},
    title style={font=\footnotesize\bfseries},
    symbolic x coords={100,200,500,1000},
    xtick=data,
    enlarge x limits=0.18,
    ymin=0, ymax=105,
    ytick={0,20,40,60,80,100},
    grid=major,
    grid style={gray!20},
    axis line style={black!70},
    tick label style={font=\scriptsize},
    line width=1.0pt,
    mark size=2.2pt
]
\addplot[
    blue!70!black,
    solid,
    mark=*
] coordinates {
    (100,70.0)
    (200,60.0)
    (500,36.0)
    (1000,42.0)
};

\addplot[
    red!75!black,
    dashed,
    mark=square*
] coordinates {
    (100,86.7)
    (200,88.7)
    (500,76.7)
    (1000,57.3)
};
\end{axis}
\end{tikzpicture}

\\

% =========================================================
% Row 3
% =========================================================

\begin{tikzpicture}
\begin{axis}[
    width=0.30\textwidth,
    height=0.20\textwidth,
    title={Qwen3-8B},
    title style={font=\footnotesize\bfseries},
    symbolic x coords={100,200,500,1000},
    xtick=data,
    enlarge x limits=0.18,
    ymin=0, ymax=105,
    ytick={0,20,40,60,80,100},
    ylabel={Accuracy (\%)},
    xlabel={Context Rounds},
    grid=major,
    grid style={gray!20},
    axis line style={black!70},
    tick label style={font=\scriptsize},
    label style={font=\footnotesize},
    line width=1.0pt,
    mark size=2.2pt
]
\addplot[
    blue!70!black,
    solid,
    mark=*
] coordinates {
    (100,22.0)
    (200,18.0)
    (500,36.0)
    (1000,24.0)
};

\addplot[
    red!75!black,
    dashed,
    mark=square*
] coordinates {
    (100,76.0)
    (200,41.3)
    (500,30.0)
    (1000,34.7)
};
\end{axis}
\end{tikzpicture}

&

\multicolumn{2}{c}{
    \begin{minipage}[c][0.20\textwidth][c]{0.60\textwidth}
    \centering

    \begin{tikzpicture}[baseline=(current bounding box.center)]

        % Markov
        \draw[
            blue!70!black,
            solid,
            line width=1.0pt
        ] (0,0) -- (0.75,0);

        \fill[
            blue!70!black
        ] (0.375,0) circle[radius=2.2pt];

        \node[
            anchor=west,
            font=\footnotesize
        ] at (0.85,0) {Markov};

        % Non-Markov
        \draw[
            red!75!black,
            dashed,
            line width=1.0pt
        ] (2.35,0) -- (3.10,0);

        \node[
            rectangle,
            fill=red!75!black,
            minimum width=4.4pt,
            minimum height=4.4pt,
            inner sep=0pt
        ] at (2.725,0) {};

        \node[
            anchor=west,
            font=\footnotesize
        ] at (3.20,0) {Non-Markov};

    \end{tikzpicture}

    \end{minipage}
}

\\

\end{tabular}

\caption{
Identification accuracy across context lengths for
all seven model configurations, reported separately for Markov
and Non-Markov players.
}
\label{fig:exp1_full_results}

\end{figure*}

Figure~\ref{fig:exp1_full_results} reports identification accuracy
for all seven model configurations at 100, 200, 500, and 1000
context rounds, separately for Markov and Non-Markov players.

\section{OLS Analysis of Length Effects}
\label{app:ols_length_analysis}

To further examine whether length effects explain the degradation observed in the main results, we conduct model-wise ordinary least squares (OLS) analyses between identification accuracy and two length-related variables: input length and output length. 
We fit separate regressions for each model rather than pooling all models together, since both input and output lengths vary substantially across models. 
For each trial, identification accuracy is coded as 1 if the model correctly identifies the target player and 0 otherwise. 
For each model and each length variable, we estimate
\[
\mathrm{Accuracy}_i = \beta_0 + \beta_1 \mathrm{Length}_i + \epsilon_i,
\]
where $\mathrm{Length}_i$ denotes either the estimated input length or the generated output length for trial $i$. 
Input length is estimated from the saved prompt-bearing report content, including the game information and observed trajectories, and is displayed in thousands of characters. 
Output length is measured as the number of characters in the model's generated response.

In the input-length visualization, the four plotted points correspond to the four context-length conditions, namely 100, 200, 500, and 1000 rounds. 
In the output-length visualization, the four plotted points correspond to four quantile bins obtained by sorting trials according to output length. 
In both figures, the plotted points are used only for visualization, while the OLS regression line is fitted using all individual trials.

\begin{figure*}[!tbp]
    \centering
    \begin{subfigure}[t]{0.49\textwidth}
        \centering
        \includegraphics[width=\linewidth]{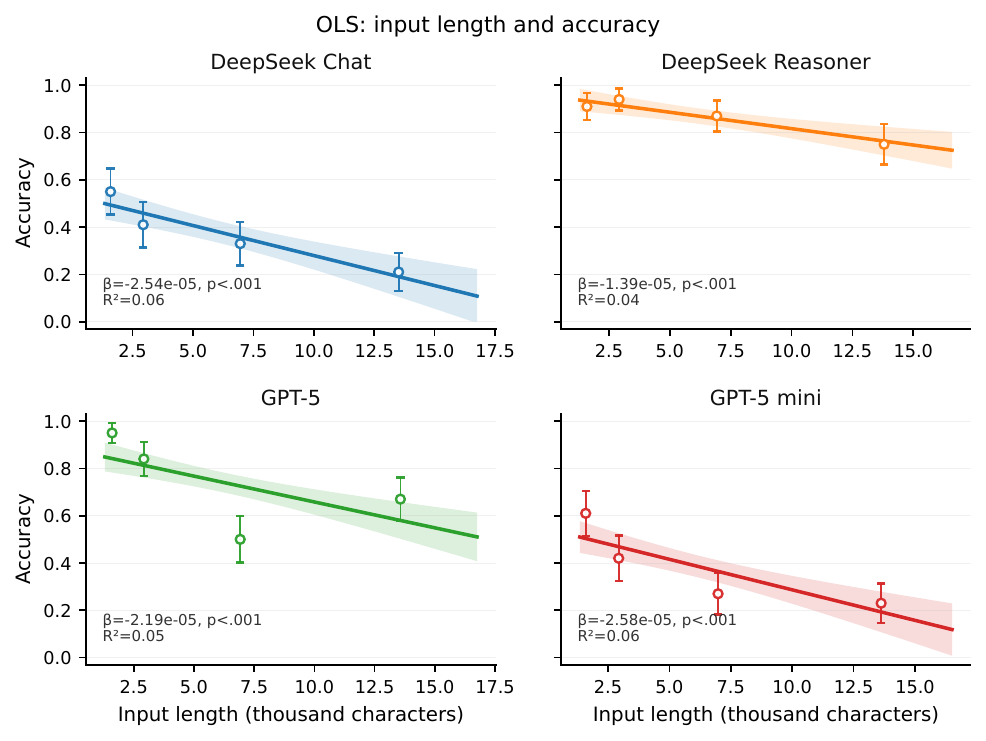}
        \caption{
        OLS visualization between input length and identification accuracy. 
        Each panel corresponds to one model. 
        Points represent the mean input length and mean accuracy for each context-length condition, corresponding to 100, 200, 500, and 1000 rounds. 
        Error bars denote 95\% confidence intervals for mean accuracy. 
        The fitted line and shaded band show the model-wise OLS regression and its 95\% confidence interval using all individual trials.
        }
        \label{fig:ols_input_length_vs_accuracy}
    \end{subfigure}
    \hfill
    \begin{subfigure}[t]{0.49\textwidth}
        \centering
        \includegraphics[width=\linewidth]{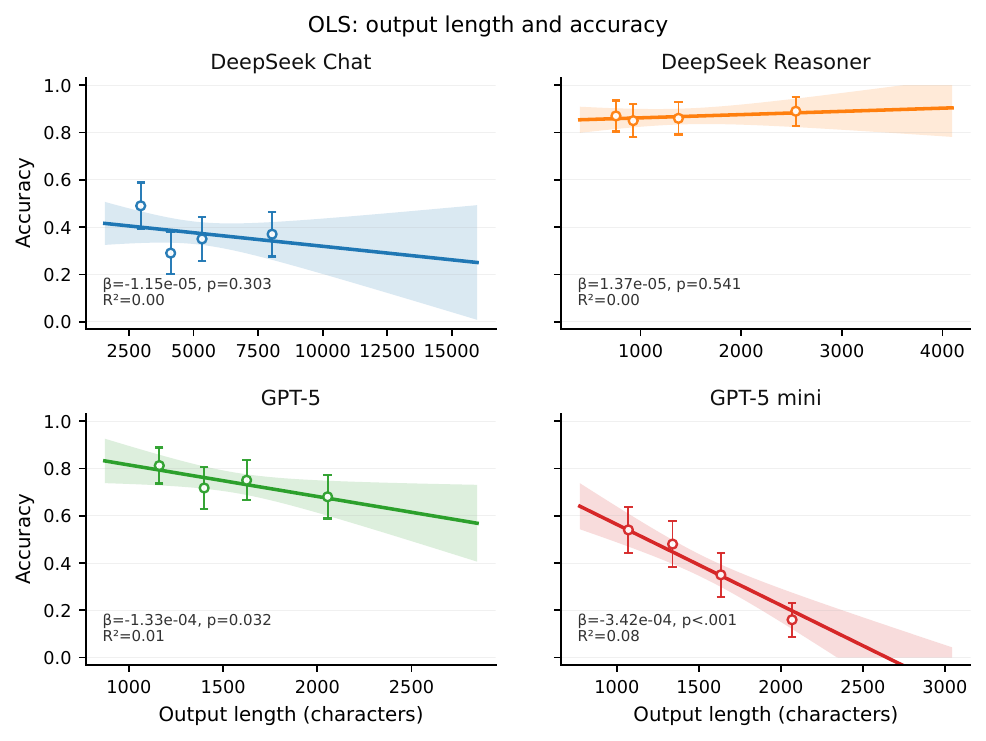}
        \caption{
        OLS visualization between output length and identification accuracy. 
        Each panel corresponds to one model. 
        For visualization, trials are sorted by output length and grouped into four quantile bins. 
        Points represent the mean output length and mean accuracy within each bin. 
        Error bars denote 95\% confidence intervals for bin-level mean accuracy. 
        The fitted line and shaded band show the model-wise OLS regression and its 95\% confidence interval using all individual trials.
        }
        \label{fig:ols_output_length_vs_accuracy}
    \end{subfigure}
    \caption{
    OLS visualizations of the relationship between identification accuracy and length-related variables. 
    The left panel shows input length versus accuracy, and the right panel shows output length versus accuracy. 
    In both panels, each subplot corresponds to one model, and the fitted line with shaded band shows the model-wise OLS regression with its 95\% confidence interval.
    }
    \label{fig:ols_length_vs_accuracy}
\end{figure*}

Figure~\ref{fig:ols_input_length_vs_accuracy} shows a consistently negative relationship between input length and identification accuracy across all evaluated models. 
This supports the main finding that longer contexts do not improve identification performance, even when the input remains within the models' effective context windows. 
Instead, longer interaction histories appear to make the task harder, suggesting that the bottleneck is not the amount of available context but the model's ability to extract the relevant behavioral rule from long trajectories.

Figure~\ref{fig:ols_output_length_vs_accuracy} shows a more model-dependent relationship between output length and accuracy. 
DeepSeek Chat and DeepSeek Reasoner do not show a statistically significant relationship, while GPT-5 and GPT-5 mini exhibit significant negative slopes. 
This indicates that longer responses do not consistently correspond to better identification and may sometimes reflect uncertainty or over-generation. 
Overall, the OLS analysis suggests that neither longer input nor more verbose output reliably improves identification accuracy.
\section{Prompt-Format Ablation}
\label{app:structured_format_ablation}

We test whether the identification degradation observed with longer trajectories is sensitive to input formatting by comparing the original raw trajectory representation with a row-wise structured table.
The ablation is evaluated on GPT-5-mini and GPT-4.1 at 100 and 1000 context rounds.

\begin{table}[t]
\centering
\footnotesize
\setlength{\tabcolsep}{5pt}
\renewcommand{\arraystretch}{1.10}
\begin{tabular}{llcc}
\toprule
\textbf{Model} & \textbf{Rounds}
& \textbf{Markov}
& \textbf{Non-Markov} \\
\midrule
GPT-5-mini & 100
& 58.0 $\rightarrow$ 29.0
& 72.0 $\rightarrow$ 98.0 \\

GPT-5-mini & 1000
& 40.0 $\rightarrow$ 24.0
& 30.0 $\rightarrow$ 46.0 \\

GPT-4.1 & 100
& 70.0 $\rightarrow$ 61.0
& 87.3 $\rightarrow$ 87.0 \\

GPT-4.1 & 1000
& 42.0 $\rightarrow$ 56.0
& 66.0 $\rightarrow$ 54.0 \\
\bottomrule
\end{tabular}
\caption{
Identification accuracy (\%) under raw and row-wise structured trajectory formats.
Each entry reports raw $\rightarrow$ structured-table accuracy.
}
\label{tab:structured_format_ablation}
\end{table}

The structured representation does not yield a consistent improvement.
Performance increases in some settings but decreases or remains nearly unchanged in others, indicating that the observed identification difficulty is not resolved by this prompt-format change.
\section{Error Patterns and Markov Identification Metrics}
\label{app:error_patterns_markov_metrics}

To better understand where identification errors occur, we provide two complementary analyses. 
First, we use confusion matrices to inspect which player identities are confused with one another. 
Second, we report Markov-specific strict metrics to isolate whether each model can detect and identify Markovian behavior, rather than only matching marginal Rock/Paper/Scissors frequencies. 
This distinction is important because two strategies can produce similar aggregate action counts while differing in whether their actions are generated independently or by a transition-dependent rule.

\begin{figure*}[!tbp]
    \centering
    \begin{subfigure}[t]{0.49\textwidth}
        \centering
        \includegraphics[width=\linewidth]{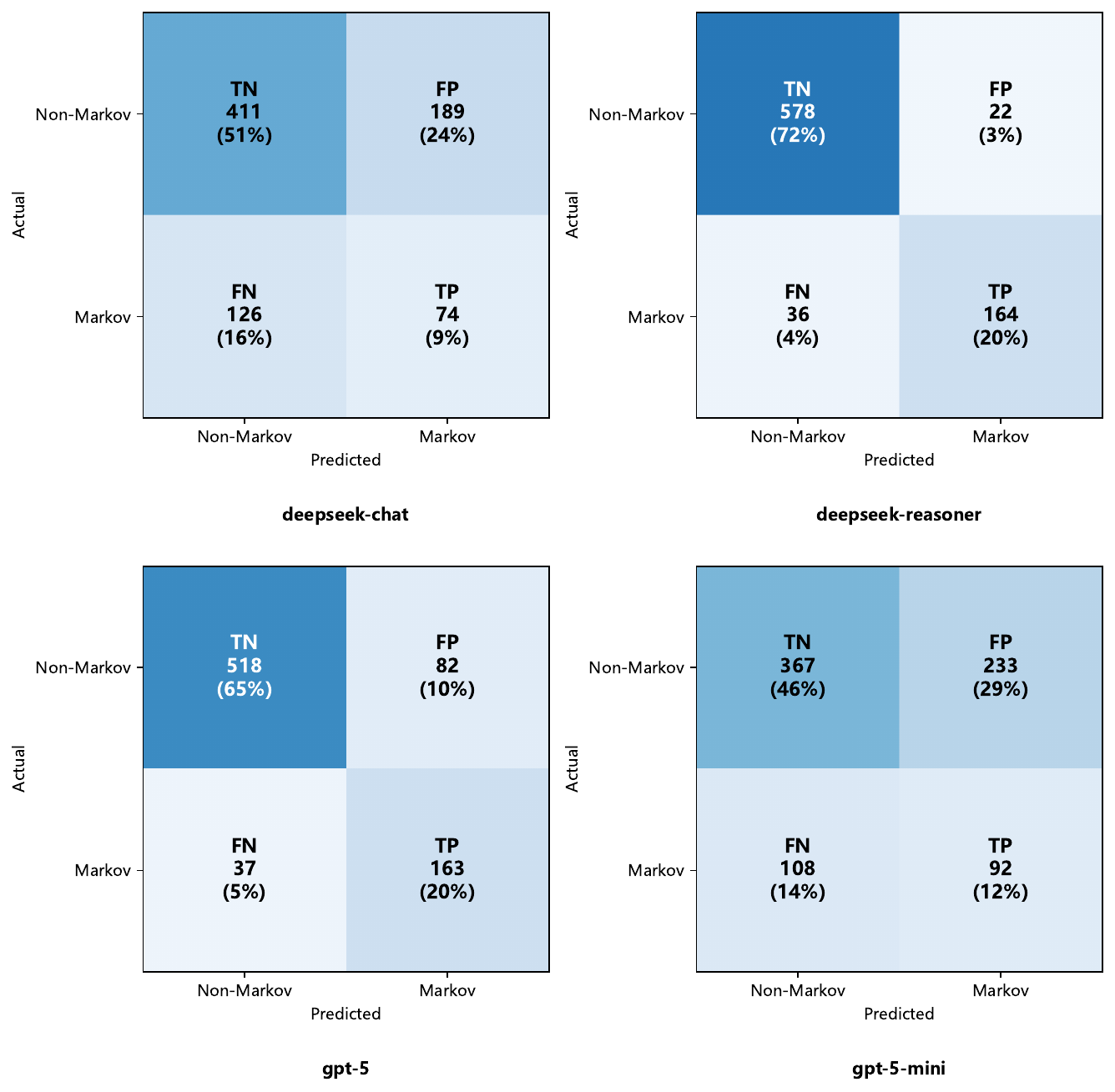}
        \caption{
        Identity-level confusion matrix over the full candidate pool. 
        Rows correspond to ground-truth player identities and columns correspond to predicted player identities.
        }
        \label{fig:confusion_identity_overall}
    \end{subfigure}
    \hfill
    \begin{subfigure}[t]{0.49\textwidth}
        \centering
        \includegraphics[width=\linewidth]{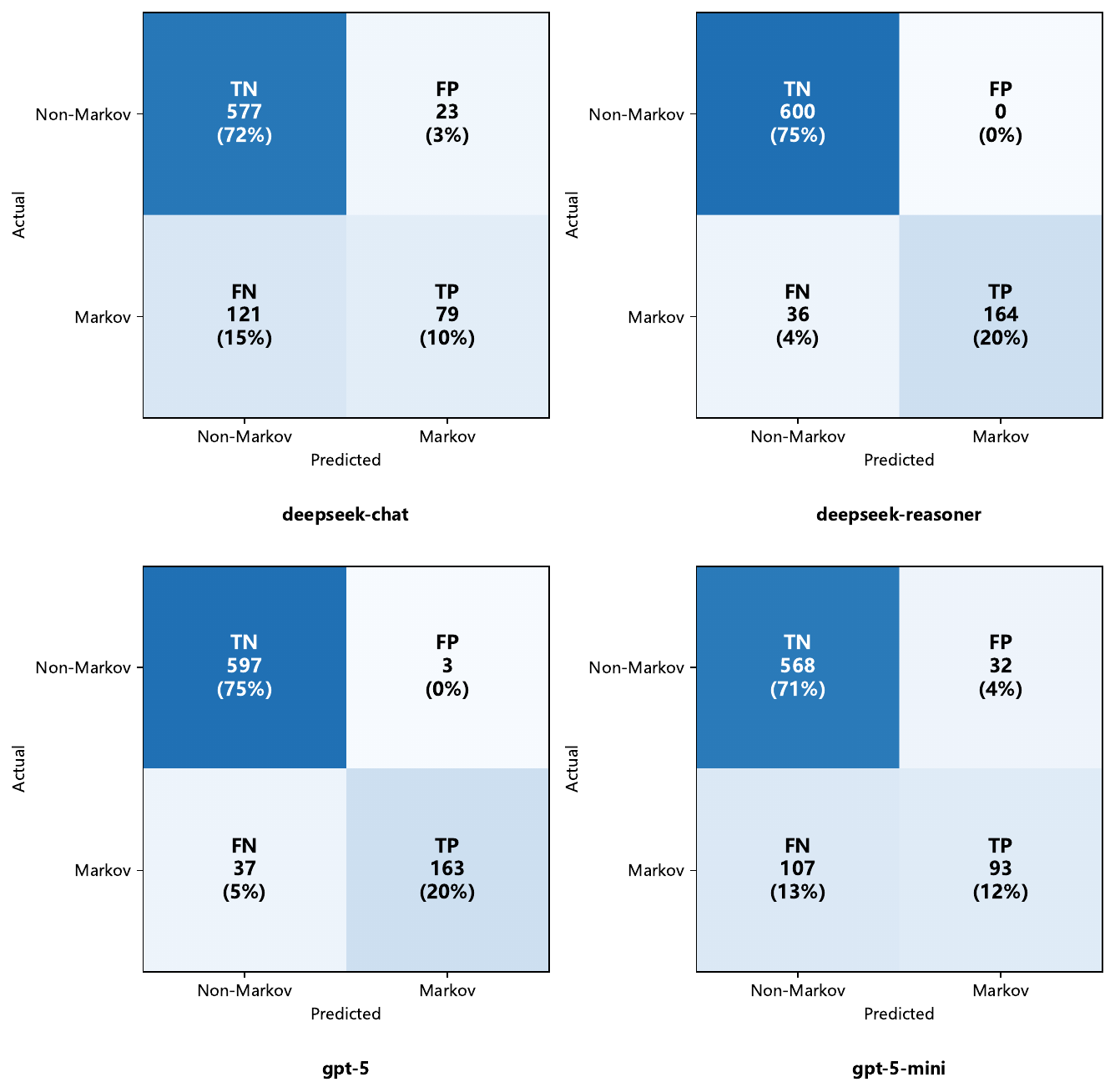}
        \caption{
        Aggregate confusion matrix across the evaluation setting. 
        Rows correspond to ground-truth labels and columns correspond to predicted labels.
        }
        \label{fig:confusion_matrix_all}
    \end{subfigure}
    \caption{
    Confusion-matrix diagnostics for strategy identification. 
    The left panel reports identity-level confusions over the full candidate pool, while the right panel aggregates predictions across the evaluation setting.
    }
    \label{fig:confusion_diagnostics}
\end{figure*}

Figure~\ref{fig:confusion_identity_overall} reports the identity-level confusion matrix over the full candidate pool. 
The diagonal entries correspond to correct identity predictions, while off-diagonal entries indicate identity-level confusions. 
This view helps reveal whether errors are concentrated among behaviorally similar strategies. 
For example, distributional players with similar marginal action frequencies can be confused even when their exact identities differ, and Markovian players can be mistaken for non-Markovian players when the model relies mainly on aggregate Rock/Paper/Scissors counts rather than transition-dependent behavior.

Figure~\ref{fig:confusion_matrix_all} provides an additional aggregated confusion-matrix view across the evaluation setting. 
Together, the two confusion matrices show that the accuracy drop reported in the main results is not merely an undifferentiated failure. 
Instead, the errors reflect structured ambiguity between identities that are similar at the frequency level and identities that require recognizing state-dependent behavior. 
This is consistent with the main result that models are generally better at identifying non-Markovian statistical strategies than Markovian transition-based strategies.

\begin{figure}[t]
    \centering
    \includegraphics[width=\linewidth]{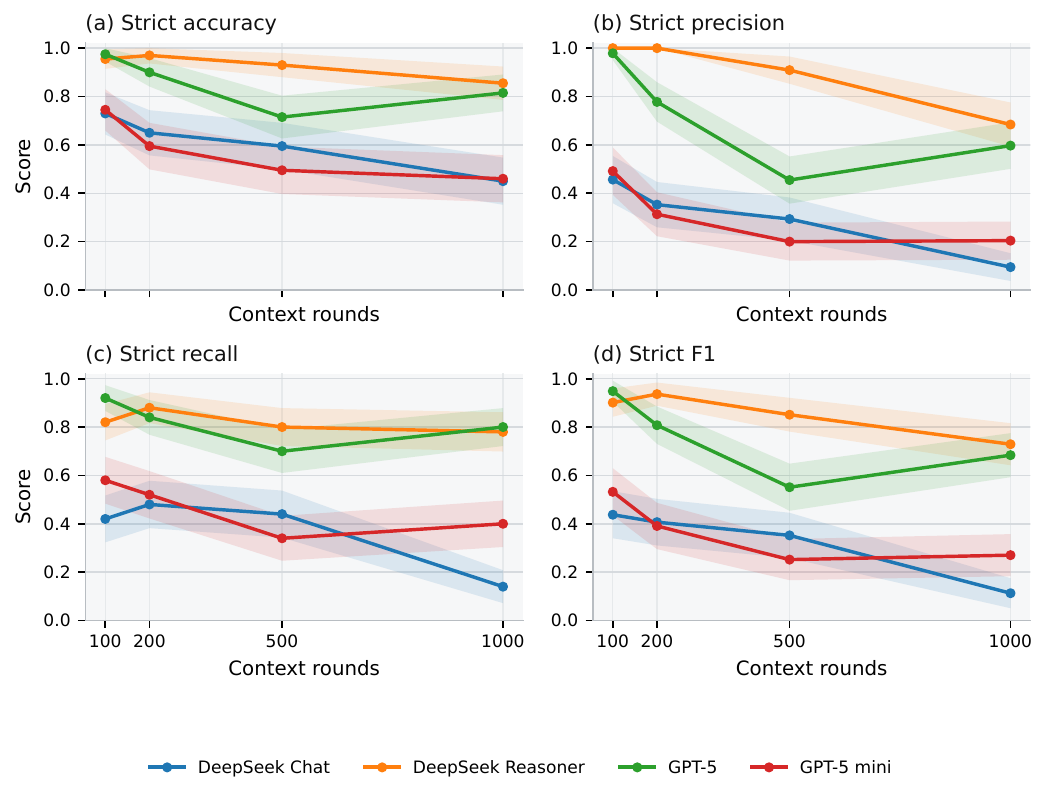}
    \caption{
    Markov-specific strict identification metrics across context lengths. 
    The four panels report strict accuracy, precision, recall, and F1 for Markov player identification. 
    Each curve corresponds to one model, and the x-axis indicates the number of context rounds.
    }
    \label{fig:markov_strict_metrics}
\end{figure}

We further report Markov-specific strict identification metrics in Figure~\ref{fig:markov_strict_metrics}. 
Strict accuracy measures whether the Markovian player is identified correctly. 
Precision measures whether predicted Markov identities are reliable. 
Recall measures whether true Markov players are successfully detected. 
F1 summarizes the balance between precision and recall. 
These metrics separate different failure modes. 
For example, low recall indicates that the model often misses true Markov players, while low precision indicates that the model often predicts Markov identities when the target behavior is not actually Markovian.

The strict Markov results reveal clear model-dependent differences. 
DeepSeek Chat is the least robust under longer contexts. 
Its Markov recall is 0.42 at 100 rounds, 0.48 at 200 rounds, 0.44 at 500 rounds, and drops sharply to 0.14 at 1000 rounds. 
Its strict F1 also decreases from 0.437 at 100 rounds to 0.113 at 1000 rounds, and its strict precision drops to 0.095 at 1000 rounds. 
This pattern indicates that DeepSeek Chat does not simply become less accurate overall. 
At long context length, it becomes much less reliable at detecting Markovian structure itself.

DeepSeek Reasoner behaves very differently. 
Its Markov recall remains high across all context lengths, with values of 0.82 at 100 rounds, 0.88 at 200 rounds, 0.80 at 500 rounds, and 0.78 at 1000 rounds. 
Its strict accuracy remains strong as well, decreasing from 0.955 at 100 rounds to 0.855 at 1000 rounds. 
Although its strict precision decreases at the longest context length, from 1.000 at both 100 and 200 rounds to 0.684 at 1000 rounds, its recall remains comparatively stable. 
This suggests that DeepSeek Reasoner is still able to detect most true Markov players even under longer histories, although its Markov predictions become less clean at the longest context.

GPT-5 shows strong Markov identification at short contexts but a non-monotonic trend as context length increases. 
Its Markov recall is 0.92 at 100 rounds and 0.84 at 200 rounds, then decreases to 0.70 at 500 rounds before recovering to 0.80 at 1000 rounds. 
This indicates that GPT-5 remains comparatively robust, but additional context does not produce a consistent improvement. 
Instead, longer histories can still introduce enough noise to weaken Markov detection at intermediate context lengths.

GPT-5 mini is weaker and less stable than the larger or reasoning-oriented models. 
Its Markov recall is 0.58 at 100 rounds, 0.52 at 200 rounds, 0.34 at 500 rounds, and 0.40 at 1000 rounds. 
Compared with DeepSeek Reasoner and GPT-5, GPT-5 mini has more difficulty detecting transition-dependent structure, especially at longer context lengths. 
This supports the observation that smaller models are more likely to rely on surface-level frequency patterns and less able to recover the underlying Markov rule from long trajectories.

The confusion matrices and Markov-specific metrics support the same conclusion from two angles. 
The confusion matrices show that errors often occur between behaviorally similar identities, especially when frequency-level evidence is insufficient to distinguish the true strategy. 
The strict Markov metrics show that the harder part of the task is not only choosing the exact identity, but first detecting whether the behavior is governed by a Markovian transition rule at all. 
Thus, the Markov difficulty observed in the main results reflects a real structural inference problem, not only a drop in aggregate accuracy.
% =========================
% Appendix D
% =========================
\section{Forecasting and Distributional Prediction Metrics}
\label{app:forecasting_distributional_metrics}

In addition to identity-level accuracy, we evaluate forecasting-oriented metrics to examine whether models preserve useful predictive information about future behavior. 
These metrics are computed from the model's predicted future action counts and the ground-truth player behavior. 
They complement identification accuracy by measuring whether the model can recover useful distributional and decision-level information, even when the final identity prediction is imperfect. 
All three metrics are error metrics, so lower values indicate better forecasting quality.

We first compute total variation distance for players whose ground-truth behavior has a well-defined static action distribution. 
This metric is applied only to non-Markovian/static distribution players, since Markovian players do not have a single fixed action distribution independent of history. 
Let $q$ denote the ground-truth action distribution over Rock, Paper, and Scissors, and let $p$ denote the model-predicted action distribution obtained by normalizing the generated action counts. 
We compute
\[
\mathrm{TV}(q,p) = \frac{1}{2}\sum_{c \in \{\mathrm{R},\mathrm{P},\mathrm{S}\}} |q_c - p_c|.
\]
This value lies in $[0,1]$, where lower values indicate that the predicted action distribution is closer to the true player distribution.

We also compute Brier score over the induced outcome distribution. 
Unlike total variation distance, this metric is not computed directly over Rock/Paper/Scissors actions. 
Instead, we first convert the ground-truth and predicted action distributions of the two players into outcome distributions over win, draw, and loss. 
Let $p^* = (P_{\mathrm{win}}^{*}, P_{\mathrm{draw}}^{*}, P_{\mathrm{loss}}^{*})$ be the ground-truth outcome distribution and let $\hat{p} = (\hat{P}_{\mathrm{win}}, \hat{P}_{\mathrm{draw}}, \hat{P}_{\mathrm{loss}})$ be the predicted outcome distribution. 
We compute
\[
\mathrm{Brier}(p^*,\hat{p}) =
\sum_{o \in \{\mathrm{win},\mathrm{draw},\mathrm{loss}\}}
(\hat{p}_o - p^*_o)^2.
\]
Lower Brier score indicates that the model's predicted outcome probabilities are closer to the true outcome probabilities.

Finally, we compute expected value loss to measure downstream decision-level error. 
For an outcome distribution $p$, we define its expected value as
\[
\mathrm{EV}(p) = P_{\mathrm{win}} - P_{\mathrm{loss}}.
\]
We then compute the squared difference between the ground-truth and predicted expected values:
\[
\mathrm{EVLoss}(p^*,\hat{p}) =
\left(\mathrm{EV}(p^*) - \mathrm{EV}(\hat{p})\right)^2.
\]
This metric captures whether forecasting errors translate into incorrect estimates of the strategic value of the interaction. 
Lower EVLoss indicates that the model better preserves the win-loss utility implied by the true behavior.

\begin{figure*}[!tbp]
    \centering
    \includegraphics[width=0.78\textwidth]{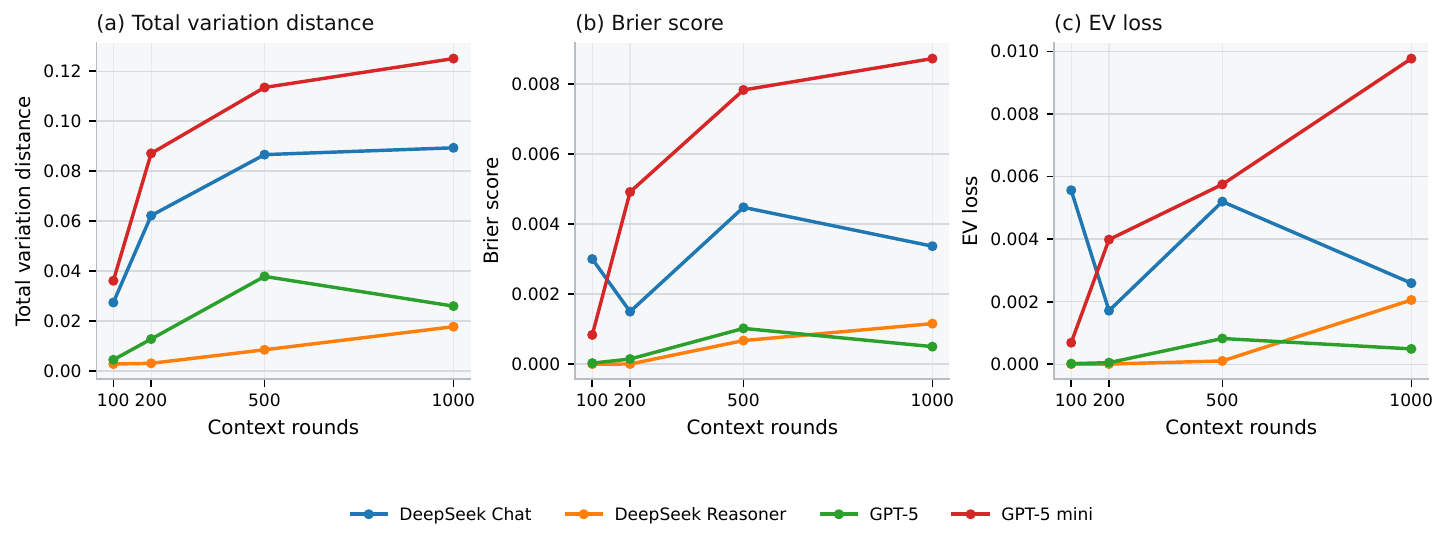}
    \caption{
    Forecasting and distributional prediction metrics across context lengths. 
    The panels report total variation distance, Brier score, and expected value loss. 
    Each point shows the mean metric value for one model under one context-length condition. 
    Each curve corresponds to one model, and the x-axis indicates the number of context rounds.
    }
    \label{fig:forecasting_metrics_3panel}
\end{figure*}

Figure~\ref{fig:forecasting_metrics_3panel} reports the average total variation distance, Brier score, and expected value loss for each model and context length. 
Each point corresponds to the mean metric value over all valid examples for that model and round condition. 
The results show that degradation under longer contexts is not limited to final identity selection. 
For weaker models, especially DeepSeek Chat and GPT-5 mini, longer contexts can also correspond to larger distributional mismatch or higher prediction loss. 
In contrast, stronger or reasoning-oriented models such as DeepSeek Reasoner and GPT-5 remain more stable across these forecasting metrics. 
This supports the broader conclusion that long-context difficulty appears both in identity recognition and in preserving useful distributional information for future prediction.

% =========================
% Appendix E
% =========================
\section{Additional Distributional Diagnostics}
\label{app:distributional_diagnostics}

This appendix provides additional distributional diagnostics for the simulation experiments in Section~\ref{sec:simulation}. The main text focuses on the MSE gap between Markov and non-Markov players. Here, we verify whether the same qualitative pattern appears under CE, inspect the raw CE/MSE values behind the relative gaps, and examine whether distributional difficulty is concentrated in specific strategies.

\begin{figure}[t]
    \centering
    \includegraphics[width=\linewidth]{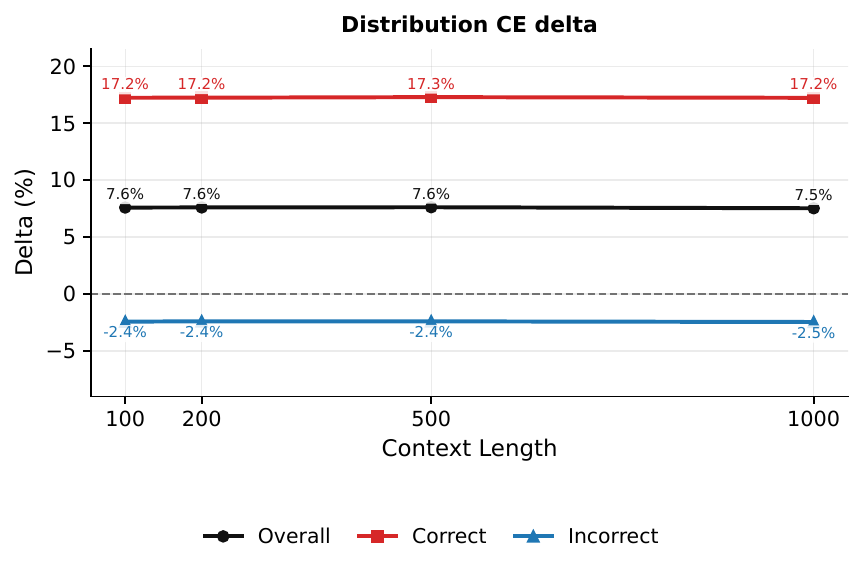}
    \caption{
    Relative CE gap between Markov and non-Markov players. Positive values indicate higher Markov error. The CE-based result follows the same qualitative pattern as the MSE-based result in Figure~\ref{fig:mse_delta}.
    }
    \label{fig:app_distribution_ce_delta}
\end{figure}

Figure~\ref{fig:app_distribution_ce_delta} reports the CE counterpart of the MSE delta in Figure~\ref{fig:mse_delta}. The pattern is consistent with the main MSE result: Markov players have higher error in the overall and correct-identity conditions, while the incorrect-identity condition shows a smaller or negative gap. This supports the claim that Markov simulation is distributionally harder when the model is expected to reproduce the exact target behavior.

\section{Additional Identification Diagnostics}
\label{app:identification_diagnostics}

This appendix provides additional recognition-level diagnostics for the simulation experiments. These analyses clarify whether model errors arise from confusing the broad Markov/non-Markov player families or from finer-grained mistakes among specific player identities. This distinction matters because the main text argues that exact identity recognition and rule-level simulation are related but not equivalent.

\begin{figure*}[t]
    \centering
    \includegraphics[width=0.88\textwidth]{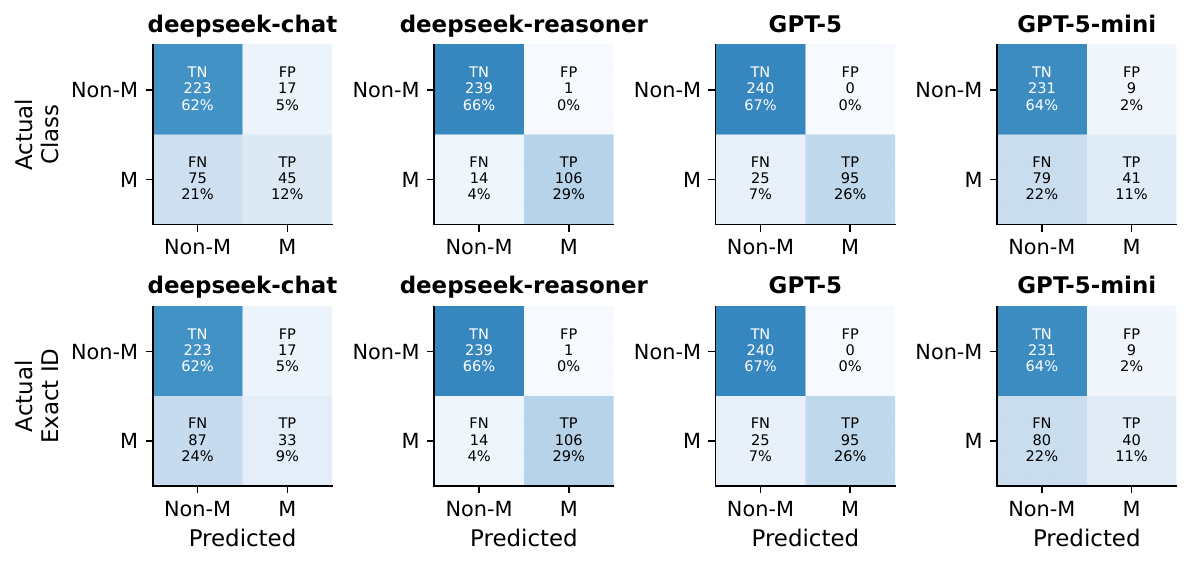}
    \caption{
    Binary Markov/non-Markov confusion matrices. These diagnostics separate broad class recognition from exact identity recognition and show whether model errors come from confusing the two player families or from finer-grained identity mistakes.
    }
    \label{fig:app_binary_confusion}
\end{figure*}

Figure~\ref{fig:app_binary_confusion} reports binary Markov/non-Markov confusion matrices. These matrices show whether each model recognizes the broad player family. The results help separate broad class recognition from exact identity recognition, which is useful because a model may correctly identify the family while still confusing specific players within that family.

\begin{figure*}[t]
    \centering
    \includegraphics[width=0.88\textwidth]{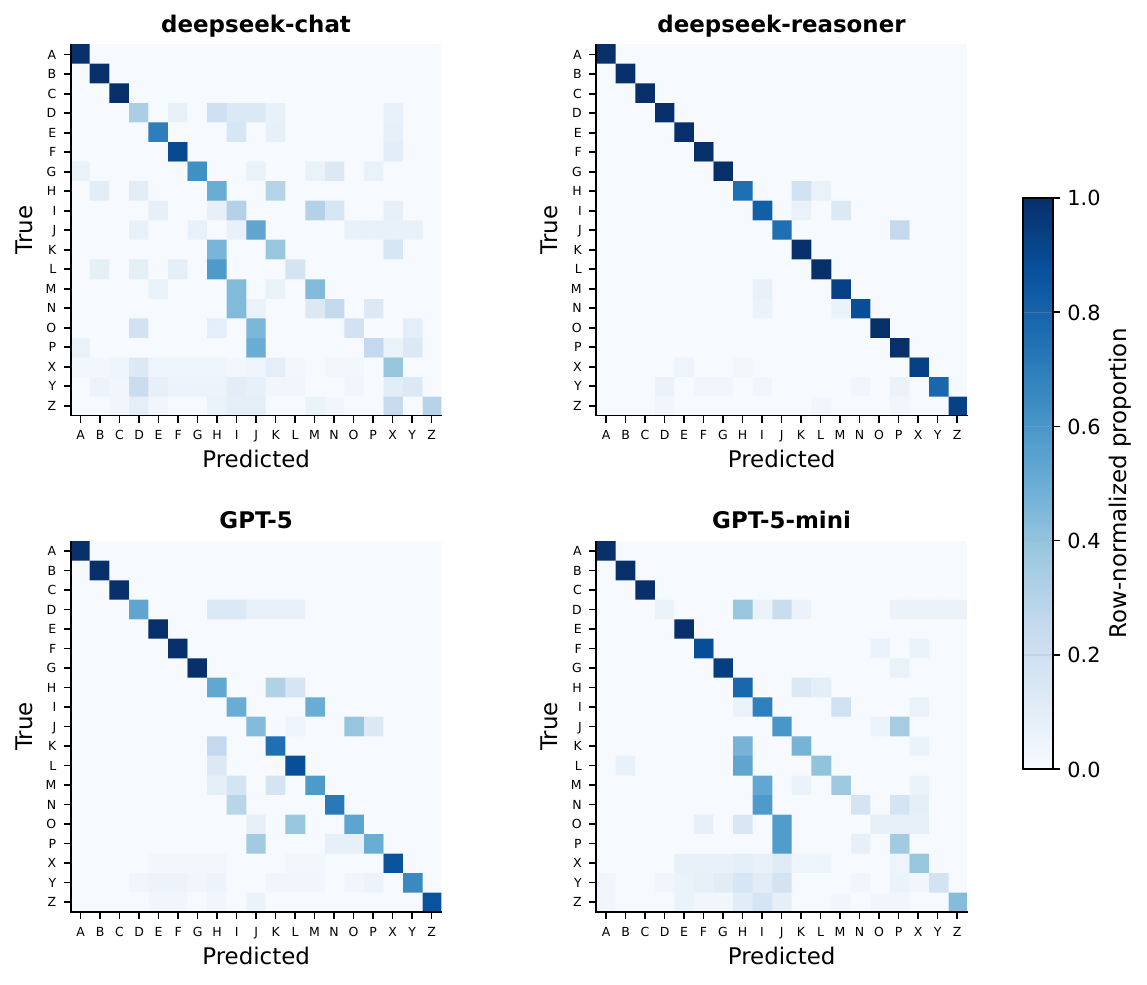}
    \caption{
    Full identity-level confusion matrices. 
    Rows correspond to true player identities and columns correspond to predicted identities. 
    Off-diagonal structure indicates that identity errors are not uniformly random, but often occur among behaviorally similar players.
    }
    \label{fig:app_full_identity_confusion}
\end{figure*}

Figure~\ref{fig:app_full_identity_confusion} reports full identity-level confusion matrices. Unlike the binary matrices, these show which specific player identities are confused with one another. Off-diagonal structure indicates that identity errors are not uniformly random, but often concentrate among behaviorally similar players. This supports the main-text observation that wrong identity predictions can still lead to non-trivial rule-following behavior.

\section{Complex Markov Sets}
\label{app:complex_markov_sets}

Table~\ref{tab:complex_markov_sets_app} summarizes the Markov sets used in the complex input-structure analysis. These sets vary along two factors: dependency order and whether the rule conditions on opponent-only or joint-state information.

\begin{table}[t]
\centering
\small
\begin{tabular}{lcc}
\hline
\textbf{Set} & \textbf{Order} & \textbf{State representation} \\
\hline
XYZ & First-order & Opponent-only \\
QRS & First-order & Joint-state \\
TUV & Second-order & Joint-state \\
XYZ-opp & Second-order & Opponent-only \\
\hline
\end{tabular}
\caption{Complex Markov sets used to stress test conditional generation. The sets separate the effect of joint-state representation from the effect of higher-order historical dependency.}
\label{tab:complex_markov_sets_app}
\end{table}
% =========================
% Appendix H
% =========================
\section{Complex Markov Identification Diagnostics}
\label{app:complex_identification_diagnostics}
\label{app:complex_markov_diagnostics}

This appendix provides diagnostic evidence for whether errors come from class-level Markov recognition or exact player attribution. These diagnostics complement the main results in Section~\ref{sec:complex_input_structures}, where identity accuracy remains high even when strict rule following decreases.

\begin{figure*}[t] \centering \includegraphics[width=0.95\textwidth]{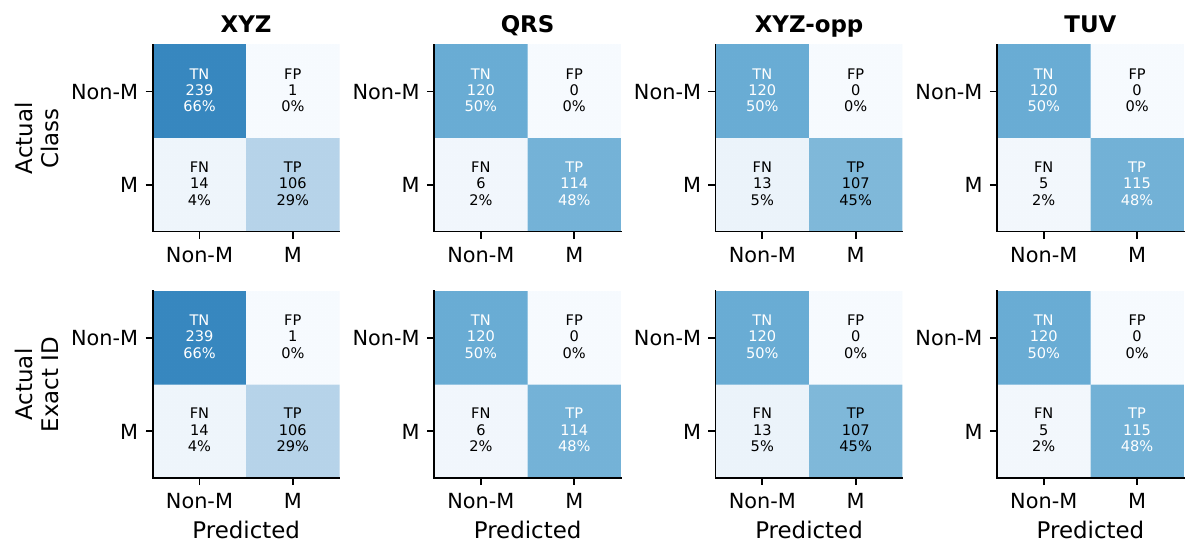} \caption{Binary Markov versus non-Markov confusion matrices for the complex Markov sets. The matrices provide a class-level diagnostic of whether the model identifies Markovian behavior.} \label{fig:app_complex_binary_confusion} \end{figure*}

Figure~\ref{fig:app_complex_binary_confusion} reports binary Markov versus non-Markov confusion matrices. These matrices examine whether the model separates Markovian and non-Markovian players at the class level. The errors are concentrated more in missed Markov cases than in false Markov predictions, which suggests that failures in the main text are not primarily caused by the model broadly confusing non-Markovian behavior with Markovian behavior.

\begin{figure*}[t]
\centering
\includegraphics[width=0.95\textwidth]{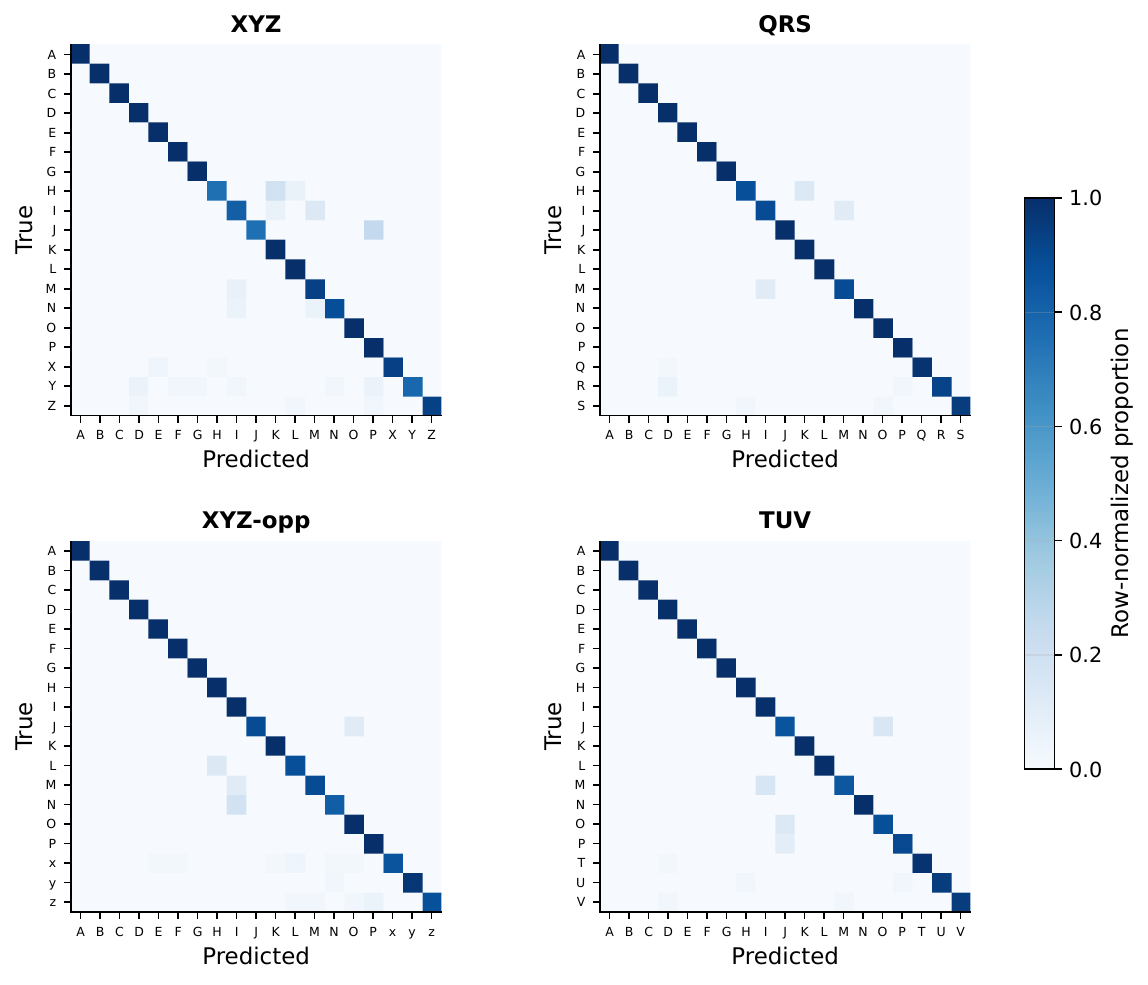}
\caption{Full identity confusion matrices for the complex Markov sets. These matrices show exact player-level attribution errors and complement the binary Markov versus non-Markov view.}
\label{fig:app_complex_full_identity_confusion}
\end{figure*}

Figure~\ref{fig:app_complex_full_identity_confusion} reports full identity confusion matrices. This diagnostic is more fine-grained than binary classification because it asks whether the model selects the exact player identity. These results help distinguish player-level attribution errors from rule-execution errors. A wrong identity does not necessarily imply that the generated trajectory violates the target rule; it may indicate that the model recovered the behavioral rule but assigned it to the wrong player label.

% =========================
% Appendix I
% =========================
\section{Complex Markov Distributional and Temporal Diagnostics}
\label{app:complex_distribution_temporal}

This appendix examines whether stable strict-rule behavior is compatible with trajectory-level distributional drift. Strict rule match evaluates whether the correct transition rule is followed, while CE and MSE evaluate whether the marginal action distribution of the generated trajectory remains close to the context distribution.

% Figure~\ref{fig:app_distribution_diagnostics} reports distributional CE and MSE across the four complex Markov sets. These metrics capture a different aspect of generation from strict rule match, because they measure trajectory-level distributional fidelity rather than exact transition-rule correctness.

% \begin{figure*}[t]
% \centering
% \includegraphics[width=0.85\textwidth]{latex/figures/paper_plots_exp3/appendix/pdf/figA3_exp2_exp3_distribution_diagnostics.pdf}
% \caption{Distributional diagnostics across complex Markov sets. CE and MSE measure trajectory-level distributional fidelity rather than strict transition-rule correctness.}
% \label{fig:app_distribution_diagnostics}
% \end{figure*}

\begin{figure*}[t]
\centering
\includegraphics[width=0.9\textwidth]{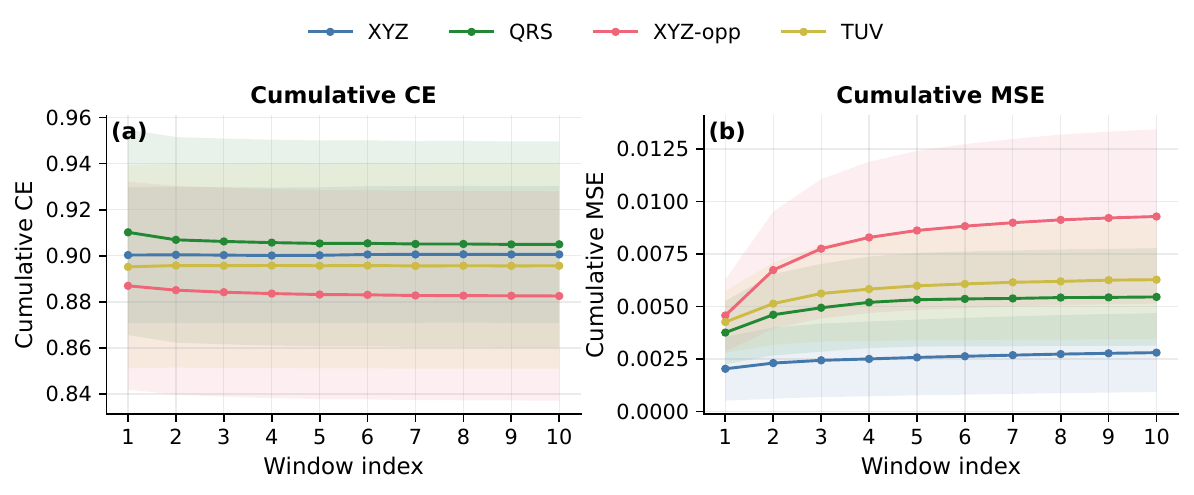}
\caption{Cumulative CE and MSE over generated windows. Distributional drift can reflect changes in state visitation or trajectory-level marginal distributions, even when strict rule following remains stable.}
\label{fig:app_cumulative_distribution}
\end{figure*}

Figure~\ref{fig:app_cumulative_distribution} shows cumulative distributional dynamics across generated windows. The cumulative MSE can increase even when strict rule match remains stable, because marginal action distributions depend not only on the transition rule but also on the states visited by the generated trajectory. If the generated trajectory visits states at different frequencies from the context trajectory, the marginal distribution can drift while the transition rule remains locally valid.

\begin{figure*}[t]
\centering
\begin{minipage}[t]{0.49\textwidth}
\centering
\includegraphics[width=\linewidth]{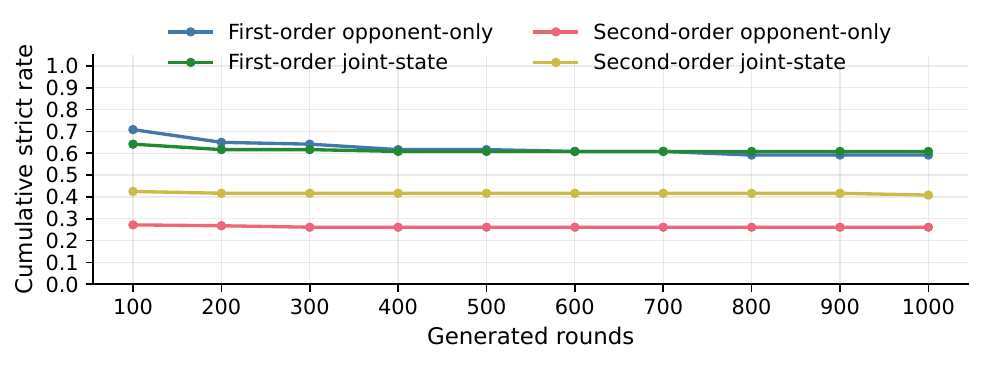}
\smallskip
\textbf{(a)} Cumulative strict-rule dynamics.
\end{minipage}
\hfill
\begin{minipage}[t]{0.47\textwidth}
\centering
\includegraphics[width=\linewidth]{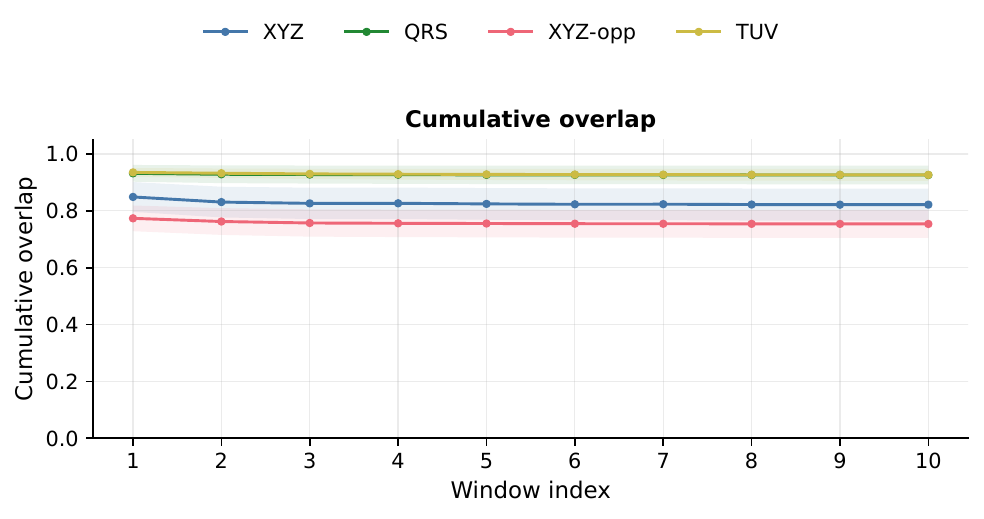}
\smallskip
\textbf{(b)} Cumulative rule-overlap dynamics.
\end{minipage}
\caption{
Cumulative generation dynamics over generated windows. 
Panel (a) reports cumulative strict-rule dynamics, showing that rule-following behavior is largely determined early in the generated trajectory. 
Panel (b) reports cumulative rule-overlap dynamics, which captures local rule-component similarity and should be interpreted separately from strict rule execution.
}
\label{fig:app_cumulative_dynamics}
\end{figure*}

Figure~\ref{fig:app_cumulative_dynamics} reports two complementary cumulative diagnostics over generated windows. 
Panel (a) shows cumulative strict-rule dynamics, where the curves become relatively stable after the early windows. 
This suggests that the model often enters a stable rule-following or rule-approximation regime early in generation, and that many failures are caused more by the initial abstraction of the rule than by gradual collapse in later rounds. 
Panel (b) shows cumulative rule-overlap dynamics, which complements strict-rule behavior by measuring whether the generated sequence continues to share local rule components with the target rule. 
The distinction is important because overlap can remain high even when exact rule execution fails.

% =========================
% Appendix J
% =========================
\section{Complex Markov Identity-Conditioned and Rule-Level Diagnostics}
\label{app:complex_identity_rule_diagnostics}

This appendix separates player-level attribution errors from rule-execution errors and examines whether high local overlap implies exact rule execution. These diagnostics explain why identity-wrong cases can still produce nonzero strict rates and why overlap can remain high even when strict match is lower.

% Figure~\ref{fig:app_identity_condition_strict} reports cumulative strict rates conditioned on whether the model selected the correct identity. Identity-correct cases usually achieve higher strict rates, but identity-wrong cases can still have nonzero strict rates. This occurs because identity selection and rule execution are not equivalent. The model can assign the behavior to the wrong player label while still generating a trajectory that follows the target transition rule. In this case, the error is primarily player-level attribution rather than behavioral rule execution.

% \begin{figure*}[t]
% \centering
% \includegraphics[width=0.9\textwidth]{latex/figures/paper_plots_exp3/appendix/pdf/figA5_exp2_exp3_identity_condition_strict.pdf}
% \caption{Identity-conditioned cumulative strict rates. Identity errors do not always imply rule-execution failure, because the model may follow the correct behavioral rule while assigning it to the wrong player identity.}
% \label{fig:app_identity_condition_strict}
% \end{figure*}

\begin{figure*}[t]
\centering
\includegraphics[width=0.95\textwidth]{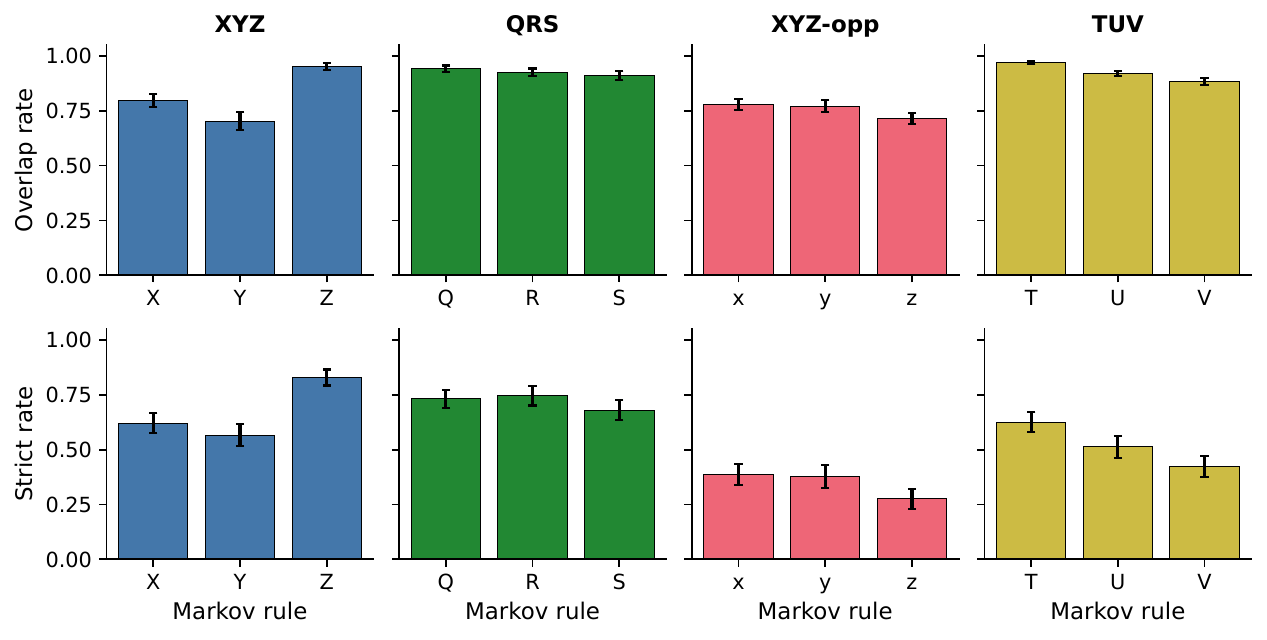}
\caption{Rule-level overlap and strict-match breakdown. The comparison shows that local rule overlap can remain high even when exact rule execution is substantially lower.}
\label{fig:app_rule_breakdown}
\end{figure*}

Figure~\ref{fig:app_rule_breakdown} provides a rule-level breakdown of overlap and strict match. This diagnostic shows that difficulty can vary within the same Markov set. Some rules preserve high overlap but lower strict match, indicating that the model captures local transition components more easily than it executes the exact rule across all required states or histories.

\section{Statistical Significance Test for Recognition and Simulation}
\label{app:recognition_simulation_significance}

To examine whether the recognition accuracy differences between Exp1 and Exp2 are statistically reliable, we conduct a two-proportion z-test for each model. Because Exp1 and Exp2 contain different numbers of samples, with $n_1=100$ and $n_2=120$, the two settings cannot be treated as paired observations. We therefore use an independent two-proportion z-test.

For each model, let $\hat{p}_1$ denote the overall recognition accuracy in Exp1 and $\hat{p}_2$ denote the overall recognition accuracy in Exp2. The pooled proportion is computed as
\[
\hat{p}
=
\frac{n_1\hat{p}_1+n_2\hat{p}_2}{n_1+n_2}.
\]
The test statistic is
\[
z
=
\frac{\hat{p}_1-\hat{p}_2}
{
\sqrt{
\hat{p}(1-\hat{p})
\left(
\frac{1}{n_1}+\frac{1}{n_2}
\right)
}
}.
\]

Table~\ref{tab:recognition_simulation_ztest} reports the resulting statistics. Across all four models, none of the Exp1--Exp2 differences are statistically significant at the conventional 0.05 level. GPT-5 shows the largest difference between Exp1 and Exp2, but this difference remains only marginal and does not reach statistical significance. Therefore, while Figure~\ref{fig:recognition_vs_simulation} shows model-dependent numerical changes, the statistical test suggests that these changes should be interpreted cautiously.

\begin{table*}[t]
\centering
\small
\begin{tabular}{lccccc}
\toprule
Model & Exp1 Acc. & Exp2 Acc. & $z$ & $p$ & Significant? \\
\midrule
DeepSeek-Chat     & 0.21 & 0.18 &  0.56 & 0.576 & No \\
DeepSeek-Reasoner & 0.75 & 0.78 & -0.53 & 0.600 & No \\
GPT-5             & 0.67 & 0.56 &  1.67 & 0.096 & No \\
GPT-5-mini        & 0.23 & 0.26 & -0.52 & 0.607 & No \\
\bottomrule
\end{tabular}
\caption{Two-proportion z-test comparing recognition accuracy between Exp1 and Exp2.}
\label{tab:recognition_simulation_ztest}
\end{table*}
% =========================
% Appendix K
% =========================
% \section{One-Player Full Metric Grid}
% \label{app:one_player_diagnostics}
% \label{app:one_player_full_metric_grid}
%The main text focuses on context-conditioned likelihood gain (CCLG) and state-frequency-weighted Jensen--Shannon divergence (Weighted JS). Here we provide the full metric grid, raw likelihood diagnostics, entropy-shape diagnostics, validity checks, degeneration diagnostics, and detailed breakdowns of prefix-length and rule-access effects.

% Figure~\ref{fig:app_one_player_full_metric_grid} reports the full metric grid across models, prompting settings, prefix lengths, and n-gram orders. This figure expands the main-text results by showing CE, CCLG, Weighted JS, entropy gap, and valid completion rate in a single view. The grid is included as a reference for checking whether the main trends hold across the full set of evaluated conditions.

% \begin{figure}[t]
% \centering
% \includegraphics[width=\linewidth]{figures/paper_plots_followup/appendix/pdf/figA2_full_metric_grid.pdf}
% \caption{
% Full metric grid across models, prompting settings, prefix lengths, and n-gram orders. 
% This diagnostic provides a complete view of generation quality, transition recovery, distribution-shape mismatch, and validity.
% }
% \label{fig:app_one_player_full_metric_grid}
% \end{figure}

% =========================
% Appendix L
% =========================
\section{One-Player Raw Likelihood and Stochasticity Diagnostics}
\label{app:one_player_likelihood_recovery}

This appendix provides the full definitions of the likelihood-based metrics
and additional diagnostics used for the one-player n-gram continuation
experiment in Section~\ref{sec:one_player_ngram}.

Let $\hat{x}_{1:M}$ denote a generated continuation of length $M$.
For each generated position $t$, let $s_t$ denote the preceding $n$-gram
history, and let $q(\cdot \mid s_t)$ denote the true next-token distribution.

\paragraph{Context-conditioned likelihood gain.}
Context-conditioned likelihood gain (CCLG) measures how much the
history-conditioned transition model improves the likelihood of the generated
continuation relative to a unigram baseline. We compute

\begin{equation}
\label{eq:app_cclg}
\mathrm{CCLG}
=
\frac{1}{M}
\sum_{t=1}^{M}
\left[
\log q(\hat{x}_t \mid s_t)
-
\log q_{\mathrm{uni}}(\hat{x}_t)
\right],
\end{equation}

where $q_{\mathrm{uni}}$ denotes the unigram baseline.
Positive CCLG indicates that the generated continuation is better explained
by the true context-conditioned process than by unigram frequencies alone.
Values near zero indicate little additional likelihood gain from conditioning
on history, while negative values indicate that the continuation is less
compatible with the conditional process than with the unigram baseline.

\paragraph{Cross-entropy.}
Cross-entropy (CE) measures the likelihood of the generated continuation
under the true n-gram process:

\begin{equation}
\label{eq:app_cross_entropy}
\mathrm{CE}
=
-\frac{1}{M}
\sum_{t=1}^{M}
\log q(\hat{x}_t \mid s_t).
\end{equation}

Lower CE indicates better generation quality under the true conditional
transition process. Whereas CCLG measures the benefit of conditioning on
history relative to unigram frequencies, CE measures absolute likelihood
under the true n-gram process.

\paragraph{Entropy gap.}
Entropy gap measures whether the generated transition distributions have a
similar degree of stochasticity to the true transition distributions.
For an n-gram history state $s$, let $q(\cdot\mid s)$ be the true next-token
distribution and let $\hat{q}(\cdot\mid s)$ be the empirical next-token
distribution induced by the generated continuation. We compute entropy gap as

\begin{equation}
\label{eq:app_entropy_gap}
\mathrm{EntGap}
=
\sum_s w_s
\left|
H\big(q(\cdot\mid s)\big)
-
H\big(\hat{q}(\cdot\mid s)\big)
\right|,
\end{equation}

where $w_s$ is the empirical frequency of state $s$ in the generated
continuation. Lower entropy gap indicates closer stochasticity to the true
n-gram process.

Together, these metrics capture complementary aspects of conditional
generation. CCLG measures context use beyond unigram frequencies, CE measures
raw likelihood under the true transition process, and entropy gap measures
whether the stochasticity of the generated conditional distributions matches
the ground-truth process.

\begin{figure}[!tbp]
    \centering
    \includegraphics[width=\linewidth]{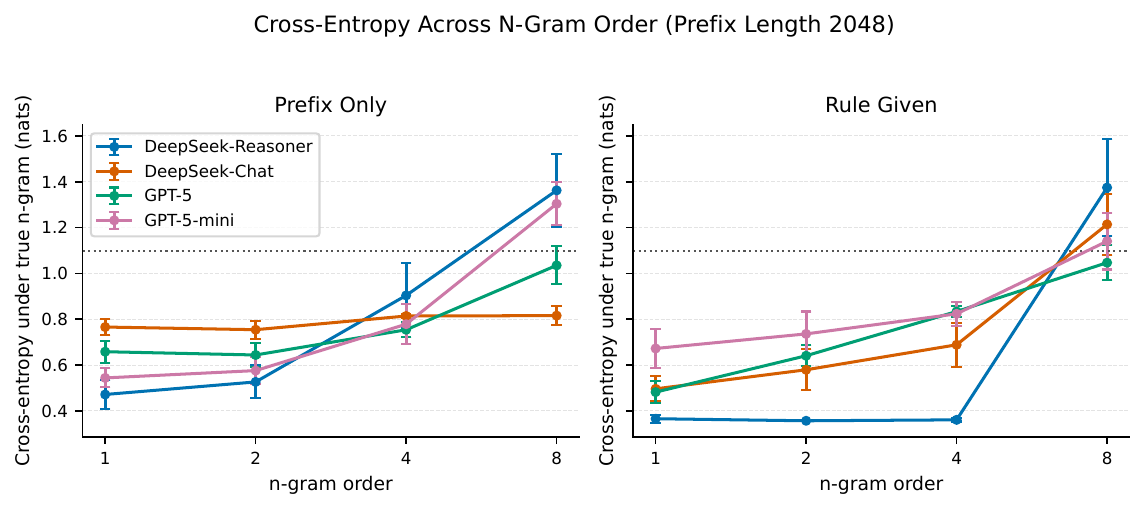}
    \caption{
    Cross-entropy as n-gram order increases, using prefix length 2048.
    Lower values indicate better generation quality under the true n-gram
    process.
    The results show that raw generation quality deteriorates at higher order,
    especially at $n=8$.
    }
    \label{fig:app_one_player_generation_quality}
\end{figure}

\begin{figure}[!tbp]
    \centering
    \includegraphics[width=\linewidth]{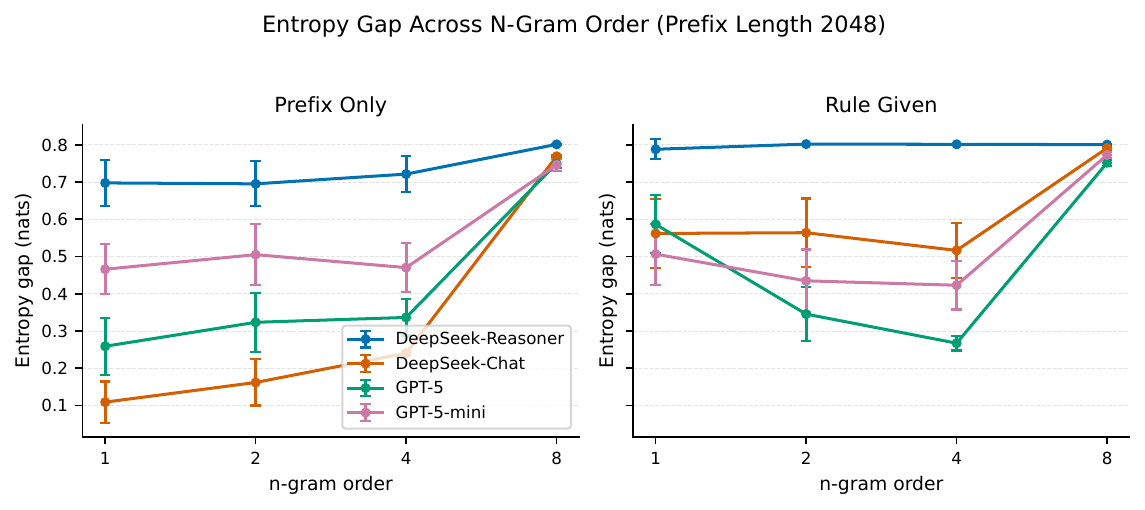}
    \caption{
    Entropy gap as n-gram order increases, using prefix length 2048.
    Lower values indicate closer stochasticity to the true n-gram process.
    Higher-order processes lead to larger stochasticity mismatch.
    }
    \label{fig:app_one_player_recovery}
\end{figure}

Figure~\ref{fig:app_one_player_generation_quality} reports CE across n-gram
orders, and Figure~\ref{fig:app_one_player_recovery} reports entropy gap.
Together, these figures show that higher-order processes reduce raw generation
quality and also increase stochasticity mismatch.

% =========================
% Appendix M
% =========================
\section{One-Player Validity and Coverage}
\label{app:one_player_validity}

\begin{figure}[t]
\centering
\includegraphics[width=\linewidth]{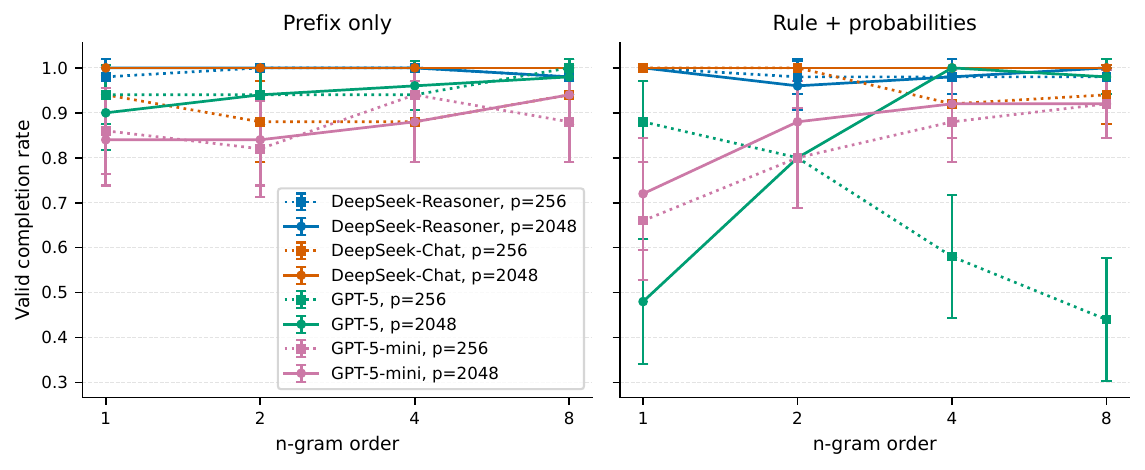}
\caption{
Valid completion rates across n-gram orders, prompting settings, models, and prefix lengths. 
This figure serves as a sanity check for whether metric differences are driven by invalid completions.
}
\label{fig:app_one_player_validity_coverage}
\end{figure}

Figure~\ref{fig:app_one_player_validity_coverage} reports valid completion rates across n-gram orders. This diagnostic checks whether the reported generation and recovery metrics are affected by invalid outputs. Most conditions maintain high valid completion rates, although some model-setting combinations show lower validity under the rule-and-probs setting. These cases are useful for interpreting whether poor metric values reflect rule-following failure or output-format failure.

% =========================
% Appendix N
% =========================
\section{One-Player Degeneration Diagnostics}
\label{app:one_player_degeneracy}

\begin{figure}[t]
\centering
\includegraphics[width=\linewidth]{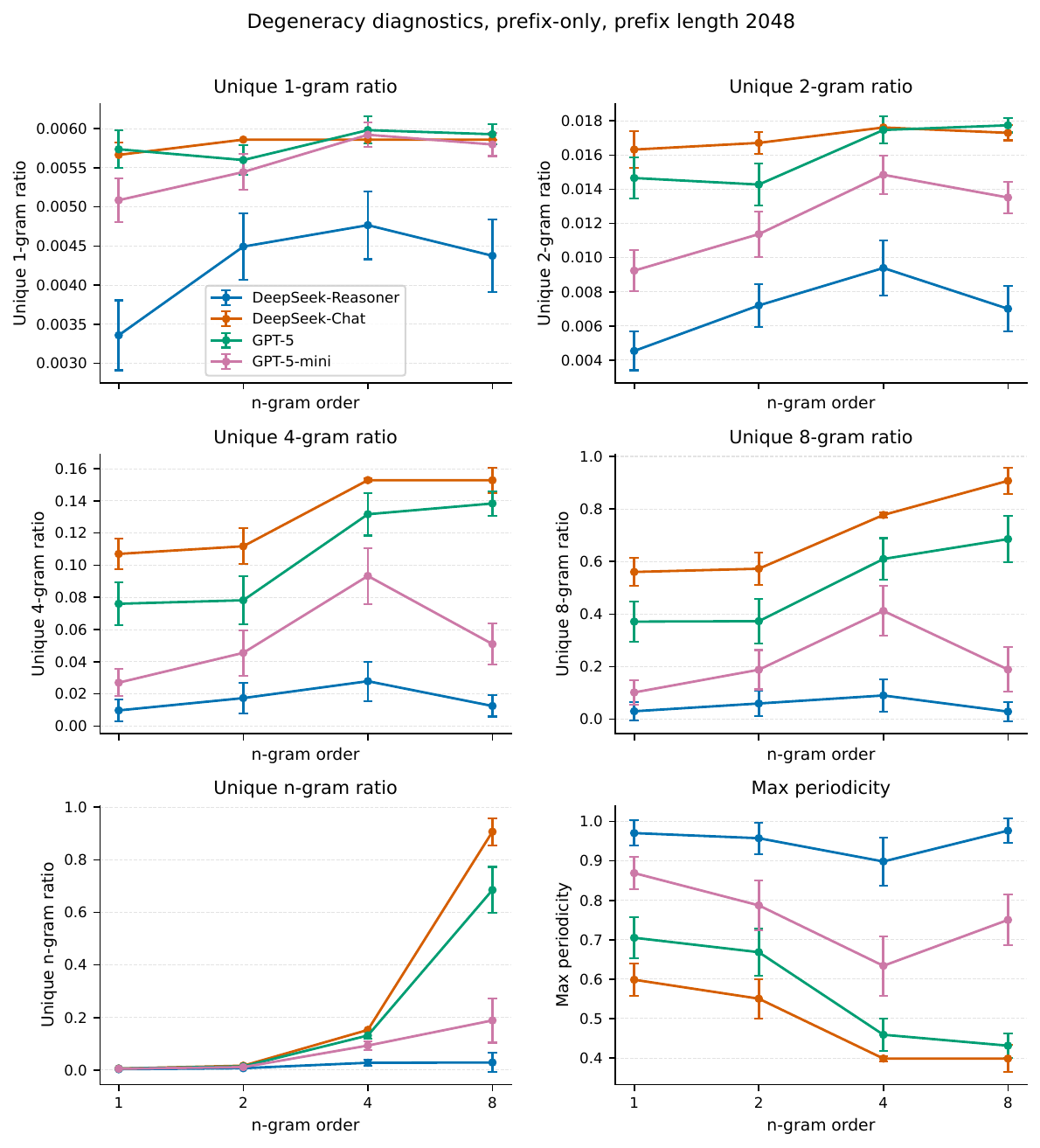}
\caption{
Degeneration diagnostics for the prefix-only setting with prefix length 2048. 
Unique n-gram ratios and maximum periodicity measure whether high-order generation leads to repetition, periodicity, or diversity collapse.
}
\label{fig:app_one_player_degeneracy_diagnostics}
\end{figure}

Figure~\ref{fig:app_one_player_degeneracy_diagnostics} reports degeneration diagnostics for the prefix-only setting with prefix length 2048. We include unique n-gram ratios and maximum periodicity to examine whether models collapse into repetitive or periodic continuations at higher orders. These diagnostics complement the likelihood and transition-recovery metrics by showing whether the generated sequences preserve diversity or instead become overly regular.

% =========================
% Appendix O
% =========================
\section{One-Player Rule-Access Effects}
\label{app:one_player_rule_access_effects}

\begin{figure}[t]
\centering
\includegraphics[width=\linewidth]{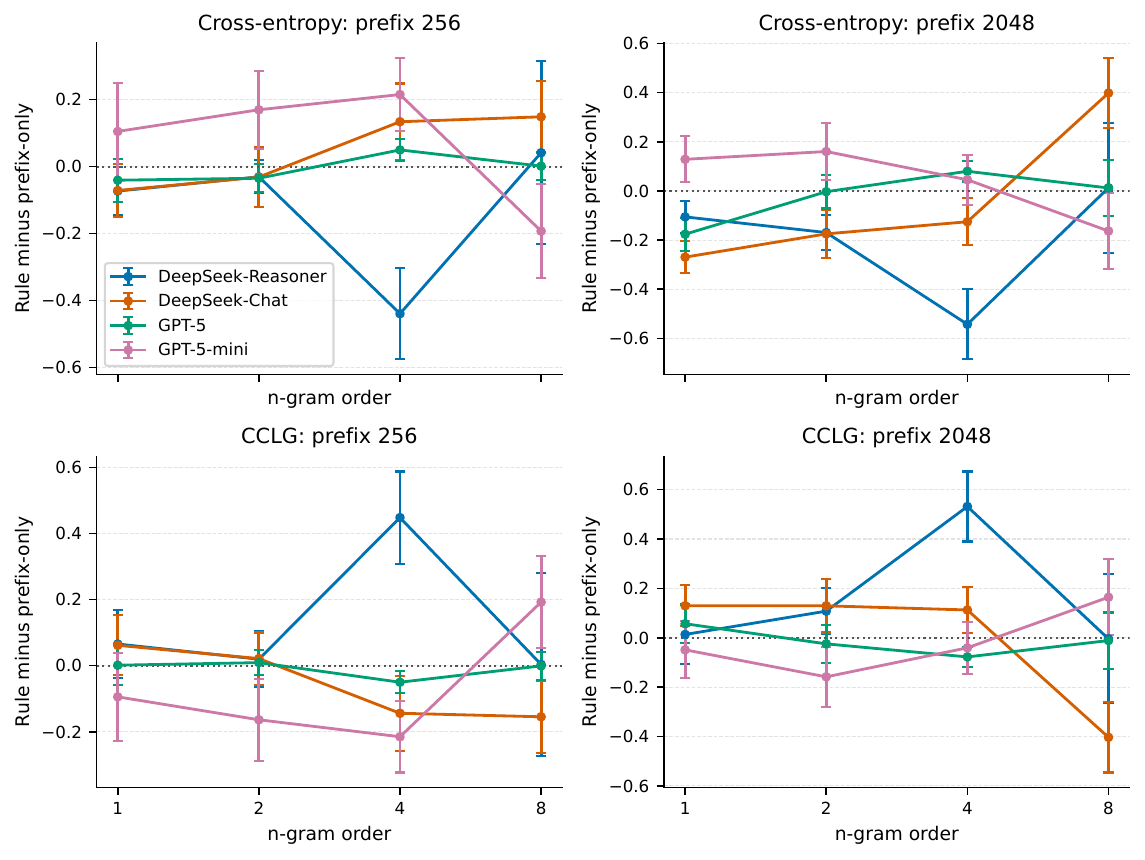}
\caption{
Detailed rule-access effects by model and prefix length. 
Values show rule-and-probs minus prefix-only differences. 
Explicit rule access does not consistently improve CE or CCLG, suggesting that high-order failure is not only a rule-inference problem.
}
\label{fig:app_one_player_rule_access_effects}
\end{figure}

Figure~\ref{fig:app_one_player_rule_access_effects} provides a detailed comparison between the prefix-only and rule-and-probs settings. The figure reports rule minus prefix-only differences for CE and CCLG under prefix lengths 256 and 2048. These results show that explicitly providing the transition rule and probabilities does not uniformly improve generation quality, indicating that failures can persist even when the rule is directly available.

% =========================
% Appendix P
% =========================
\section{One-Player Prefix-Length Effects}
\label{app:one_player_prefix_effects}

\begin{figure}[t]
\centering
\includegraphics[width=\linewidth]{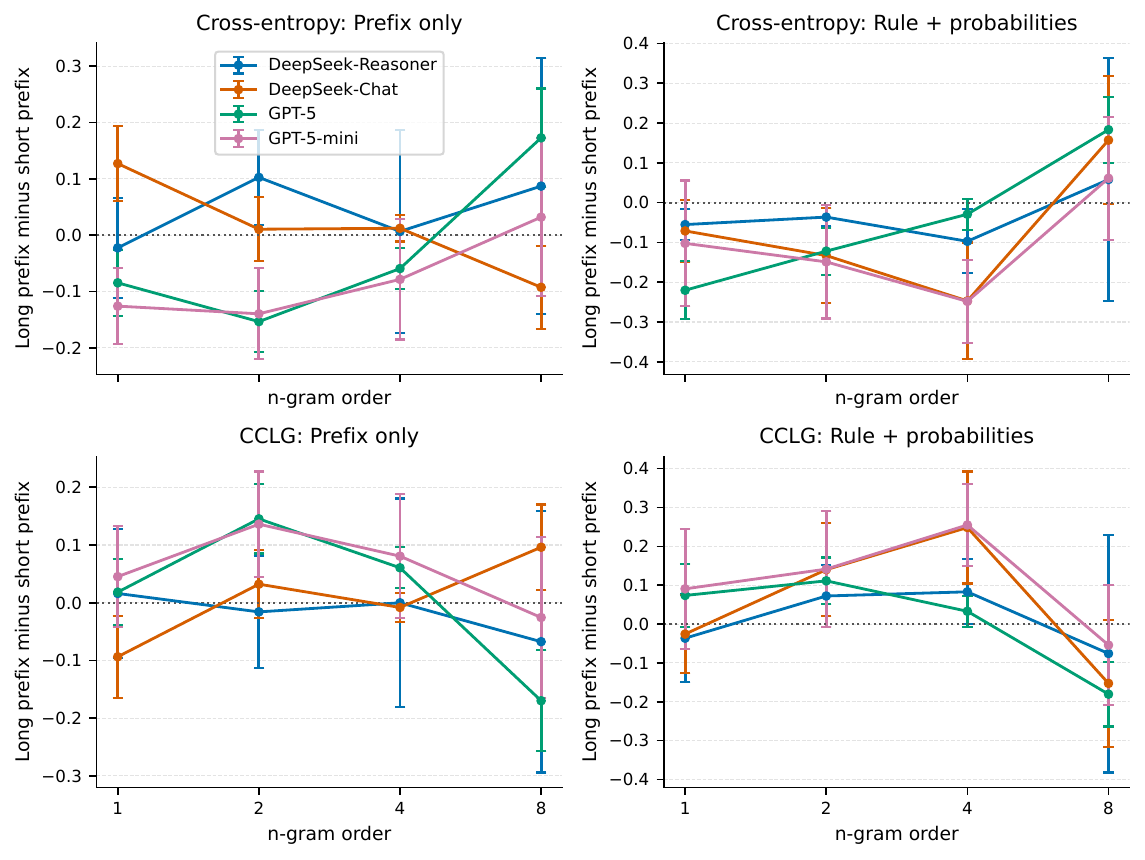}
\caption{
Detailed prefix-length effects by model. 
Values show long-prefix minus short-prefix differences. 
Longer prefixes do not consistently improve CE or CCLG, especially at higher n-gram orders.
}
\label{fig:app_one_player_prefix_length_effects}
\end{figure}

Figure~\ref{fig:app_one_player_prefix_length_effects} provides a detailed view of the effect of increasing prefix length from 256 to 2048 tokens. 
The figure reports long-prefix minus short-prefix differences for CE and CCLG under both prefix-only and rule-and-probs settings.

\paragraph{Longer prefixes do not consistently improve generation quality.}
The effect of increasing prefix length is mixed across models, prompting settings, and n-gram orders. 
Although longer prefixes can sometimes improve CE or CCLG, the improvement is not consistent, especially for higher-order processes. 
This suggests that high-order failure is not simply due to insufficient prefix evidence.

% =========================
% Appendix Q
% =========================
\section{One-Player Information and Context Effects}
\label{app:one_player_information_effects}

\begin{figure}[!tbp]
    \centering
    \includegraphics[width=\linewidth]{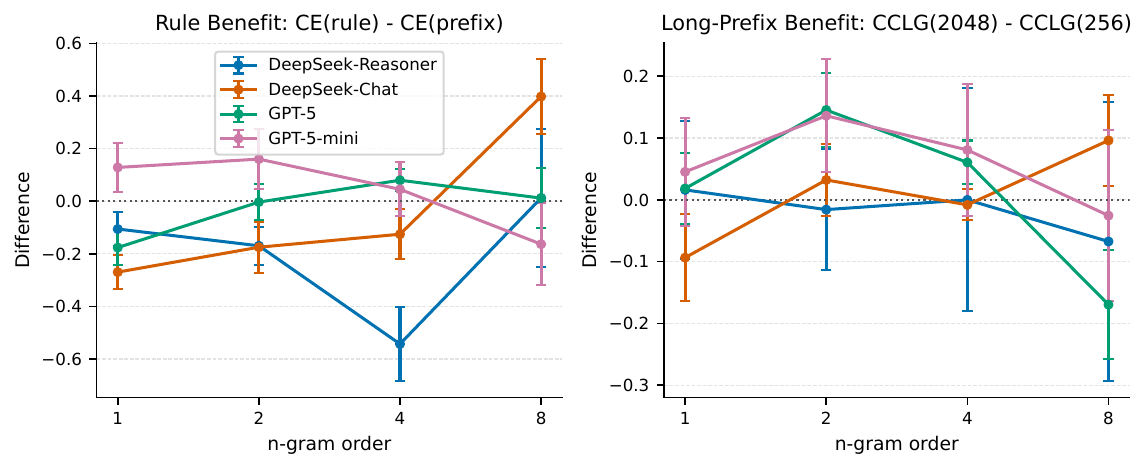}
    \caption{
    Effects of explicit rule access and longer prefix evidence. 
    Negative values in CE(rule) - CE(prefix) indicate that providing the rule lowers CE, while positive values in CCLG(2048) - CCLG(256) indicate benefit from a longer prefix. 
    The effects are inconsistent across models and n-gram orders, showing that explicit rules and longer prefixes do not reliably solve high-order dependency.
    }
    \label{fig:app_one_player_information_effects}
\end{figure}

Figure~\ref{fig:app_one_player_information_effects} jointly summarizes whether additional information resolves the high-order failure. 
The left panel compares the effect of explicitly providing transition rules and probabilities, while the right panel compares the effect of increasing prefix length from 256 to 2048 tokens. 
Negative values in CE(rule) - CE(prefix) indicate that rule access improves raw likelihood, while positive values in CCLG(2048) - CCLG(256) indicate that a longer prefix improves context-conditioned likelihood gain.

\paragraph{More information does not reliably solve high-order dependency failure.}
The effects of rule access and longer prefix evidence vary across models and n-gram orders. 
Explicit transition rules help in some settings but do not uniformly improve generation quality, and longer prefixes do not consistently increase context-conditioned likelihood gain. 
Thus, high-order failure cannot be explained solely by rule inference difficulty or insufficient evidence length.
%, but also reflects difficulty using and maintaining high-order transition structure during generation.

\section{One-Player Prefix-Only Structured Baseline}
\label{app:one_player_no_seed_mle}

\begin{figure}[!tbp]
    \centering
    \includegraphics[width=\linewidth]{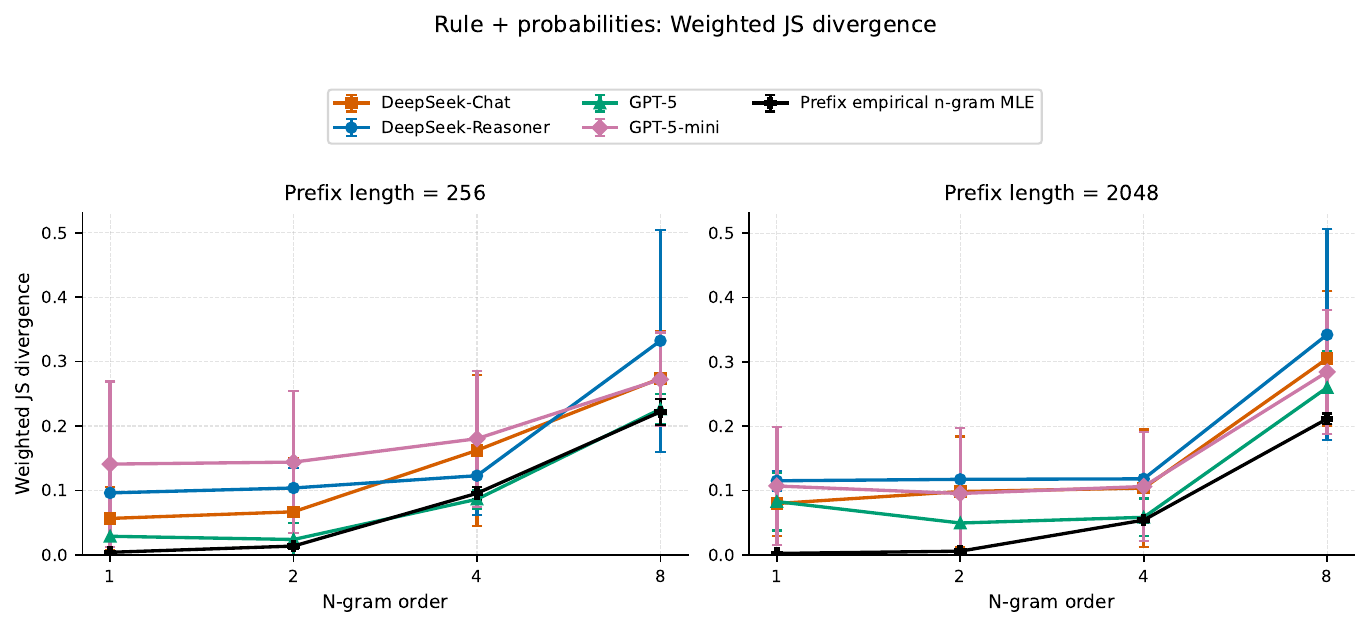}
    \caption{
    Prefix-only empirical n-gram MLE baseline under the rule-and-probabilities setting, evaluated by weighted JS divergence. 
    The baseline uses only the observed prefix and remains competitive with or better than LLM generations, especially at higher orders.
    }
    \label{fig:app_one_player_no_seed_mle_wjs}
\end{figure}

\begin{figure}[!tbp]
    \centering
    \includegraphics[width=\linewidth]{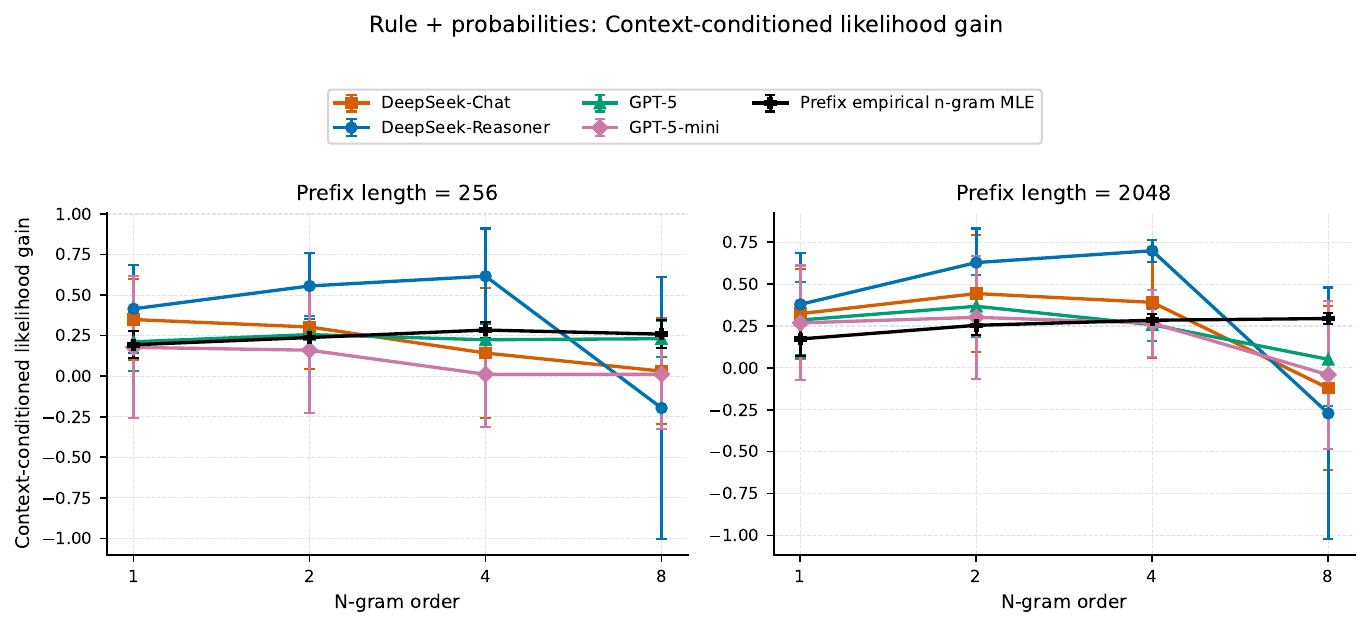}
    \caption{
    Prefix-only empirical n-gram MLE baseline under the rule-and-probabilities setting, evaluated by context-conditioned likelihood gain. 
    LLMs can benefit from low-order context, but this benefit often collapses at order 8, while the empirical n-gram baseline remains comparatively stable.
    }
    \label{fig:app_one_player_no_seed_mle_cclg}
\end{figure}

Figure~\ref{fig:app_one_player_no_seed_mle_wjs} and Figure~\ref{fig:app_one_player_no_seed_mle_cclg} compare LLM generations with a prefix empirical n-gram MLE baseline. 
Unlike the closed-pool MLE rule executor, this baseline does not use transition seeds, candidate rule pools, hidden transition tables, gold continuations, or base probabilities. 
It estimates transition distributions only from the observed prefix. 
For seen contexts, it samples from the empirical n-gram distribution; for unseen contexts, it backs off to lower-order empirical contexts, then prefix unigrams, and finally a uniform vocabulary distribution.

\paragraph{A prefix-only structured estimator remains more stable than LLM generation.}
The no-seed empirical n-gram baseline provides a non-oracle structured reference for testing whether prefix evidence can support faithful context-conditioned generation. 
On weighted JS divergence, the baseline remains competitive with or better than LLMs across orders, especially at higher orders where several LLMs diverge more strongly from the true transition process. 
At order 8, it achieves the best weighted JS divergence among all evaluated methods, showing that the observed prefix still contains usable high-order context information when handled by a structured estimator. 

On CCLG, LLMs sometimes obtain larger gains in low-order settings, indicating that they can exploit simple context structure. 
However, this advantage does not persist at order 8, where several LLMs collapse or become negative, while the empirical n-gram baseline remains relatively stable. 
These results suggest that high-order failure is not merely caused by unavailable hidden rules or insufficient output formatting, since a simple prefix-only estimator can use the observed context more reliably. 
At the same time, the baseline should not be interpreted as recovering the full high-order transition table; rather, it gives a structured reference that uses observed contexts when available and backs off under sparsity.

\end{document}